\documentclass{article}

\usepackage[utf8]{inputenc}
\usepackage{kotex}
\usepackage{microtype}
\usepackage{graphicx}
\graphicspath{{images_paper/}}
\usepackage{subcaption}
\usepackage{booktabs}
\usepackage[breaklinks=true]{hyperref}

\usepackage[accepted]{icml2026}

\makeatletter
\g@addto@macro{\UrlBreaks}{\UrlOrds}
\makeatother

\usepackage{amsmath}
\usepackage{amssymb}
\usepackage{mathtools}
\usepackage{amsthm}
\usepackage[capitalize,noabbrev]{cleveref}
\usepackage{xcolor}
\definecolor{customgreen}{HTML}{4EA72E}
\definecolor{customblue}{HTML}{4E95D9}
\definecolor{customred}{HTML}{FF0000}
\usepackage{tikz}
\usepackage[most]{tcolorbox}

\usepackage{changepage}
\usepackage{multirow}
\usepackage{array}
\usepackage{float}
\usepackage{longtable}
\usepackage{algorithm}
\usepackage{algorithmic}

\theoremstyle{plain}

\theoremstyle{definition}

\theoremstyle{remark}

\usepackage[textsize=tiny]{todonotes}

\newcommand{\bluecheck}{\textcolor{customblue}{\checkmark}}

\icmltitlerunning{Discovering Ordinary Differential Equations with LLM-Based Qualitative and Quantitative Evaluation}

\begin{document}

\twocolumn[
\icmltitle{Discovering Ordinary Differential Equations with LLM-Based Qualitative and Quantitative Evaluation}
  
  \icmlsetsymbol{equal}{*}

  \begin{icmlauthorlist}
    \icmlauthor{Sum Kyun Song}{equal,cau}
    \icmlauthor{Bong Gyun Shin}{equal,dju}
    \icmlauthor{Jae Yong Lee}{cau}
  \end{icmlauthorlist}

  \icmlaffiliation{cau}{Department of Artificial Intelligence, Chung-Ang University, Seoul, Republic of Korea}
  \icmlaffiliation{dju}{Department of AI Big Data, Daejin University, Pocheon-si, Republic of Korea}

  \icmlcorrespondingauthor{Jae Yong Lee}{jaeyong@cau.ac.kr}
  \icmlkeywords{Symbolic Regression, Ordinary Differential Equations, Large Language Models, Scientific Discovery}
  \vskip 0.3in
]

\printAffiliationsAndNotice{\icmlEqualContribution}

\begin{abstract}
  Discovering governing differential equations from observational data is a fundamental challenge in scientific machine learning. Existing symbolic regression approaches rely primarily on quantitative metrics; however, real-world differential equation modeling also requires incorporating domain knowledge to ensure physical plausibility. To address this gap, we propose DoLQ, a method for discovering ordinary differential equations with LLM-based qualitative and quantitative evaluation. DoLQ employs a multi-agent architecture: a Sampler Agent proposes dynamic system candidates, a Parameter Optimizer refines equations for accuracy, and a Scientist Agent leverages an LLM to conduct both qualitative and quantitative evaluations and synthesize their results to iteratively guide the search. Experiments on multi-dimensional ordinary differential equation benchmarks demonstrate that DoLQ achieves superior performance compared to existing methods, not only attaining higher success rates but also more accurately recovering the correct symbolic terms of ground truth equations. Our code is available at \url{https://github.com/Bon99yun/DoLQ}.
\end{abstract}

\section{Introduction}


To understand complex dynamical systems, scientists have long pursued concise mathematical laws \citep{wigner1960unreasonable}. Differential equations are essential tools for describing the dynamic behavior of systems across various fields including physics, chemistry, and biology, and when the governing equations are known, it becomes possible to predict and control the future states of the system. However, in many real-world systems, the form of governing equations is not clearly known, and one must infer them from observational data. Consequently, there is a growing need for methodologies capable of deriving interpretable, explicit mathematical formulas directly from data \citep{rudin2019stop}.

\begin{figure}[t]
  \centering
  \includegraphics[width=0.9\columnwidth]{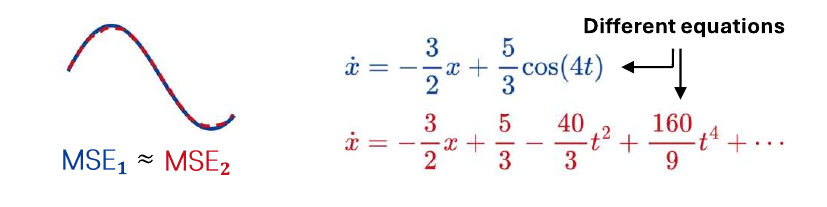}
  \caption{Similar MSE does not imply correct discovery: even with a low MSE, the identified equation may differ from the true one, suggesting that quantitative metrics alone are insufficient and that qualitative evaluation is therefore necessary.}
  \label{fig:intro_physical_meaning_main}
\end{figure}

\begin{figure*}[t]
  \centering
  \includegraphics[width=\textwidth]{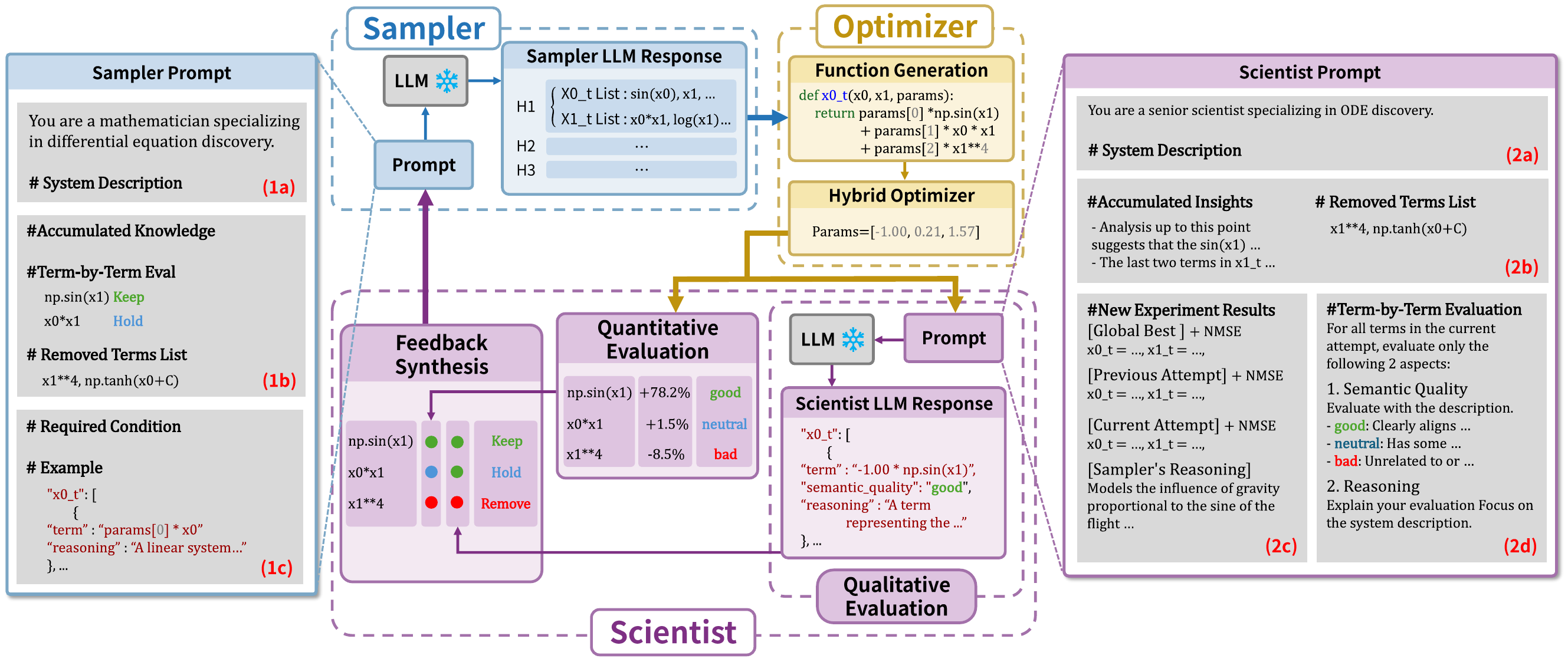}
  \caption{Overview of the DoLQ framework for LLM-based ODE discovery. The framework operates through an iterative loop among three components: (1) the Sampler Agent proposes candidate terms with physical justifications based on the system description and Scientist Agent's feedback; (2) the Parameter Optimizer makes functions and fits their parameters; and (3) the Scientist Agent evaluates each term through qualitative semantic assessment and quantitative iterative comparison, determining actions that guide subsequent iterations.}
  \label{fig:overview}
\end{figure*}

Sparse Identification of Nonlinear Dynamics (SINDy) \citep{brunton2016discovering} addresses this equation discovery problem by constructing a predefined library of basis functions and identifying system dynamics via sparse regression. While efficient, SINDy struggles with selecting appropriate libraries for systems with unknown functional forms. Alternatively, Symbolic Regression (SR) approaches seek to find a symbolic function $f$ that maps input variables $\boldsymbol{x}$ to observed outputs $y$, satisfying $y \approx f(\boldsymbol{x})$, from given data \citep{koza1992genetic, schmidt2009distilling}. SR extends to dynamical system discovery by identifying the right-hand side $\boldsymbol{f}$ of an ordinary differential equation (ODE), formulated as $\dot{\boldsymbol{x}} = \boldsymbol{f}(t, \boldsymbol{x})$, which governs the system. SR-based methods overcome the library constraint through autonomous discovery of differential equations. For instance, ODEformer \citep{d2023odeformer} leverages the Transformer architecture to generate ODEs from observational data, enabling the discovery of diverse equation structures. However, relying solely on numerical data, these methods often fail to verify the physical meaning or consistency of the generated terms.

More recently, LLMs have entered the SR landscape. These approaches replace the traditional evolutionary algorithms used for equation generation with LLMs \citep{shojaee2024llm, grayeli2024symbolic}. By leveraging the extensive prior knowledge and hypothesis generation capabilities of LLMs, these studies have outperformed traditional approaches. However, their evaluation of candidate equations still relies primarily on quantitative criteria such as mean squared error (MSE) or model complexity. As Figure~\ref{fig:intro_physical_meaning_main} illustrates, equations with similarly low numerical error can nevertheless imply different physical mechanisms. This highlights the limitation of purely numerical evaluation and motivates the need for qualitative assessment. Since differential equations in science are often grounded in natural phenomena and human intuition, the physical plausibility of discovered terms becomes essential.

To address these limitations, we propose \textbf{DoLQ}, a novel framework that \textbf{D}iscovers \textbf{O}rdinary differential equations with \textbf{L}LM-based \textbf{Q}ualitative and \textbf{Q}uantitative evaluation. DoLQ is a collaborative framework composed of three key components: a Sampler Agent, a Parameter Optimizer, and a Scientist Agent. The Sampler Agent leverages the semantic reasoning of LLMs to propose candidate terms grounded in qualitative physical justification. The Parameter Optimizer determines the optimal numerical parameters for proposed terms. Crucially, the Scientist Agent enriches the search with qualitative evaluation by assessing whether candidate equations are semantically and physically plausible rather than relying on numerical fit alone. By combining physical reasoning with numerical validation, DoLQ can identify governing laws for multi-dimensional ODEs beyond simple polynomial forms across diverse dynamical systems.

Our main contributions are summarized as follows:

\begin{itemize}
    \item \textbf{Integrated qualitative-quantitative framework:} DoLQ explicitly embeds qualitative physical reasoning into the search loop, guiding the discovery process toward equations that satisfy both numerical accuracy and physical interpretability while maintaining compact equation structures.
    
    \item \textbf{Superior performance on complex systems:} We demonstrate that DoLQ achieves the highest success rates in identifying correct governing equations, particularly excelling where existing methods struggle to capture the correct dynamics.
    
    \item \textbf{Impact of qualitative evaluation:} We demonstrate the effectiveness of qualitative reasoning in filtering out physically implausible terms, validating that semantic assessment plays a crucial role in guiding the discovery process toward correct governing equations.
\end{itemize}

\section{Problem Formulation}

We consider a data-driven discovery task where the input consists of two parts: numerical observations and a semantic description. Let $\mathcal{D} = \{(t_i, \boldsymbol{x}(t_i))\}_{i=0}^{N-1}$ be the time-series state observations, where $\boldsymbol{x}(t_i) \in \mathbb{R}^d$ represents the system state at time $t_i$. Additionally, a system description $\mathcal{T}$ is provided, which represents the domain-specific prior knowledge that a practitioner would possess when approaching the discovery task.

Our goal is to discover an interpretable system of ODEs $\dot{\boldsymbol{x}}(t) = \boldsymbol{f}(t, \boldsymbol{x})$, where $\boldsymbol{f} = [f_0, \dots, f_{d-1}]^T$ is composed of functional terms proposed by an LLM that synthesizes the domain knowledge from $\mathcal{T}$ with numerical patterns in $\mathcal{D}$. The objective is to identify $\boldsymbol{f}$ that minimizes the Mean Squared Error (MSE). For the $j$-th dimension, the integral MSE is defined as
\begin{equation}\label{eq:mse}
\sum_{i=0}^{N-1} \left( x_j(t_i) - \left(x_j(t_0) + \int_{t_0}^{t_i} f_j(s, \boldsymbol{x}) \, ds\right) \right)^2.
\end{equation}
Here, the integral represents the state trajectory obtained via numerical integration of the ODE system starting from the initial condition $\boldsymbol{x}(t_0)$.

Specifically, $\mathcal{T}$ comprises facts known before data collection and without reference to mathematical formulas, such as physical principles and qualitative assessments. While providing this context, $\mathcal{T}$ strictly prohibits specifying mathematical structures; for example, describing ``air resistance'' is permitted, whereas explicit terms like $-cx^2$ or $\sin(x)$ are not. This ensures the model leverages physical intuition without knowledge of the ground-truth symbolic forms.

\section{Methodology}

With the problem formulation in place, we now describe the DoLQ framework and its components in detail. DoLQ is an iterative framework consisting of a Sampler Agent for generating symbolic candidates, a Parameter Optimizer for estimating coefficients, and a Scientist Agent for qualitative and quantitative evaluation (Figure~\ref{fig:overview}). This cycle progressively refines the discovered equations by leveraging both domain knowledge and numerical patterns, guiding the search towards physically consistent and accurate models.

\subsection{Sampler Agent}

The Sampler Agent is an LLM-based component that proposes candidate terms for each dimension of the system of ODEs, along with natural language justifications grounded in the system description and feedback from the Scientist Agent. See Appendix~\ref{app:sampler_prompt} for detailed prompt examples.

\textbf{Sampler prompt.} The input prompt consists of three components: (1a) the role specification stating the SR task and the system description $\mathcal{T}$ providing domain-specific context and physical principles; (1b) Scientist Agent guidance, including accumulated knowledge synthesized from previous iterations, term-by-term evaluation results (\textit{keep} or \textit{hold}), and the set of removed term skeletons marked as ineffective; (1c) technical constraints including output format requirements and illustrative examples.

\textbf{Sampler LLM response.}\label{sec:sampler_output} The Sampler Agent generates ODE candidates in a structured format, pairing each symbolic term with a natural language justification grounded in the system description. Crucially, each term is formatted as an executable Python snippet utilizing state variables \texttt{x} and parameter placeholders \texttt{params[...]} (e.g., \texttt{params[0] * x}). As illustrated in Figure~\ref{fig:overview}, the output is organized into multiple candidate hypotheses. For each hypothesis, a distinct set of terms is generated for every dimension of the system. Each term is accompanied by its physical justification and reasoning, allowing the framework to explore diverse structural assumptions while maintaining the ability to evaluate and ablate distinct physical mechanisms.

\subsection{Parameter optimizer}

The Parameter Optimizer takes term lists proposed by the Sampler Agent and refines them into accurate differential equations through two steps: constructing executable functions and optimizing their parameters.

\textbf{Function construction.} For each set of terms, we construct an executable skeleton function by instantiating the symbolic terms with trainable parameters. This function represents a candidate differential equation $f_j(t, \boldsymbol{x}; \boldsymbol{\theta})$ for the $j$-th dimension, where $\boldsymbol{\theta}$ denotes the parameters to be optimized. See Appendix~\ref{app:skeleton_conversion} for detailed examples.

\textbf{Hybrid optimization.} We find the optimal parameters $\boldsymbol{\theta}$ that minimize the residual MSE:
\begin{equation}\label{eq:residual}
\sum_{i=0}^{N-1} \left( \dot{x}_j(t_i) - f_j(t_i, \boldsymbol{x}; \boldsymbol{\theta}) \right)^2,
\end{equation}
where $\dot{x}_j(t_i)$ denotes the numerically estimated derivative at time $t_i$, computed via finite differences from $\mathcal{D}$. 

We adopt residual MSE instead of integral MSE (Eq.~\ref{eq:mse}) for two reasons: (1) integral MSE requires costly numerical integration at each optimization step, and (2) in multi-dimensional systems, errors in one dimension cause trajectory divergence during integration, resulting in poor scores even for correctly identified dimensions. Residual MSE evaluates each dimension independently, enabling proper credit assignment. We therefore use residual MSE throughout optimization.

While many SR studies utilize BFGS \citep{fletcher2013practical}, it is known to be sensitive to initialization and often fails to converge if the starting values are not well-posed (see Section~\ref{sec:ablation_optimizer}). Therefore, we employ a hybrid strategy that first applies differential evolution \citep{storn1997differential} to identify a promising parameter region, followed by BFGS optimization for refinement. In practice, we evaluate all three strategies—BFGS alone, differential evolution alone, and the hybrid method—selecting the candidate with the lowest MSE.

\subsection{Scientist Agent}

The Scientist Agent re-evaluates the optimized candidate equations $\boldsymbol{f}$ received from the Parameter Optimizer. It assesses each candidate term from two complementary perspectives: qualitative evaluation based on the system description and quantitative contribution measured through iterative comparison. Upon completing these evaluations, the agent synthesizes both results to provide feedback and recommended actions to the Sampler Agent.

\subsubsection{Quantitative evaluation} \label{sec:quantitative_eval}
For quantitative evaluation, the Scientist Agent assesses which terms in the optimized equation are truly necessary. We perform a simple ablation test. For each term in the equation, we temporarily remove it by setting its coefficient to zero and recalculate the residual MSE. If the error significantly increases without the term, it is contributing meaningfully to the fit. If the error stays the same or decreases, the term is unnecessary or harmful and should be removed. 

Based on this ablation analysis, terms are classified into three categories: \textit{good} (terms whose removal significantly increases error, indicating positive contribution), \textit{neutral} (terms whose removal has negligible impact on error), and \textit{bad} (terms whose removal decreases error, indicating negative impact). To make the ablation criterion concrete, consider the illustrative example shown in the middle panel of Figure~\ref{fig:overview}. In that example, \texttt{np.sin(x1)} is classified as \textit{good} because removing it increases the MSE by 78.2\%, whereas \texttt{x1**4} is classified as \textit{bad} because its removal decreases the overall error, indicating that the term was inflating the fit. See Appendix~\ref{app:implementation_details} for details.

\subsubsection{Qualitative evaluation}
For qualitative evaluation, the Scientist LLM assesses how well each term aligns with the physical or mathematical meaning described in the system description. See Appendix~\ref{app:scientist_prompt} for details.  

\textbf{Scientist prompt.} 
The input prompt to the Scientist Agent comprises four components: (2a) context information including the current iteration progress and system description, enabling the agent to understand the search trajectory and physical background; (2b) accumulated insights from previous evaluations and constraints listing previously removed term structures to prevent redundant proposals in subsequent iterations; (2c) experimental results presenting the Global Best equation, Previous Attempt, and Current Attempt, each with per-dimension MSE values, optimized coefficients, and the Sampler's physical justifications for each proposed term, allowing the agent to assess performance improvements and evaluate the physical plausibility of new terms relative to established baselines; (2d) evaluation instructions specifying semantic quality grading criteria and requirements for physical reasoning, ensuring consistent evaluation standards across iterations. 

\textbf{Scientist LLM response.}\label{sec:scientist_output} 
The Scientist Agent produces a structured response containing two components. The first component, term evaluations, provides dimension-specific assessments where each term receives a semantic quality grade paired with physical reasoning. Terms are assigned \textit{good} if they clearly align with physical principles described in the system description, \textit{neutral} if they are partially relevant but lack a clear physical basis, or \textit{bad} if they lack physical meaning or contradict the system description. The second component, updated insight, synthesizes the current analysis into accumulated knowledge, identifying captured physical mechanisms and aspects requiring further investigation. This structured format enables systematic tracking of modeling progress.

\subsubsection{Feedback synthesis.} The final action for each term is determined by combining both evaluations: (1) if semantic quality is \textit{bad}, immediately \textit{remove} regardless of performance impact; (2) if both semantic quality and performance impact are \textit{good}, assign \textit{keep}; (3) all other combinations are assigned \textit{hold}. Terms receiving \textit{hold} for two consecutive iterations are converted to \textit{remove}. Removed term skeletons, where parameters are replaced with placeholder symbols, are recorded to prevent re-proposal. To avoid permanently discarding potentially valid terms, a probabilistic forgetting mechanism periodically clears these entries, allowing re-exploration of previously rejected candidates. The detailed algorithm and decision logic are provided in Appendix~\ref{app:term_retention}.

\begin{table}[t]
  \centering
  \caption{Benchmark ODEs and their temporal domains for ID and ID-Ext evaluation.}
  \label{tab:benchmark_odes} 
  \small
  \resizebox{\columnwidth}{!}{
  \begin{tabular}{llcc}
  \toprule
  \multirow{2}{*}{Benchmark} & \multirow{2}{*}{Equation} & \multicolumn{2}{c}{Time Domain} \\
  \cmidrule(lr){3-4}
   & & ID & ID-Ext \\
  \midrule
  \multirow{2}{*}{SIR(2D)} & $\dot{x}_0 = -0.4 x_0 x_1$ & \multirow{2}{*}{$[0, 2]$} & \multirow{2}{*}{$[0, 4]$} \\
   & $\dot{x}_1 = 0.4 x_0 x_1 - 0.314 x_1$ & & \\
  \cmidrule(lr){1-4}
  \multirow{2}{*}{CDIMA(2D)} & $\dot{x}_0 = 8.9 - \frac{4.0 x_0 x_1}{x_0^2 + 1} - x_0$ & \multirow{2}{*}{$[0, 5]$} & \multirow{2}{*}{$[0, 10]$} \\
   & $\dot{x}_1 = 1.4 x_0 \left(1 - \frac{x_1}{x_0^2 + 1}\right)$ & & \\
  \cmidrule(lr){1-4}
  \multirow{2}{*}{Glider(2D)} & $\dot{x}_0 = -x_0^2/5.0 - \sin(x_1)$ & \multirow{2}{*}{$[0, 5]$} & \multirow{2}{*}{$[0, 10]$} \\
   & $\dot{x}_1 = x_0 - \cos(x_1)/x_0$ & & \\
  \cmidrule(lr){1-4}
  \multirow{4}{*}{Glider(4D)} & $\dot{x}_0 = -9.81 \sin(x_1) - 0.030625 x_0^2$ & \multirow{4}{*}{$[0, 5]$} & \multirow{4}{*}{$[0, 10]$} \\
   & $\dot{x}_1 = -\frac{9.81 \cos(x_1)}{x_0} + 0.6125 x_0$ & & \\
   & $\dot{x}_2 = x_0 \cos(x_1)$ & & \\
   & $\dot{x}_3 = x_0 \sin(x_1)$ & & \\
  \bottomrule
  \end{tabular}
  }
  \end{table}

  \begin{table*}[t]
    \centering
    \caption{Quantitative performance comparison measured by NMSE. We report the dimension-averaged results. Bold with underline indicates the best, bold indicates the second best. NaN indicates solver failure or numerical instability.}
    \label{tab:quantitative_full_split}
    \small
    \resizebox{\textwidth}{!}{
    \begin{tabular}{lcccccccccccc}
    \toprule
    \multirow{3}{*}{Model} 
    & \multicolumn{4}{c}{SIR(2D)} 
    & \multicolumn{4}{c}{CDIMA(2D)} 
    & \multicolumn{4}{c}{Glider(4D)} \\
    \cmidrule(lr){2-5} \cmidrule(lr){6-9} \cmidrule(lr){10-13}
    & \multicolumn{2}{c}{Residual} & \multicolumn{2}{c}{Integral} 
    & \multicolumn{2}{c}{Residual} & \multicolumn{2}{c}{Integral} 
    & \multicolumn{2}{c}{Residual} & \multicolumn{2}{c}{Integral} \\
    \cmidrule(lr){2-3} \cmidrule(lr){4-5} \cmidrule(lr){6-7} \cmidrule(lr){8-9} \cmidrule(lr){10-11} \cmidrule(lr){12-13}
    & ID & ID-Ext & ID & ID-Ext & ID & ID-Ext & ID & ID-Ext & ID & ID-Ext & ID & ID-Ext \\
    \midrule
    ICSR   
    & 2.80e-8 & 4.62e-9 & \textbf{1.42e-8} & \textbf{1.19e-8} 
    & 5.16e-4 & \textbf{6.06e-5} & 3.78e-4 & \textbf{2.56e-4} 
    & 2.55e-2 & 1.16e0 & 2.55e-1 & 3.19e0 \\
    L\small{A}SR 
    & 1.78e-8 & 7.19e-9 & \textbf{\underline{7.60e-9}} & \textbf{\underline{8.61e-9}} 
    & 2.91e-1 & 1.14e1 & 2.90e-1 & 4.85e1 
    & 2.50e-2 & 3.52e0 & 4.94e-1 & 1.00e3 \\
    LLM-SR 
    & \textbf{\underline{1.65e-8}} & \textbf{4.14e-9} & 1.61e-8 & 2.79e-8 
    & \textbf{2.67e-3} & 5.08e-3 & \textbf{3.14e-3} & 9.38e-3 
    & \textbf{\underline{9.95e-7}} & \textbf{\underline{1.12e-6}} & \textbf{\underline{8.54e-7}} & \textbf{\underline{4.76e-6}} \\
    EDL    
    & 1.86e1 & NaN & 1.40e1 & NaN 
    & 1.07e5 & NaN & 1.33e5 & NaN 
    & 1.48e4 & NaN & 1.37e4 & NaN \\
    \midrule
    DoLQ(ours)
    & \textbf{1.71e-8} & \textbf{\underline{3.98e-9}} & 1.55e-8 & 2.63e-8 
    & \textbf{\underline{2.29e-8}} & \textbf{\underline{2.69e-8}} & \textbf{\underline{2.35e-8}} & \textbf{\underline{1.15e-7}} 
    & \textbf{1.22e-6} & \textbf{1.53e-6} & \textbf{8.29e-7} & \textbf{5.21e-6} \\
    \bottomrule
    \end{tabular}
    }
    \end{table*}

\section{Experiments}

\subsection{Benchmark and dataset}

To evaluate our method, we select seven ODE problems from ODEbench \citep{d2023odeformer}, a comprehensive benchmark for ODE discovery, with their physical descriptions and domain knowledge supplemented by referencing Wikipedia and related literature. For four-dimensional systems, we additionally construct benchmark problems following the same approach. Thus, we use a total of eight ODE datasets. We focus on four representative systems for detailed analysis in the main experiments. We use three systems for quantitative performance comparison as shown in Table~\ref{tab:quantitative_full_split}: SIR(2D), CDIMA(2D), and Glider(4D). Additionally, we use Glider(2D) for structural equation analysis as shown in Figure~\ref{fig:equation_comparison}. Details on the complete benchmark suite are provided in Appendix~\ref{appendix:dataset}.

\textbf{Benchmark problems.}
Table~\ref{tab:benchmark_odes} shows four representative systems from the benchmark suite. We categorize problems by their functional complexity: systems with polynomial terms versus those with rational, trigonometric, or exponential terms. Table~\ref{tab:quantitative_full_split} presents quantitative evaluation on three systems: SIR(2D) models disease spread, CDIMA(2D) models chemical kinetics with nonlinear saturation terms, and Glider(4D) captures flight dynamics. Figure~\ref{fig:equation_comparison} shows structural equation comparison on a dimensionless Glider(2D) variant for clearer visualization.

\textbf{Extended in-domain evaluation.} To assess the generalization capability of discovered equations, we build upon the evaluation framework of \citet{shojaee2024llm}, which proposes in-domain (ID) and out-of-domain (OOD) test sets. While OOD evaluation is ideal for testing generalization, it is impractical in our setting because integral MSE-based validation requires initial conditions that may be unavailable in the extended regime. To address this limitation, we introduce extended-ID (ID-Ext) test sets that extend the temporal or state-space coverage while retaining observed initial conditions, enabling rigorous testing of whether discovered equations maintain predictive accuracy over longer horizons.

\textbf{Data generation.} 
State trajectories for both ID and ID-Ext regimes were generated using the fourth-order Runge--Kutta method. The time derivatives $\dot{x}$ in Eq.~\eqref{eq:residual} were obtained through numerical differentiation based on the finite difference method. For each ODE, we prepared textual descriptions by referencing relevant research literature, with equation names and explicit formula structures removed to prevent direct inference of the symbolic form. The ID regime is used for model training and selection, while the ID-Ext regime serves to evaluate extrapolation capability under extended temporal or spatial ranges. The time domain ranges for ID and ID-Ext in Table~\ref{tab:benchmark_odes} were selected based on the temporal intervals where the data distribution exhibited the most significant differences, ensuring rigorous evaluation of extrapolation capability.

\begin{figure*}[t]
  \centering
  \includegraphics[width=\textwidth]{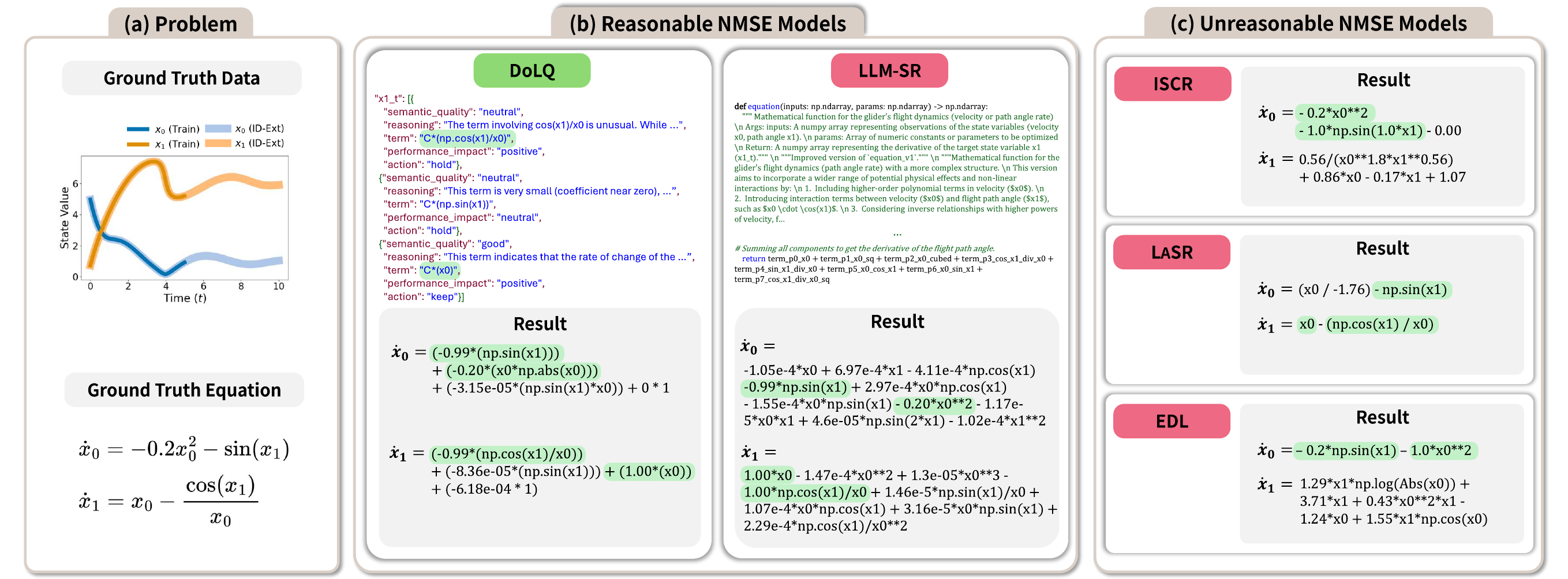}
  \caption{Equation comparison on the 2D dimensionless Glider system. Ground truth terms are highlighted in green. Group (b) shows models with reasonable NMSE: DoLQ achieves integral NMSE $<10^{-3}$ across all dimensions, and LLM-SR exhibits the lowest NMSE among the remaining methods. Group (c) shows models with higher NMSE. Each model's discovered equation terms and their corresponding optimized coefficients are displayed.}
  \label{fig:equation_comparison}
\end{figure*}

\subsection{Baseline models and experimental setup}

We evaluate DoLQ against representative LLM-based SR methods: ICSR~\cite{merler2024context}, L{\small{A}}SR~\cite{grayeli2024symbolic}, LLM-SR~\cite{shojaee2024llm}, and EDL~\cite{du2024large}. We selected these baselines based on their capability to incorporate textual descriptions of the system $\mathcal{T}$ via prompts and their ability to generate symbolic ODE expressions. Notably, among these baselines, LLM-SR and EDL were originally proposed and evaluated specifically for symbolic discovery of ordinary differential equations. The remaining methods, L{\small{A}}SR and ICSR, were primarily designed for general symbolic regression tasks; nevertheless, we adapt them to our problem setting to enable a fair comparison on ODE discovery.

Most existing SR methods target scalar-valued functions ($f:\mathbb{R}^d \rightarrow \mathbb{R}$), whereas our task requires discovering vector-valued functions ($f:\mathbb{R}^d \rightarrow \mathbb{R}^d$) for ODE systems. To ensure fair comparison, we adapted the baselines to support vector-valued outputs by modifying their prompts and providing the same system description $\mathcal{T}$. All experiments were conducted using Gemini 2.5 Flash Lite~\citep{team2023gemini}. To ensure equivalent LLM call conditions for a fair comparison, we configured the search budget as follows. For DoLQ, we perform 100 iterations. In contrast, baseline models LLM-SR, EDL, and ICSR are configured to perform 100 iterations per dimension, with L{\small{A}}SR set to a comparable expected number of iterations per dimension. A similar budget strategy was maintained for 4D systems; further details are provided in Appendix~\ref{app:implementation_details}.

\subsection{Performance comparison}

We use normalized mean squared error (NMSE) to enable fair comparison across different problems with varying scales.
We evaluate performance using two metrics: integral NMSE and residual NMSE, with the results summarized in Table~\ref{tab:quantitative_full_split}.
For the SIR(2D) system, since the equation form is readily inferable from the description, most models achieve low error rates, successfully identifying the governing dynamics. Notably, many models identify the correct mathematical structure practically from the first iteration, demonstrating that existing methods are capable when the problem structure is relatively straightforward. 

However, the CDIMA(2D) system presents a distinct challenge due to its complex nonlinear saturation terms. In this case, a significant divergence emerges between residual NMSE and integral NMSE. Baseline models often exhibit low residual NMSE yet high integral NMSE, indicating a failure to capture the true underlying physics. Only DoLQ consistently achieves superior performance with stable integral NMSE scores, demonstrating its effectiveness in discovering equations with complex functional forms.

For the Glider(4D) system, both LLM-SR and DoLQ successfully identify the correct governing equations with low NMSE.

\begin{figure}[t]
  \centering
  \includegraphics[width=0.8\columnwidth]{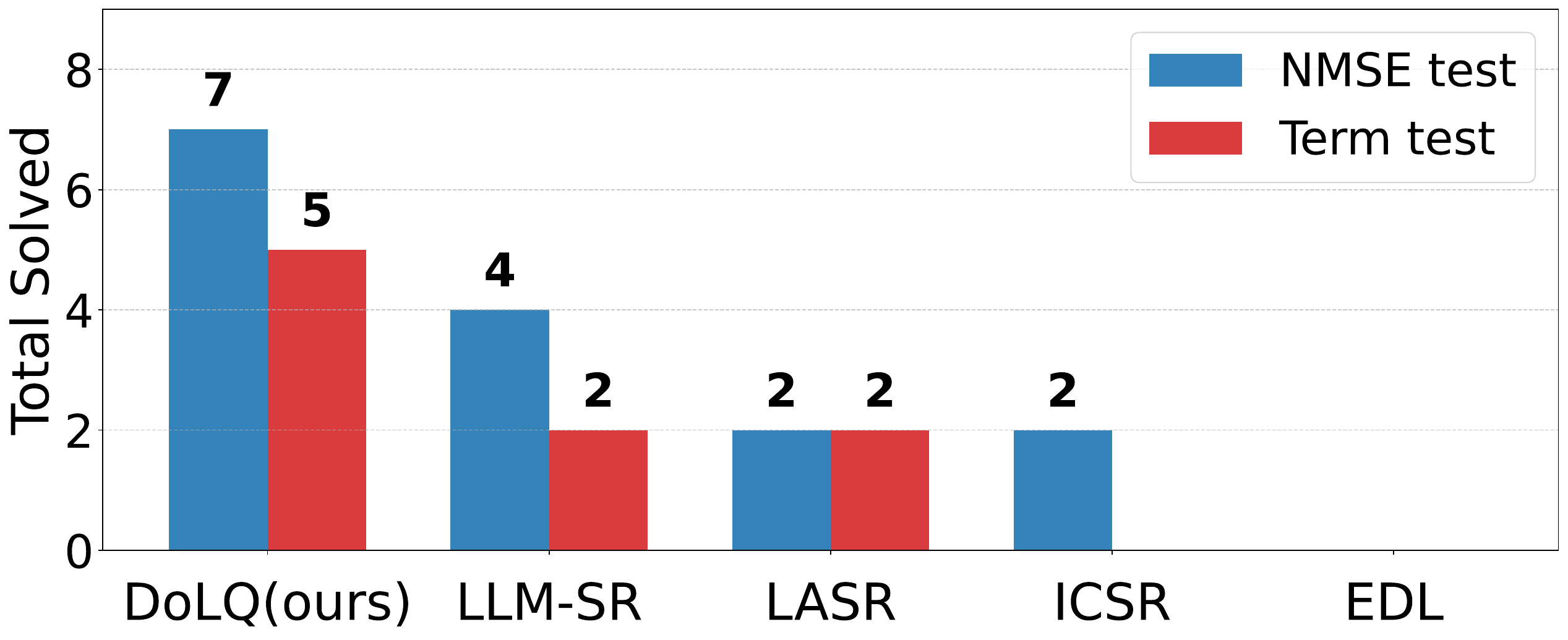}
  \caption{Success scores measured across eight benchmark ODEs. NMSE test: success is achieved if the integral NMSE across all dimensions is $< 10^{-3}$. Term test: success is achieved if the discovered equation matches the ground truth structure after excluding terms with negligible impact.}
  \label{fig:success_scores}
\end{figure}

\begin{figure}[t]
  \centering
  \includegraphics[width=\columnwidth]{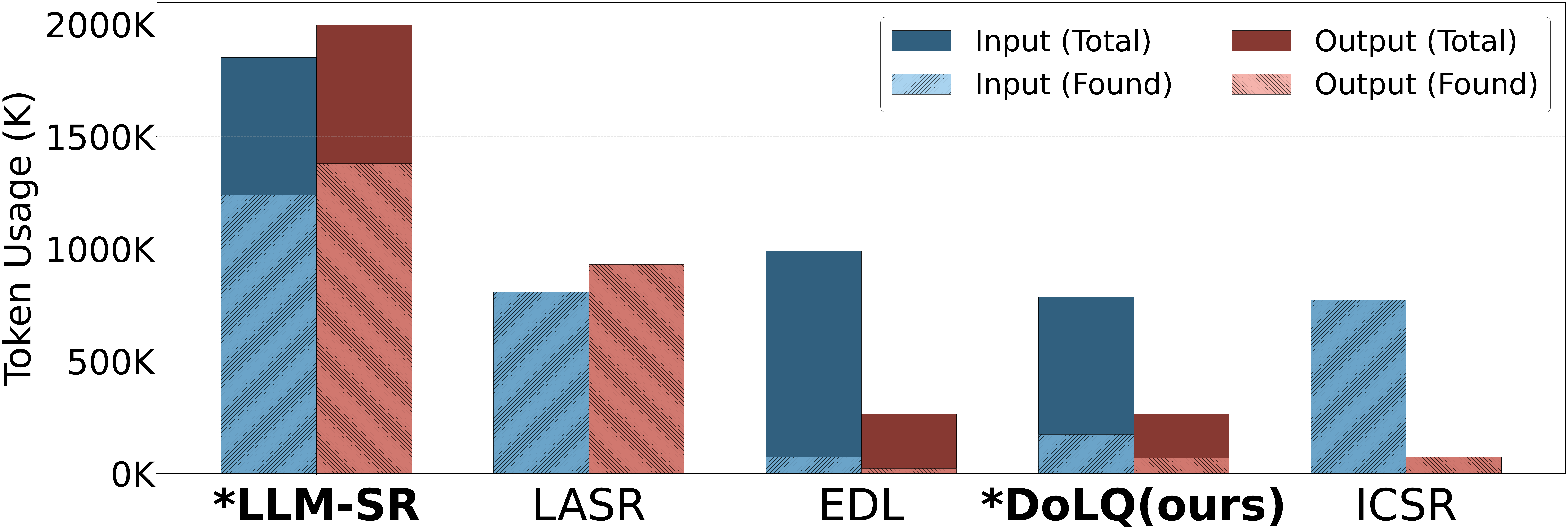}
  \caption{Cumulative token usage on the 4D Glider system. Stars (*) indicate that LLM-SR and DoLQ successfully identified the governing equation. NMSE values reach their minimum on the Glider system.}
  \label{fig:token_usage}
\end{figure}

\subsection{Structural comparison}

Beyond the three representative systems in Table~\ref{tab:benchmark_odes}, we further analyze the learning outcomes on an additional ODE problem: a 2D dimensionless variant of the Glider system. This simplified version retains the essential nonlinear trigonometric dynamics while enabling clearer visualization and interpretation of the discovered terms compared to the full 4D system.

Qualitative examination reveals critical differences in the discovered physical laws. In Figure~\ref{fig:equation_comparison}, we categorize the results on this system into groups with reasonable NMSE (b) and those without (c). Both DoLQ and LLM-SR successfully identify all ground truth terms, whereas other baselines capture only partial dynamics. However, LLM-SR often proposes equation skeletons containing numerous unnecessary terms. In contrast, DoLQ proposes significantly fewer terms while successfully including all ground truth components. This efficiency stems from the Scientist Agent's rigorous quantitative and qualitative assessment; beneficial terms receive positive evaluations, while detrimental terms are explicitly identified and removed, preventing their recurrence in subsequent iterations. Consequently, DoLQ achieves superior performance with a more focused and efficient search process.

\subsection{Overall comparison and discussion}

As shown in Figure~\ref{fig:success_scores}, DoLQ achieves the highest success rate across all evaluation criteria, outperforming baseline methods on both NMSE test and Term test across diverse problem types.

Analyzing baseline characteristics reveals distinct limitations. L{\small{A}}SR demonstrates reasonable performance on problems with simple polynomial terms, producing clean equations with minimal unnecessary components. However, the evolutionary algorithm's search capability is fundamentally limited, failing to discover governing equations for complex systems involving non-polynomial functional forms. LLM-SR achieves moderate NMSE performance (4/8 success rate), yet shows limited success in capturing the correct equation structure. This discrepancy arises from proposing excessive candidate terms, which not only complicates parameter optimization but also generates unnecessarily verbose prompts. As evidenced by Figure~\ref{fig:token_usage}, LLM-SR consumes substantially more tokens than DoLQ, with detailed statistics provided in Appendix~\ref{app:token_usage}.

DoLQ addresses these fundamental limitations through the Scientist Agent's evaluation framework. By rigorously assessing both qualitative evaluation and quantitative evaluation of each proposed term, the Scientist Agent effectively filters out unnecessary candidates early in the search process. This prevents the bloated equation skeletons that plague LLM-SR while maintaining the capacity to discover complex functional forms beyond the reach of evolutionary methods. The result is a framework that achieves superior accuracy across both simple and complex systems, consistently identifying the correct governing equations where baseline methods fail. DoLQ's multi-agent architecture not only discovers physically consistent governing equations with higher success rates but also enables more interpretable results through compact equation structures. By combining qualitative reasoning with quantitative validation, DoLQ establishes a new standard for reliable and accurate ODE discovery in dynamical systems, demonstrating robust performance across diverse problem types while maintaining computational efficiency as an additional benefit.

\section{Component validation}

\subsection{Necessity of the Scientist Agent}
\begin{figure}[t]
  \centering
  \includegraphics[width=\columnwidth]{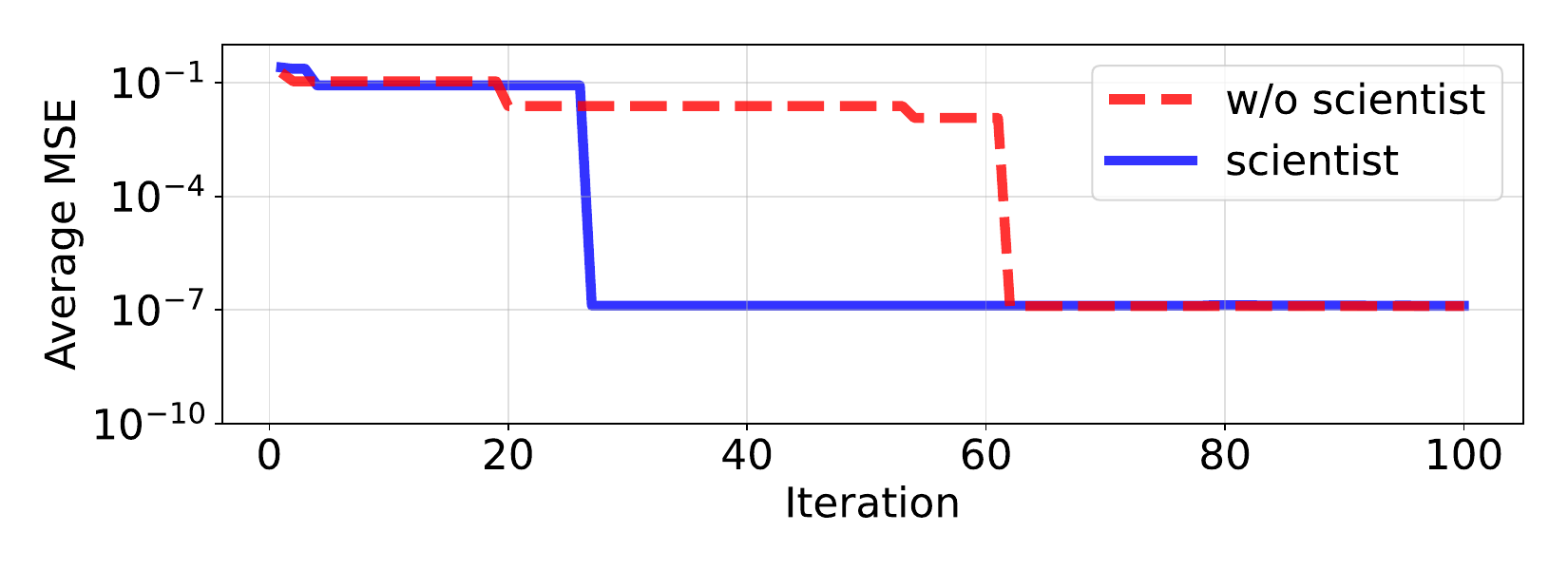}
  \caption{Impact of the Scientist Agent on search convergence for the Glider(2D) problem. DoLQ with the Scientist Agent discovers the correct equation at iteration 27, while the baseline without the Scientist Agent finds it at iteration 62, demonstrating that qualitative feedback accelerates convergence.}
  \label{fig:ablation_scientist}
\end{figure}

To demonstrate the critical role of qualitative reasoning, we conduct an ablation study comparing DoLQ against a baseline variant labeled ``w/o Scientist,'' where the Scientist Agent is removed from the loop. In this configuration, the Sampler Agent directly receives the equation skeletons, optimized coefficients, and NMSE scores from the Parameter Optimizer as part of the prompt for the next iteration, bypassing the qualitative evaluation step. As illustrated in Figure~\ref{fig:ablation_scientist}, the full DoLQ framework containing the Scientist Agent exhibits significantly faster convergence than the baseline. An analysis of the reasoning logs confirms that the Scientist Agent plays a pivotal role in the discovery process; its feedback effectively filters out physically implausible terms early in the search, guiding the model toward the correct governing equations more efficiently.


Unless otherwise noted, all methods were executed three times under the same configuration, and we report the best run; Figure~\ref{fig:ablation_scientist} shows one representative convergence trace generated under this protocol rather than an average over runs. Specifically, the plotted DoLQ trace corresponds to the best of these three runs.

\begin{figure}[t]
  \centering
  \includegraphics[width=\columnwidth]{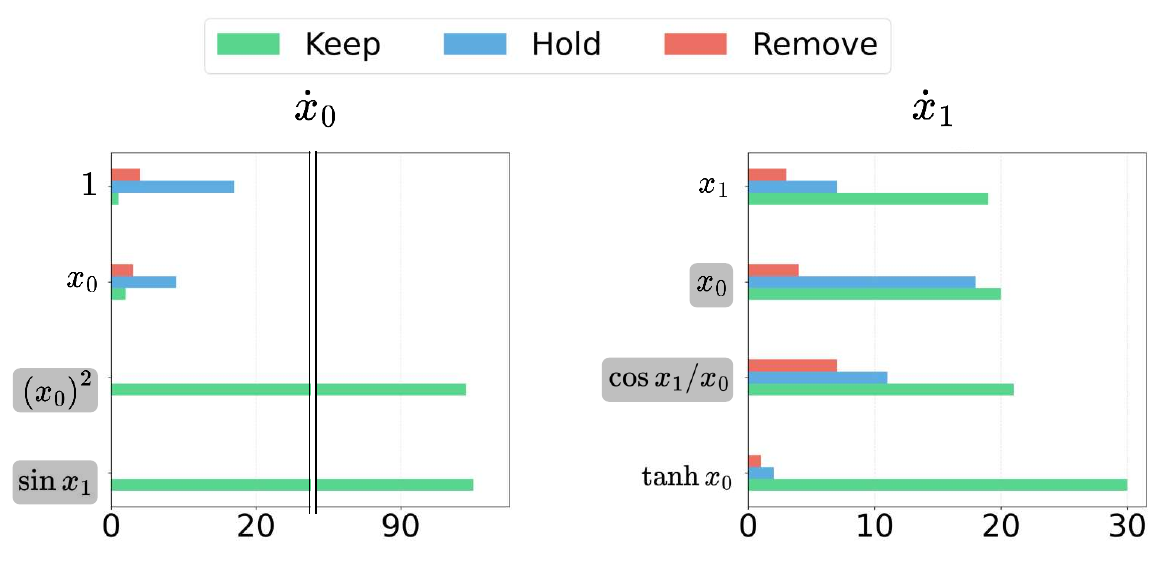}
  \caption{Action frequencies for the top-ranked terms in each target dimension during DoLQ execution on the Glider(2D) problem. Highlighted terms appear in the ground-truth equation and consistently receive high \textit{keep} frequency, indicating that the Scientist Agent effectively preserves physically meaningful terms.}
  \label{fig:scientist_analysis}
\end{figure}

To further understand how the Scientist Agent operates within DoLQ, we analyze its decision patterns during the discovery process. Figure~\ref{fig:scientist_analysis} reveals that ground truth terms consistently receive high \textit{keep} frequencies, demonstrating the Scientist Agent's effectiveness in identifying physically relevant terms. Notably, even when these terms occasionally receive \textit{remove} recommendations, they reappear frequently across iterations due to the probabilistic sampling mechanism that considers keep/remove probabilities, global best information, and previous iteration history, enabling robust convergence toward correct equations.

\subsection{Effectiveness of hybrid parameter optimization}
\label{sec:ablation_optimizer} 
\begin{figure}[t]
  \centering
  \includegraphics[width=\columnwidth]{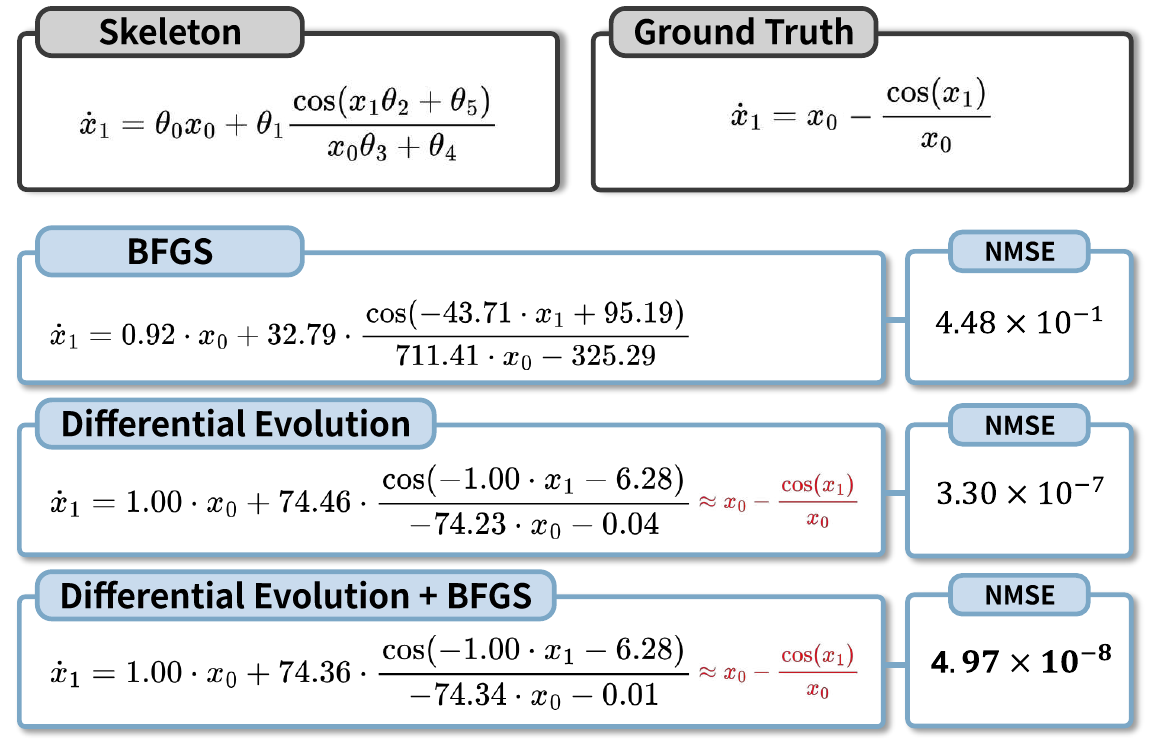}
  \caption{BFGS alone can fail even with a structurally correct skeleton, whereas the hybrid optimizer successfully recovers the correct solution. This illustrates why hybrid parameter optimization is necessary.}
  \label{fig:de_optimization}
\end{figure}

Our choice of combining differential evolution with BFGS is motivated by the need for robust parameter estimation in the rugged loss landscapes of symbolic regression. While the LLM may propose a structurally correct equation skeleton, standard gradient-based optimization like BFGS often fails to identify the optimal parameters due to its sensitivity to initialization and susceptibility to local minima. As shown in Figure~\ref{fig:de_optimization}, even when the correct functional form is provided, BFGS yields suboptimal parameter estimation, whereas our hybrid approach—combining the global search of differential evolution with the local refinement of BFGS—successfully locates the global optimum. Crucially, the successful optimization by differential evolution verifies that the skeleton proposed by the LLM is indeed structurally similar to the ground truth. Relying solely on BFGS could lead to the incorrect rejection of a reasonable skeleton due to optimization failure, concealing the true quality of the proposed structure. We provide further analysis on the adoption frequency of each optimizer across different problem complexities in Appendix~\ref{app:optimizer_adoption}.

\section{Related work}

\textbf{Symbolic regression.} 
SR aims to recover governing equations of dynamical systems in the form of explicit mathematical formulas, evolving from evolutionary algorithms \citep{cranmerInterpretableMachineLearning2023} to incorporate physical constraints \citep{udrescu2020aifeynman}, reinforcement learning \citep{petersen2021deep,mundhenk2021symbolic,xu2024reinforcement}, and neural sequence models \citep{biggio2021neural,kamienny2022end,valipour2021symbolicgpt,shojaee2023transformer,yu2024symbolic,xiang2025graph,huang2025improving}. For dynamical systems, recent methods combine neural networks with genetic programming \citep{gaodiscovering} and Transformer-based architectures \citep{d2023odeformer}. Recent advances include symbolic-numeric pre-training \citep{meidani2024snip}, interactive offline reinforcement learning \citep{tian2025interactive}, and grammar-based discovery \citep{omejc2023probabilistic}. However, most existing methods prioritize point-wise or derivative-based fitting metrics without validating long-horizon trajectory consistency. While functional SR has been extensively studied with established benchmarks \citep{zhang2012srbench}, ODE-specific benchmarks \citep{matsubara2022srsd} remain relatively limited, requiring further research \citep{oliveira2018analysing}.

\textbf{LLMs for scientific discovery.} 
LLMs leverage encoded scientific knowledge for equation discovery, optimizing skeletons via evolutionary search \citep{shojaee2024llm}, self-improvement \citep{du2024large}, in-context learning \citep{merler2024context}, concept libraries \citep{grayeli2024symbolic}, and symbol-numeric pre-training \citep{meidani2024snip}. Recent work shows LLMs can interpret discovered expressions \citep{guo2025sr} and diagnose errors through multi-agent reasoning \citep{park2026self}, while formal verification enables structured validation \citep{sun2025enumerate}. Beyond SR, LLMs translate natural language to formal mathematics through autoformalization and reasoning \citep{soroco2025pde} and automate research pipelines \citep{lu2024ai}. However, evaluation still relies on derivative matching rather than trajectory integration, and LLMs' semantic reasoning capacity for validation remains underutilized.

\textbf{Modeling and evaluation of dynamical systems.}
Deep learning approaches for differential equation modeling \citep{czarnecki2017sobolev,dupont2019augmented,greydanus2019hamiltonian,RAISSI2019686,rubanova2019latent,cranmer2020lagrangian,karniadakis2021physics,linot2023stabilized,lee2025finite} achieve robust predictions but lack interpretability. In contrast, SR discovers interpretable governing equations in closed form. Our approach integrates LLM-based physical reasoning into symbolic discovery to improve equation accuracy.

\section{Conclusion}

This paper presents DoLQ, a multi-agent framework integrating qualitative and quantitative evaluation for discovering governing ODEs. By employing LLMs to validate physical plausibility, DoLQ demonstrates substantial improvements in discovery success rates and equation interpretability, particularly for systems with complex functional forms. The systematic integration of semantic reasoning with numerical validation enables consistent identification of correct governing equations across diverse problem types, achieving the highest performance on comprehensive benchmarks. Several limitations warrant further investigation: parameter optimization depends on numerical differentiation, introducing sensitivity to measurement noise, and reasoning fixation occasionally causes premature convergence to implausible hypotheses. Future research will address these through integral-based parameter estimation and diversified hypothesis generation with refined feedback mechanisms. While ODEs benefit from established numerical solvers, extension to PDEs presents considerable difficulty due to the absence of universal solution methods. We plan to investigate advances in PDE solver research to enable reliable equation discovery in spatiotemporal dynamical systems.

\section*{Impact Statement}

This work advances automated scientific discovery by enabling interpretable equation learning from observational data. The enhanced ability to discover governing equations can accelerate research in physics, biology, and engineering, reducing dependence on domain expertise for model development. However, practitioners should remain cautious of potential risks: automated discovery systems may generate spurious correlations or physically implausible models if deployed without proper validation. We emphasize the importance of human oversight in applying discovered equations to real-world systems, particularly in safety-critical domains.

\section*{Acknowledgments}

This work was supported by the Institute of Information \& Communications Technology Planning \& Evaluation (IITP) grant funded by the Korea government (MSIT) [RS-2021-II211341, Artificial Intelligence Graduate School Program (Chung-Ang University)]. This work was supported by the National Research Foundation of Korea (NRF) grant funded by the Korea government (MSIT) (RS-2025-02303239 and RS-2026-25497362).


\bibliography{example_paper}

@article{wigner1960unreasonable,
    title={The unreasonable effectiveness of mathematics in the natural sciences},
    author={Wigner, Eugene P and others},
    journal={Mathematics and science},
    volume={13},
    pages={1--14},
    year={1990},
    publisher={World Scientific}
}

@article{rudin2019stop,
  title={Stop explaining black box machine learning models for high stakes decisions and use interpretable models instead},
  author={Rudin, Cynthia},
    journal={Nature machine intelligence},
    volume={1},
    number={5},
    pages={206--215},
    year={2019},
    publisher={Nature Publishing Group UK London}
}

@book{koza1992genetic,
    title     = {Genetic Programming: On the Programming of Computers by Means of Natural Selection},
    author    = {Koza, John R.},
    year      = {1992},
    publisher = {MIT Press},
    address   = {55 Hayward St, Cambridge, MA, United States},
    isbn      = {978-0-262-11170-6},
    pages     = {680}
}

@article{schmidt2009distilling,
    author = {Michael Schmidt  and Hod Lipson },
    title = {Distilling Free-Form Natural Laws from Experimental Data},
    journal = {Science},
    volume = {324},
    number = {5923},
    pages = {81-85},
    year = {2009},
    doi = {10.1126/science.1165893},
    URL = {https://www.science.org/doi/abs/10.1126/science.1165893},
    eprint = {https://www.science.org/doi/pdf/10.1126/science.1165893}
}

@article{brunton2016discovering,
    author = {Steven L. Brunton  and Joshua L. Proctor  and J. Nathan Kutz },
    title = {Discovering governing equations from data by sparse identification of nonlinear dynamical systems},
    journal = {Proceedings of the National Academy of Sciences},
    volume = {113},
    number = {15},
    pages = {3932-3937},
    year = {2016},
    doi = {10.1073/pnas.1517384113},
    URL = {https://www.pnas.org/doi/abs/10.1073/pnas.1517384113},
    eprint = {https://www.pnas.org/doi/pdf/10.1073/pnas.1517384113}
}

@inproceedings{d2023odeformer,
    title     = {ODEFormer: Symbolic Regression of Dynamical Systems with Transformers},
    author    = {d'Ascoli, St{\'e}phane and others},
    booktitle = {Proceedings of the 12th International Conference on Learning Representations (ICLR)},
    year      = {2024},
    address   = {Vienna, Austria},
    url       = {https://openreview.net/forum?id=TzoHLiGVMo},
    note      = {ICLR 2024}
}

@inproceedings{shojaee2024llm,
    title={{LLM}-{SR}: Scientific Equation Discovery via Programming with Large Language Models},
    author={Parshin Shojaee and Kazem Meidani and Shashank Gupta and Amir Barati Farimani and Chandan K. Reddy},
    booktitle={The Thirteenth International Conference on Learning Representations},
    year={2025},
    url={https://openreview.net/forum?id=m2nmp8P5in}
}

@inproceedings{grayeli2024symbolic,
    author = {Grayeli, Arya and Sehgal, Atharva and Costilla-Reyes, Omar and Cranmer, Miles and Chaudhuri, Swarat},
    booktitle = {Advances in Neural Information Processing Systems},
    doi = {10.52202/079017-1419},
    editor = {A. Globerson and L. Mackey and D. Belgrave and A. Fan and U. Paquet and J. Tomczak and C. Zhang},
    pages = {44678--44709},
    publisher = {Curran Associates, Inc.},
    title = {Symbolic Regression with a Learned Concept Library},
    url = {https://proceedings.neurips.cc/paper_files/paper/2024/file/4ec3ddc465c6d650c9c419fb91f1c00a-Paper-Conference.pdf},
    volume = {37},
    year = {2024}
}

@article{du2024large,
    author={Du, Mengge and Chen, Yuntian and Wang, Zhongzheng and Nie, Longfeng and Zhang, Dongxiao},
    title = {Large language models for automatic equation discovery of nonlinear dynamics},
    journal = {Physics of Fluids},
    volume = {36},
    number = {9},
    pages = {097121},
    year = {2024},
    month = {09},
    issn = {1070-6631},
    doi = {10.1063/5.0224297},
    url = {https://doi.org/10.1063/5.0224297},
    eprint = {https://pubs.aip.org/aip/pof/article-pdf/doi/10.1063/5.0224297/20152367/097121_1_5.0224297.pdf},
}

@inproceedings{merler2024context,
    title = "In-Context Symbolic Regression: Leveraging Large Language Models for Function Discovery",
    author = "Merler, Matteo  and
      Haitsiukevich, Katsiaryna  and
      Dainese, Nicola  and
      Marttinen, Pekka",
    editor = "Fu, Xiyan  and
      Fleisig, Eve",
    booktitle = "Proceedings of the 62nd Annual Meeting of the Association for Computational Linguistics (Volume 4: Student Research Workshop)",
    month = aug,
    year = "2024",
    address = "Bangkok, Thailand",
    publisher = "Association for Computational Linguistics",
    url = "https://aclanthology.org/2024.acl-srw.49/",
    doi = "10.18653/v1/2024.acl-srw.49",
    pages = "427--444",
    ISBN = "979-8-89176-097-4"
}

@article{storn1997differential,
    title   = {Differential Evolution -- A Simple and Efficient Heuristic for Global Optimization over Continuous Spaces},
    author  = {Storn, Rainer and Price, Kenneth},
    journal = {Journal of Global Optimization},
    year    = {1997},
    volume  = {11},
    number  = {4},
    pages   = {341--359},
    doi     = {10.1023/A:1008202821328},
    issn    = {1573-2916},
    url     = {https://doi.org/10.1023/A:1008202821328}
}

@inproceedings{rubanova2019latent,
    author = {Rubanova, Yulia and Chen, Ricky T. Q. and Duvenaud, David K},
    booktitle = {Advances in Neural Information Processing Systems},
    publisher = {Curran Associates, Inc.},
    title = {Latent Ordinary Differential Equations for Irregularly-Sampled Time Series},
    url = {https://proceedings.neurips.cc/paper_files/paper/2019/file/42a6845a557bef704ad8ac9cb4461d43-Paper.pdf},
    volume = {32},
    year = {2019}
}

@inproceedings{dupont2019augmented,
    author = {Dupont, Emilien and Doucet, Arnaud and Teh, Yee Whye},
    booktitle = {Advances in Neural Information Processing Systems},
    publisher = {Curran Associates, Inc.},
    title = {Augmented Neural ODEs},
    url = {https://proceedings.neurips.cc/paper_files/paper/2019/file/21be9a4bd4f81549a9d1d241981cec3c-Paper.pdf},
    volume = {32},
    year = {2019}
}

@article{RAISSI2019686,
    title = {Physics-informed neural networks: A deep learning framework for solving forward and inverse problems involving nonlinear partial differential equations},
    journal = {Journal of Computational Physics},
    volume = {378},
    pages = {686-707},
    year = {2019},
    issn = {0021-9991},
    doi = {https://doi.org/10.1016/j.jcp.2018.10.045},
    url = {https://www.sciencedirect.com/science/article/pii/S0021999118307125},
    author = {M. Raissi and P. Perdikaris and G.E. Karniadakis}
}

@article{karniadakis2021physics,
    title={Physics-informed machine learning},
    author={Karniadakis, George Em and Kevrekidis, Ioannis G and Lu, Lu and others},
    journal={Nature Reviews Physics},
    volume={3},
    pages={422--440},
    year={2021}
}

@inproceedings{greydanus2019hamiltonian,
    author = {Greydanus, Samuel and Dzamba, Misko and Yosinski, Jason},
    booktitle = {Advances in Neural Information Processing Systems},
    publisher = {Curran Associates, Inc.},
    title = {Hamiltonian Neural Networks},
    url = {https://proceedings.neurips.cc/paper_files/paper/2019/file/26cd8ecadce0d4efd6cc8a8725cbd1f8-Paper.pdf},
    volume = {32},
    year = {2019}
}

@inproceedings{cranmer2020lagrangian,
    title={Lagrangian Neural Networks},
    author={Miles Cranmer and Sam Greydanus and Stephan Hoyer and Peter Battaglia and David Spergel and Shirley Ho},
    booktitle={ICLR 2020 Workshop on Integration of Deep Neural Models and Differential Equations},
    year={2019},
    url={https://openreview.net/forum?id=iE8tFa4Nq}
}

@article{udrescu2020aifeynman,
    author = {Silviu-Marian Udrescu  and Max Tegmark },
    title = {AI Feynman: A physics-inspired method for symbolic regression},
    journal = {Science Advances},
    volume = {6},
    number = {16},
    pages = {eaay2631},
    year = {2020},
    doi = {10.1126/sciadv.aay2631},
    URL = {https://www.science.org/doi/abs/10.1126/sciadv.aay2631},
    eprint = {https://www.science.org/doi/pdf/10.1126/sciadv.aay2631}}

@inproceedings{petersen2021deep,
    title={Deep symbolic regression: Recovering mathematical expressions from data via risk-seeking policy gradients},
    author={Brenden K Petersen and Mikel Landajuela Larma and Terrell N. Mundhenk and Claudio Prata Santiago and Soo Kyung Kim and Joanne Taery Kim},
    booktitle={International Conference on Learning Representations},
    year={2021},
    url={https://openreview.net/forum?id=m5Qsh0kBQG}
}

@inproceedings{oliveira2018analysing,
  title={Analysing symbolic regression benchmarks under a meta-learning approach},
  author={Oliveira, Luiz Otavio VB and Martins, Joao Francisco BS and Miranda, Luis F and Pappa, Gisele L},
  booktitle={Proceedings of the Genetic and Evolutionary Computation Conference Companion},
  pages={1342--1349},
  year={2018}
}

@inproceedings{mundhenk2021symbolic,
    author = {Mundhenk, Terrell and Landajuela, Mikel and Glatt, Ruben and Santiago, Claudio P and faissol, Daniel and Petersen, Brenden K},
    booktitle = {Advances in Neural Information Processing Systems},
    editor = {M. Ranzato and A. Beygelzimer and Y. Dauphin and P.S. Liang and J. Wortman Vaughan},
    pages = {24912--24923},
    publisher = {Curran Associates, Inc.},
    title = {Symbolic Regression via Deep Reinforcement Learning Enhanced Genetic Programming Seeding},
    url = {https://proceedings.neurips.cc/paper_files/paper/2021/file/d073bb8d0c47f317dd39de9c9f004e9d-Paper.pdf},
    volume = {34},
    year = {2021}
}

@inproceedings{xu2024reinforcement,
    author = {Xu, Yilong and Liu, Yang and Sun, Hao},
    booktitle = {International Conference on Learning Representations},
    editor = {B. Kim and Y. Yue and S. Chaudhuri and K. Fragkiadaki and M. Khan and Y. Sun},
    pages = {54416--54440},
    title = {Reinforcement Symbolic Regression Machine},
    url = {https://proceedings.iclr.cc/paper_files/paper/2024/file/ef3a55fa15aa5fe39b7a2617b3a5d06e-Paper-Conference.pdf},
    volume = {2024},
    year = {2024}
}

@InProceedings{biggio2021neural,
    title = 	 {Neural Symbolic Regression that scales},
    author =       {Biggio, Luca and Bendinelli, Tommaso and Neitz, Alexander and Lucchi, Aurelien and Parascandolo, Giambattista},
    booktitle = 	 {Proceedings of the 38th International Conference on Machine Learning},
    pages = 	 {936--945},
    year = 	 {2021},
    editor = 	 {Meila, Marina and Zhang, Tong},
    volume = 	 {139},
    series = 	 {Proceedings of Machine Learning Research},
    month = 	 {18--24 Jul},
    publisher =    {PMLR},
    url = 	 {https://proceedings.mlr.press/v139/biggio21a.html}
}

@inproceedings{kamienny2022end,
    author={Kamienny, Pierre-Alexandre and d'Ascoli, St{\'e}phane and Lample, Guillaume and Charton, Fran{\c{c}}ois},
    booktitle = {Advances in Neural Information Processing Systems},
    editor = {S. Koyejo and S. Mohamed and A. Agarwal and D. Belgrave and K. Cho and A. Oh},
    pages = {10269--10281},
    publisher = {Curran Associates, Inc.},
    title = {End-to-end Symbolic Regression with Transformers},
    url = {https://proceedings.neurips.cc/paper_files/paper/2022/file/42eb37cdbefd7abae0835f4b67548c39-Paper-Conference.pdf},
    volume = {35},
    year = {2022}
}

@misc{valipour2021symbolicgpt,
  title={SymbolicGPT: A Generative Transformer Model for Symbolic Regression}, 
  author={Mojtaba Valipour and Bowen You and Maysum Panju and Ali Ghodsi},
  year={2021},
  eprint={2106.14131},
  archivePrefix={arXiv},
  primaryClass={cs.LG},
  url={https://arxiv.org/abs/2106.14131}
}

@inproceedings{shojaee2023transformer,
    author = {Shojaee, Parshin and Meidani, Kazem and Barati Farimani, Amir and Reddy, Chandan},
    booktitle = {Advances in Neural Information Processing Systems},
    editor = {A. Oh and T. Naumann and A. Globerson and K. Saenko and M. Hardt and S. Levine},
    pages = {45907--45919},
    publisher = {Curran Associates, Inc.},
    title = {Transformer-based Planning for Symbolic Regression},
    url = {https://proceedings.neurips.cc/paper_files/paper/2023/file/8ffb4e3118280a66b192b6f06e0e2596-Paper-Conference.pdf},
    volume = {36},
    year = {2023}
}

@inproceedings{yu2024symbolic,
    author = {Yu, Zihan and Ding, Jingtao and Li, Yong and Jin, Depeng},
    booktitle = {International Conference on Learning Representations},
    editor = {Y. Yue and A. Garg and N. Peng and F. Sha and R. Yu},
    pages = {65306--65333},
    title = {Symbolic regression via MDLformer-guided search: from minimizing prediction error to minimizing description length},
    url = {https://proceedings.iclr.cc/paper_files/paper/2025/file/a402493de088886740b5939f666a6e56-Paper-Conference.pdf},
    volume = {2025},
    year = {2025}
}

@inproceedings{xiang2025graph,
    title={Graph-based Symbolic Regression with Invariance and Constraint Encoding},
    author={Ziyu Xiang and Kenna Ashen and Xiaofeng Qian and Xiaoning Qian},
    booktitle={The Thirty-ninth Annual Conference on Neural Information Processing Systems},
    year={2025},
    url={https://openreview.net/forum?id=JYB6wFcbky}
}

@inproceedings{huang2025improving,
    title={Improving Monte Carlo Tree Search for Symbolic Regression},
    author={Zhengyao Huang and Daniel Zhengyu Huang and Tiannan Xiao and Dina Ma and Zhenyu Ming and Hao Shi and Yuanhui Wen},
    booktitle={The Thirty-ninth Annual Conference on Neural Information Processing Systems},
    year={2025},
    url={https://openreview.net/forum?id=Wic0OgYsgy}
}

@inproceedings{meidani2024snip,
    title={{SNIP}: Bridging Mathematical Symbolic and Numeric Realms with Unified Pre-training},
    author={Kazem Meidani and Parshin Shojaee and Chandan K. Reddy and Amir Barati Farimani},
    booktitle={The Twelfth International Conference on Learning Representations},
    year={2024},
    url={https://openreview.net/forum?id=KZSEgJGPxu}
}

@article{tian2025interactive,
    author = {Tian, Yuan and Zhou, Wenqi and Viscione, Michele and Dong, Hao and Kammer, David S. and Fink, Olga},
    title = {Interactive symbolic regression with co-design mechanism through offline reinforcement learning},
    journal = {Nature Communications},
    year = {2025},
    volume = {16},
    number = {1},
    pages = {3930},
    doi = {10.1038/s41467-025-59288-y},
    url = {https://doi.org/10.1038/s41467-025-59288-y}
}

@article{gaodiscovering,
    title={Probabilistic grammars for modeling dynamical systems from coarse, noisy, and partial data},
    author={Gao, En-Hao and others},
    journal={Research Square},
    year={2023}
}

@article{omejc2023probabilistic,
    author={Omejc, Nina and Gec, Boštjan and Brence, Jure and others},
    title   = {Probabilistic grammars for modeling dynamical systems from coarse, noisy, and partial data},
    journal = {Machine Learning},
    year    = {2024},
    volume  = {113},
    number  = {10},
    pages   = {7689--7721},
    doi     = {10.1007/s10994-024-06522-1},
    url     = {https://doi.org/10.1007/s10994-024-06522-1}
}

@misc{lu2024ai,
    title={The AI Scientist: Towards Fully Automated Open-Ended Scientific Discovery}, 
    author={Chris Lu and Cong Lu and Robert Tjarko Lange and Jakob Foerster and Jeff Clune and David Ha},
    year={2024},
    eprint={2408.06292},
    archivePrefix={arXiv},
    primaryClass={cs.AI},
    url={https://arxiv.org/abs/2408.06292}
}

@inproceedings{czarnecki2017sobolev,
    author = {Czarnecki, Wojciech M. and Osindero, Simon and Jaderberg, Max and Swirszcz, Grzegorz and Pascanu, Razvan},
    booktitle = {Advances in Neural Information Processing Systems},
    editor = {I. Guyon and U. Von Luxburg and S. Bengio and H. Wallach and R. Fergus and S. Vishwanathan and R. Garnett},
    pages = {},
    publisher = {Curran Associates, Inc.},
    title = {Sobolev Training for Neural Networks},
    url = {https://proceedings.neurips.cc/paper_files/paper/2017/file/758a06618c69880a6cee5314ee42d52f-Paper.pdf},
    volume = {30},
    year = {2017}
}

@inproceedings{soroco2025pde,
    title={{PDE}-Controller: {LLM}s for Autoformalization and Reasoning of {PDE}s},
    author={Mauricio Soroco and Jialin Song and Mengzhou Xia and Kye Emond and Weiran Sun and Wuyang Chen},
    booktitle={Forty-second International Conference on Machine Learning},
    year={2025},
    url={https://openreview.net/forum?id=7epYTVsWEI}
}

@article{guo2025sr,
    author = {Zelin Guo  and Siqi Wang  and Yonglin Tian  and Jing Yang  and Hui Yu  and Xiaoxiang Na  and Levente Kovács  and Li Li  and Petros A. Ioannou  and Fei-Yue Wang },
    title = {SR-LLM: An incremental symbolic regression framework driven by LLM-based retrieval-augmented generation},
    journal = {Proceedings of the National Academy of Sciences},
    volume = {122},
    number = {52},
    pages = {e2516995122},
    year = {2025},
    doi = {10.1073/pnas.2516995122},
    URL = {https://www.pnas.org/doi/abs/10.1073/pnas.2516995122},
    eprint = {https://www.pnas.org/doi/pdf/10.1073/pnas.2516995122}
}

@article{park2026self,
      author  = {Park, Donggeun and Moon, Hyeonbin and Ryu, Seunghwa},
      title   = {A self-correcting multi-agent LLM framework for language-based physics simulation and explanation},
      journal = {npj Artificial Intelligence},
      year    = {2026},
      volume  = {2},
      number  = {1},
      pages   = {10},
      doi     = {10.1038/s44387-025-00057-z},
      url     = {https://doi.org/10.1038/s44387-025-00057-z},
      issn    = {3005-1460}
}

@misc{sun2025enumerate,
    title={Enumerate-Conjecture-Prove: Formally Solving Answer-Construction Problems in Math Competitions}, 
    author={Jialiang Sun and Yuzhi Tang and Ao Li and Chris J. Maddison and Kuldeep S. Meel},
    year={2025},
    eprint={2505.18492},
    archivePrefix={arXiv},
    primaryClass={cs.AI},
    url={https://arxiv.org/abs/2505.18492}
}

@article{linot2023stabilized,
    title = {Stabilized neural ordinary differential equations for long-time forecasting of dynamical systems},
    journal = {Journal of Computational Physics},
    volume = {474},
    pages = {111838},
    year = {2023},
    issn = {0021-9991},
    doi = {https://doi.org/10.1016/j.jcp.2022.111838},
    url = {https://www.sciencedirect.com/science/article/pii/S0021999122009019},
    author = {Alec J. Linot and Joshua W. Burby and Qi Tang and Prasanna Balaprakash and Michael D. Graham and Romit Maulik}
}

@article{team2023gemini,
    title={Gemini: a family of highly capable multimodal models},
    author={Team, Gemini and Anil, Rohan and Borgeaud, Sebastian and Alayrac, Jean-Baptiste and Yu, Jiahui and Soricut, Radu and Schalkwyk, Johan and Dai, Andrew M and Hauth, Anja and Millican, Katie and others},
    journal={arXiv preprint arXiv:2312.11805},
    year={2023}
}

@book{fletcher2013practical,
  title={Practical methods of optimization},
  author={Fletcher, Roger},
  year={2013},
  publisher={John Wiley \& Sons}
}

@article{lee2025finite,
  title={Finite Element Operator Network for Solving Elliptic-Type Parametric PDEs},
  author={Lee, Jae Yong and Ko, Seungchan and Hong, Youngjoon},
  journal={SIAM Journal on Scientific Computing},
  volume={47},
  number={2},
  year={2025}
}

@misc{cranmerInterpretableMachineLearning2023,
    title = {Interpretable {Machine} {Learning} for {Science} with {PySR} and {SymbolicRegression}.jl},
    url = {http://arxiv.org/abs/2305.01582},
    doi = {10.48550/arXiv.2305.01582},
    urldate = {2023-07-17},
    publisher = {arXiv},
    author = {Cranmer, Miles},
    month = may,
    year = {2023},
    note = {arXiv:2305.01582 [astro-ph, physics:physics]},
}

@inproceedings{zhang2012srbench,
  title={{SRBench}: {a streaming RDF/SPARQL benchmark}},
  author={Zhang, Ying and Duc, Pham Minh and Corcho, Oscar and Calbimonte, Jean-Paul},
  booktitle={International Semantic Web Conference},
  pages={641--657},
  year={2012},
  organization={Springer}
}

@inproceedings{matsubara2022srsd,
  title={{SRSD}: {Rethinking datasets of symbolic regression for scientific discovery}},
  author={Matsubara, Yoshitomo and Chiba, Naoya and Igarashi, Ryo and Ushiku, Yoshitaka},
  booktitle={NeurIPS 2022 AI for Science: Progress and Promises},
  year={2022}
}
\bibliographystyle{icml2026}

\newpage
\appendix
\onecolumn
\raggedbottom

\section{Dataset details}
\label{appendix:dataset}

\subsection{Benchmark problems and ground truth equations}
\label{app:benchmark_problems}

Table~\ref{tab:benchmark_specs_detailed} provides the complete specifications for all eight benchmark ODE systems used in our experiments. Problems ID 1-7 are sourced from ODEbench \citep{d2023odeformer}, while ID 8 is a 4D variant of the Glider problem, constructed to evaluate performance on four-dimensional systems with non-polynomial terms. For each benchmark problem, we list its identifier (ID 1-8), descriptive title, ground truth governing equations in mathematical form, and the temporal domains used for both ID and ID-Ext evaluation regimes. These specifications serve as the reference for comparing the equations discovered by DoLQ and baseline methods, as detailed in Table~\ref{tab:equations_appendix}.

\begin{table}[H]
  \centering
  \caption{Benchmark ODEs and their ground truth governing equations and temporal domains used for in-domain (ID) and extended in-domain (ID-Ext) evaluations.}
  \label{tab:benchmark_specs_detailed}
  \small
  \begin{tabular}{rm{4.5cm}m{5.5cm}cc}
    \toprule
    ID & Title & Ground Truth Equations & $t$ (ID) & $t$ (ID-Ext) \\
    \midrule
    1 & SIR infection model only for healthy \newline and sick & $\dot{x}_0 = -0.4 x_0 x_1$ \newline $\dot{x}_1 = 0.4 x_0 x_1 - 0.314 x_1$ & $[0, 2]$ & $[0, 4]$ \\
    \midrule
    2 & Glider (dimensionless) & $\dot{x}_0 = -x_0^2/5.0 - \sin(x_1)$ \newline $\dot{x}_1 = x_0 - \cos(x_1)/x_0$ & $[0, 5]$ & $[0, 10]$ \\
    \midrule
    3 & Reduced model for chlorine \newline dioxide-iodine-malonic acid \newline reaction (dimensionless) & $\dot{x}_0 = 8.9 - \frac{4.0 x_0 x_1}{x_0^2 + 1.0} - x_0$ \newline $\dot{x}_1 = 1.4 x_0 \left(1.0 - \frac{x_1}{x_0^2 + 1.0}\right)$ & $[0, 5]$ & $[0, 10]$ \\
    \midrule
    4 & Isothermal autocatalytic reaction \newline model by Gray and Scott 1985 (di- \newline mensionless) & $\dot{x}_0 = 0.5(1.0-x_0) - x_0 x_1^2$ \newline $\dot{x}_1 = -0.02 x_1 + x_0 x_1^2$ & $[0, 2]$ & $[0, 4]$ \\
    \midrule
    5 & Interacting bar magnets & $\dot{x}_0 = 0.33 \sin(x_0 - x_1) - \sin(x_0)$ \newline $\dot{x}_1 = -0.33 \sin(x_0 - x_1) - \sin(x_1)$ & $[0, 2]$ & $[0, 4]$ \\
    \midrule
    6 & Binocular rivalry model \newline (no oscillations) & $\dot{x}_0 = -x_0 + (e^{4.89 x_1 - 1.4} + 1.0)^{-1}$ \newline $\dot{x}_1 = -x_1 + (e^{4.89 x_0 - 1.4} + 1.0)^{-1}$ & $[0, 2]$ & $[0, 4]$ \\
    \midrule
    7 & Oscillator death model by \newline Ermentrout and Kopell (1990) & $\dot{x}_0 = 1.432 + \sin(x_1)\cos(x_0)$ \newline $\dot{x}_1 = 0.972 + \sin(x_1)\cos(x_0)$ & $[0, 4]$ & $[0, 8]$ \\
    \midrule
    8 & Glider (physical units) & $\dot{x}_0 = -9.81 \sin(x_1) - 0.030625 x_0^2$ \newline $\dot{x}_1 = -\frac{9.81 \cos(x_1)}{x_0} + 0.6125 x_0$ \newline $\dot{x}_2 = x_0 \cos(x_1)$ \newline $\dot{x}_3 = x_0 \sin(x_1)$ & $[0, 5]$ & $[0, 10]$ \\
    \bottomrule
  \end{tabular}
\end{table}

\subsection{Natural language descriptions for benchmark problems}
\label{app:natural_language_descriptions}

Table~\ref{tab:benchmark_descriptions} provides the natural language descriptions ($\mathcal{T}$) for each benchmark ODE system. These descriptions contain domain-specific context and physical principles that are provided as input to the LLM-based agents during the equation discovery process.

\begin{table}[H]
  \centering
  \caption{Natural language descriptions for benchmark problems.}
  \label{tab:benchmark_descriptions}
  \small
  \begin{tabular}{cp{12cm}}
    \toprule
    ID & Natural Language Description ($\mathcal{T}$) \\
    \midrule
    1 & This model describes disease spread by dividing the population into susceptible and infected groups, with transmission proportional to their contact and infected individuals recovering at a constant rate. An epidemic occurs only when the transmission rate exceeds a threshold, revealing a critical condition for outbreak or extinction \\
    \midrule
    2 & The glider is viewed as an idealized system whose motion is expected to arise from the interplay of gravity and aerodynamic forces, with speed and flight path angle evolving from assumed initial conditions \\
    \midrule
    3 & One state variable represents the activator species concentration, which autocatalytically accelerates the reaction and has limited spatial mobility due to polymer indicator binding. Another represents the inhibitor species concentration that suppresses activator production. The activator production rate shows saturation beyond a certain level due to substrate limitations. The consumption process is negligible at low activator concentrations but changes sharply above a specific range. External halide addition directly reacts with the inhibitor, reducing its amount. \\
    \midrule
    4 & This describes an open chemical system where a precursor is continuously supplied and an autocatalytic species both self-amplifies and decays. The interplay of local activation and global depletion leads to spontaneous pattern formation such as spots, waves, spirals, and chaotic spatial behavior arising from instability of an initially uniform state \\
    \midrule
    5 & This system describes two nearby bar magnets whose orientations influence each other through distance-dependent magnetic interactions. Depending on initial conditions and external effects, the magnets may settle into stable alignments, oscillate, rotate together, or follow complex motion patterns \\
    \midrule
    6 & This describes binocular rivalry, where two competing neural populations represent different images seen by each eye and suppress each other. Depending on initial conditions and input strength, one population becomes dominant, leading to stable perception of only one image at a time \\
    \midrule
    7 & This describes a counterintuitive effect where coupling two identical oscillating systems can completely suppress their oscillations. Strong symmetric interaction forces both systems into a shared steady state, causing oscillatory activity to disappear regardless of initial motion \\
    \midrule
    8 & This is conducted to measure the physical motion characteristics of a glider in actual flight. The experimenter sets the glider's travel speed, flight path angle relative to the horizontal plane, horizontal distance, and altitude as main measurement variables. Specifically, the state variables are defined as forward velocity, flight path angle, horizontal position, and vertical altitude. The experiment is based on an aircraft of specific mass moving through the atmosphere under constant gravitational acceleration, experiencing lift and drag determined by wing configuration and air density. The aircraft's mass, wing area, and atmospheric conditions are defined beforehand, and the flight trajectory is recorded by setting initial launch velocity and altitude. \\
    \bottomrule
  \end{tabular}
\end{table}


To reduce equation-level leakage, we removed equation names, explicit symbolic forms, and direct function-form clues from all descriptions before prompting. This design does not prove the absence of memorization in a strict causal sense, but it prevents the prompt from containing the answer in symbolic form and forces discovery to depend on consistency between the description, numerical optimization, and iterative term evaluation.

\subsection{Data generation procedure}
\label{app:data_generation}

We generate synthetic trajectory data for all benchmark systems using numerical integration. The data generation process consists of three main stages: ODE solving, derivative computation, and dataset partitioning.

\textbf{Numerical integration.} State trajectories are generated using the Runge--Kutta method. For each ODE system specified in Table~\ref{tab:benchmark_specs_detailed}, we integrate the ground truth equations over the temporal domains indicated in the table.

\textbf{Derivative computation.} We compute time derivatives $\dot{x}(t)$ using two complementary methods to validate numerical consistency. First, we evaluate the derivatives directly by substituting the interpolated state values into the ground truth ODE functions: $\dot{x}(t_i) = f(t_i, x(t_i))$, where $f$ represents the right-hand side of the ODE system. This provides the analytically correct derivative values denoted as $\dot{x}_{\text{gt}}$. Second, we compute numerical derivatives using NumPy's gradient function with second-order accurate edge handling (\texttt{edge\_order=2}), which applies the central difference scheme $\dot{x}(t_i) \approx (x(t_{i+1}) - x(t_{i-1}))/(2\Delta t)$ for interior points and forward/backward differences at boundaries. These numerically differentiated values, denoted as $\dot{x}_t$, serve as the target derivative values for residual-based optimization (Eq.~\ref{eq:residual} in the main text). For evaluation, we measure Residual NMSE by comparing discovered equations' predictions against $\dot{x}_{\text{gt}}$, ensuring consistency with the ground truth dynamics.

\textbf{Dataset partitioning.} Generated trajectories are partitioned into two distinct temporal regimes for model evaluation. The ID regime spans from $t_{\text{start}}$ to $t_{\text{end}}$ as specified in Table~\ref{tab:benchmark_specs_detailed}, representing the temporal range used for model training and in-distribution testing. The ID-Ext regime extends from $t_{\text{start}}$ to $t_{\text{ood\_end}}$, combining both the ID temporal range and an extended temporal domain for OOD evaluation. Specifically, for the extended portion beyond $t_{\text{end}}$, we use the state value at $t_{\text{end}}$ as the initial condition and continue the integration up to $t_{\text{ood\_end}}$, creating a continuous trajectory that tests the model's extrapolation capability over longer time horizons. All experiments use noise-free data ($\sigma = 0$) with 1000 uniformly sampled time points for each regime. Initial conditions for each system are predetermined and specified in the benchmark configuration rather than randomly sampled, ensuring reproducibility and consistency across all experimental runs.

\subsection{State trajectory visualization}
\label{app:trajectory_visualization}

\begin{figure}[H]
    \centering
    \includegraphics[width=1\linewidth]{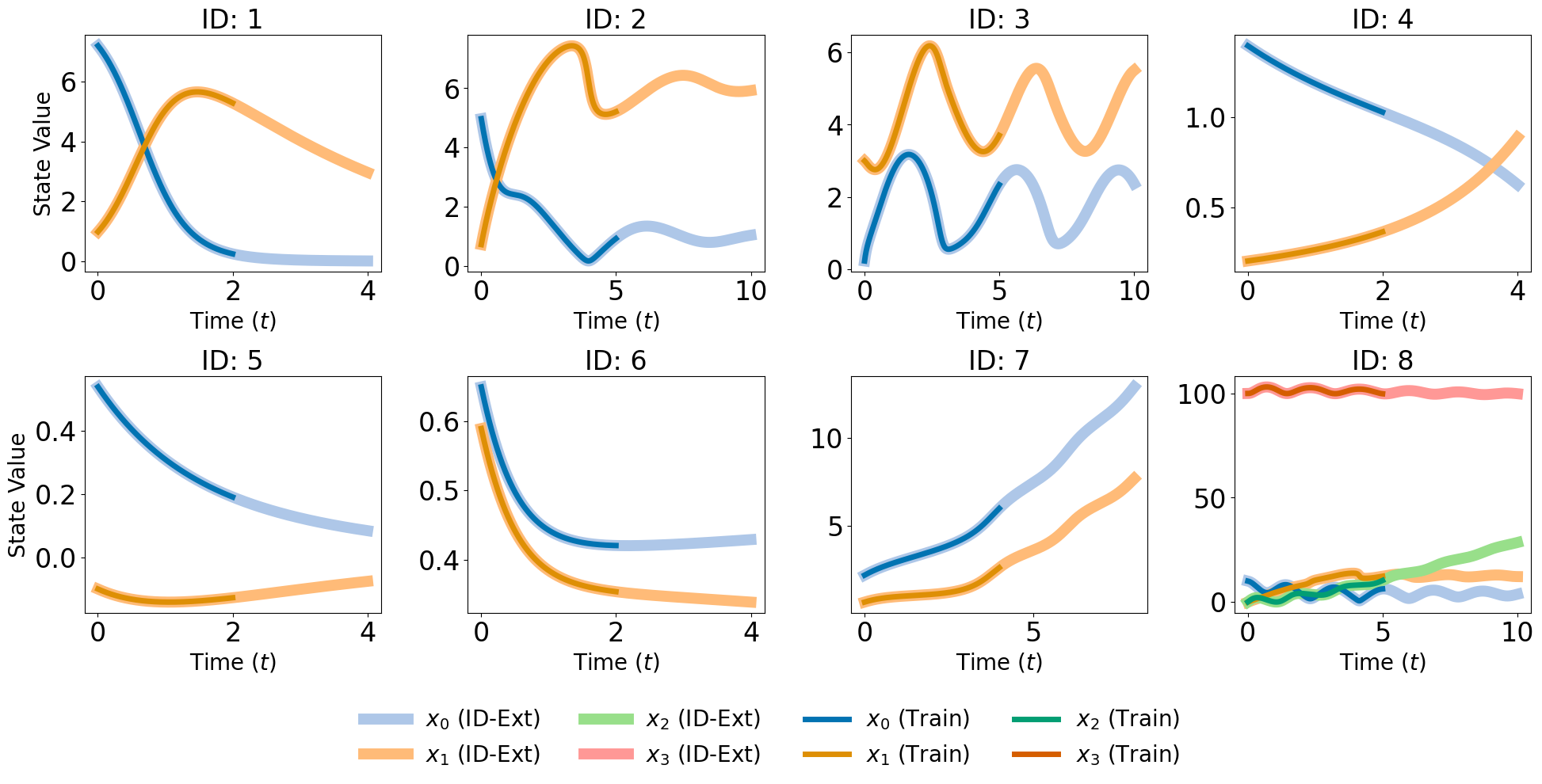}
    \caption{State trajectories for benchmark ODEs in ID and ID-Ext regimes. The figures display: ID 1: SIR infection model; ID 2: Glider (dimensionless); ID 3: Reduced model for chlorine dioxide-iodine-malonic acid reaction; ID 4: Isothermal autocatalytic reaction model; ID 5: Interacting bar magnets; ID 6: Binocular rivalry model; ID 7: Oscillator death model; ID 8: Glider (physical units).}
    \label{fig:all_ode_trajectories}
\end{figure}

We visualize the state trajectories for the benchmark ODE problems generated using the procedure described in the previous subsection. As shown in Figure~\ref{fig:all_ode_trajectories}, we present the temporal evolution of state variables for all eight benchmark problems (ID 1-8), where each subfigure displays both the ID training regime (solid lines) and the ID-Ext evaluation regime (extended temporal domain). The visualization demonstrates the diversity of dynamical behaviors in our benchmark suite, ranging from simple polynomial dynamics in the SIR model to complex nonlinear oscillations in systems like CDIMA and the Glider systems.

\vfill

\section{Implementation details}
\label{app:implementation_details}

\subsection{DoLQ hyperparameters}

The specific hyperparameters used for the DoLQ framework in our experiments are detailed in Table~\ref{tab:dolq_hyperparams}. These settings were chosen to balance exploration capability with computational efficiency.

\begin{table}[H]
    \centering
    \caption{Hyperparameters for DoLQ. Values are derived from the experimental configuration specific to the results reported.}
    \label{tab:dolq_hyperparams}
    \small
    \renewcommand{\arraystretch}{1.2}
    \resizebox{\columnwidth}{!}{%
    \begin{tabular}{l l l p{0.35\linewidth}}
        \toprule
        \textbf{Category} & \textbf{Parameter} & \textbf{Value} & \textbf{Description} \\
        \midrule
        \multirow{6}{*}{LLM Configuration} 
        & Sampler Model & \texttt{gemini-2.5-flash-lite} & Model for generating candidate terms \\
        & Scientist Model & \texttt{gemini-2.5-flash-lite} & Model for reasoning and evaluation \\
        & Sampler Temperature & $0.9$ & High temperature for diverse generation \\
        & Scientist Temperature & $0.6$ & Lower temperature for stable reasoning \\
        & Max Tokens (Sampler) & $3000$ & Constraint on generation length \\
        & Max Tokens (Scientist) & $3000$ & Constraint on evaluation length \\
        \midrule
        \multirow{5}{*}{Search \& Optimization} 
        & Evolution Iterations & $100$ & Total number of search cycles \\
        & Max Parameters & $8$ & Maximum trainable coefficients per term \\
        & Differential evolution tolerance & $10^{-5}$ & Convergence threshold for differential evolution \\
        & BFGS Tolerance & $10^{-9}$ & Convergence threshold for BFGS \\

        \midrule
        \multirow{4}{*}{System Settings} 
        & Differential evolution strategy & \texttt{best1bin} & Standard differential evolution strategy \\
        & Differential evolution population size & $20$ & Number of candidates in the differential evolution pool \\
        & Forget Probability & $0.01$ & Chance to revive a banned term \\
        & Timeout & $240$s & Maximum execution time per iteration \\
        \bottomrule
    \end{tabular}%
    }
    \vspace{0.5em}
\end{table}

In all experiments, the Sampler Agent produces 3 candidate hypotheses per iteration, and each equation in a candidate is constrained to contain at most 10 terms. We do not impose a separate sparsity regularizer; instead, sparsity is controlled operationally through this term cap together with the Scientist Agent's keep/hold/remove feedback and the subsequent ablation-based filtering. Outputs exceeding the term cap are discarded before optimization. If candidate evaluation produces overflow, underflow, division-by-zero, NaN, or $\infty$, we assign the corresponding objective value to $\infty$ and discard that candidate from further consideration in the current iteration.

\subsection{Experimental settings for baseline models}

To ensure a fair comparison with our proposed framework, DoLQ, we incorporated the system description $\mathcal{T}$ (providing domain-specific context and physical principles) into the prompts for each baseline method (L{\small{A}}SR, LLM-SR, EDL, ICSR), adapted to their respective prompt structures. All baseline methods utilize Gemini 2.5 Flash Lite~\citep{team2023gemini} as the underlying LLM. For experimental parameters not explicitly mentioned in the following tables, we followed the default settings of each respective baseline method. Specifically for ICSR, the binary and unary operators listed in Table~\ref{tab:icsr_settings} were explicitly provided in the prompt to define the permissible function space.

\begin{table}[H]
\centering
\begin{minipage}[t]{0.48\textwidth}
  \centering
  \captionof{table}{Experimental settings for L\small{A}SR.}
  \label{tab:lasr_settings}
  \small
  \resizebox{\linewidth}{!}{%
  \begin{tabular}{lcc}
    \toprule
    Setting                          & 2D                                                                              & 4D                   \\
    \midrule
    Description                      & Included                                                                        & Included             \\
    Iterations                       & 31                                                                              & 31                   \\
    Populations                      & 6                                                                               & 6                    \\
    Variable Names                   & $x_0, x_1$                                                                           & $x_0, x_1, x_2, x_3$ \\
    Binary Operators                 & \multicolumn{2}{c}{$+, -, *, /, \text{\^{}}$}                                                          \\
    \multirow{2}{*}{Unary Operators} & \multicolumn{2}{c}{$\cos, \sin, \log, \exp, \tan$,}                                                    \\
                                     & \multicolumn{2}{c}{$\text{abs}, \sinh, \cosh, \tanh, \text{sign}, \text{sqrt}$}                        \\
    Use LLM                          & \multicolumn{2}{c}{True}                                                                               \\
    Use Concepts/Evolution           & \multicolumn{2}{c}{True}                                                                               \\
    LLM Probabilities                & \multicolumn{2}{c}{0.02 (Crossover, Mutate, Randomize)}                                                \\
    \bottomrule
  \end{tabular}%
  }
\end{minipage}
\hfill
\begin{minipage}[t]{0.48\textwidth}
  \centering
  \captionof{table}{Experimental settings for LLM-SR.}
  \label{tab:llmsr_settings}
  \small
  \resizebox{\linewidth}{!}{%
  \begin{tabular}{lcc}
    \toprule
    Setting                        & 2D       & 4D       \\
    \midrule
    Description                    & Included & Included \\
    Max Samples (Iterations)       & 100      & 100       \\
    Number of Islands              & 3        & 3        \\
    Reset Period                   & 483      & 304      \\
    Cluster Sampling Temp.\ Period & 302      & 190      \\
    \bottomrule
  \end{tabular}%
  }
\end{minipage}

\vspace{1em}

\begin{minipage}[t]{0.48\textwidth}
  \centering
  \captionof{table}{Experimental settings for EDL.}
  \label{tab:edl_settings}
  \small
  \resizebox{\linewidth}{!}{%
  \begin{tabular}{lcc}
    \toprule
    Setting                          & 2D                                                                              & 4D                   \\
    \midrule
    Description                      & Included                                                                        & Included             \\
    Max Epoch (Iterations)           & 100                                                                             & 100                   \\
    Operands                         & $x_0, x_1$                                                                      & $x_0, x_1, x_2, x_3$ \\
    Binary Operators                 & \multicolumn{2}{c}{$+, -, *, /, \text{\^{}}$}                                                          \\
    \multirow{2}{*}{Unary Operators} & \multicolumn{2}{c}{$\cos, \sin, \log, \exp, \tan$,}                                                    \\
                                     & \multicolumn{2}{c}{$\text{abs}, \sinh, \cosh, \tanh, \text{sign}, \text{sqrt}$}                        \\
    Metric                           & \multicolumn{2}{c}{$R^2$}                                                                              \\
    Number of Samples ($M$)          & \multicolumn{2}{c}{10}                                                                               \\
    Optimize Type                    & \multicolumn{2}{c}{evolution\_optimize}                                                              \\
    Mode                             & \multicolumn{2}{c}{nonlinear}                                                                          \\
    \bottomrule
  \end{tabular}%
  }
\end{minipage}
\hfill
\begin{minipage}[t]{0.48\textwidth}
  \centering
  \captionof{table}{Experimental settings for ICSR.}
  \label{tab:icsr_settings}
  \small
  \resizebox{\linewidth}{!}{%
  \begin{tabular}{lcc}
    \toprule
    Setting                                   & 2D                                                                              & 4D                   \\
    \midrule
    Description                               & Included                                                                        & Included             \\
    Iterations                                & 100                                                                             & 100                   \\
    Operands                                  & $x_0, x_1$                                                                      & $x_0, x_1, x_2, x_3$ \\
    Binary Operators (Prompt)                 & \multicolumn{2}{c}{$+, -, *, /, \text{\^{}}$}                                                          \\
    \multirow{2}{*}{Unary Operators (Prompt)} & \multicolumn{2}{c}{$\cos, \sin, \log, \exp, \tan$,}                                                    \\
                                              & \multicolumn{2}{c}{$\text{abs}, \sinh, \cosh, \tanh, \text{sign}, \text{sqrt}$}                        \\
    \bottomrule
  \end{tabular}%
  }
\end{minipage}
\end{table}

The hyperparameter configurations for each baseline method are summarized in Tables~\ref{tab:lasr_settings}, \ref{tab:llmsr_settings}, \ref{tab:edl_settings}, and \ref{tab:icsr_settings}. Table~\ref{tab:lasr_settings} shows the L{\small{A}}SR settings, which employ a hybrid evolutionary approach with LLM-guided mutations at a probability of 0.02, running 31 iterations with 6 populations for both 2D and 4D systems. Table~\ref{tab:llmsr_settings} details the LLM-SR configuration, which utilizes an island-based evolutionary strategy with 3 islands and adaptive reset mechanisms (reset period of 483 for 2D and 304 for 4D). Table~\ref{tab:edl_settings} presents the EDL setup, configured with 100 epochs and using $R^2$ as the fitness metric with 10 samples per generation. Finally, Table~\ref{tab:icsr_settings} specifies the ICSR parameters, which performs 100 iterations of in-context symbolic regression with the operator set explicitly provided in the prompt.

\subsection{Algorithm for term retention strategy}
\label{app:term_retention}

This algorithm details the ``simple ablation test'' mentioned in Section~\ref{sec:quantitative_eval}, classifying terms based on their impact on residual MSE.

\begin{tcolorbox}[
    enhanced,
    title=Algorithm 1: Term Contribution Analysis via Ablation Test,
    colframe=black!70,
    colback=white!95!gray,
    coltitle=white,
    fonttitle=\bfseries\sffamily,
    boxrule=0.5pt,
    drop shadow=black!30!white
]
\ttfamily\small
\textbf{Input:} Equation $f(\mathbf{x}; \mathbf{\theta})$, Parameters $\mathbf{\theta} = \{c_1, \dots, c_k\}$, Data $\mathcal{D}$, Threshold $\delta=0.05$ \\
\textbf{Output:} Classification per term \textbf{(\textcolor{customgreen}{Good}/\textcolor{customred}{Bad}/\textcolor{customblue}{Neutral})}

\vspace{0.2cm}
1: \textbf{Calculate Baseline Error:} \\
\quad $MSE_{base} \leftarrow \text{Evaluate}(f(\mathbf{x}; \mathbf{\theta}), \mathcal{D})$

\vspace{0.1cm}
2: \textbf{For each term} $i \in \{1, \dots, k\}$: \\
\quad $c_{temp} \leftarrow \mathbf{\theta}.\text{copy}()$ \\
\quad $c_{temp}[i] \leftarrow 0$  \quad \textcolor{gray}{// Temporarily remove term $i$} \\
\quad \\
\quad $MSE_{ablated} \leftarrow \text{Evaluate}(f(\mathbf{x}; c_{temp}), \mathcal{D})$ \\
\quad \\
\quad $\Delta \leftarrow (MSE_{ablated} - MSE_{base}) / (MSE_{base} + \epsilon)$ \quad \textcolor{gray}{// Calculate Change Rate} \\
\quad \\
\quad \textbf{If} $\Delta > \delta$: \\
\quad \quad Classification $\leftarrow$ \textbf{\textcolor{customgreen}{Good}} (Positive Impact) \\
\quad \textbf{Else If} $\Delta < -\delta$: \\
\quad \quad Classification $\leftarrow$ \textbf{\textcolor{customred}{Bad}} (Negative Impact) \\
\quad \textbf{Else}: \\
\quad \quad Classification $\leftarrow$ \textbf{\textcolor{customblue}{Neutral}}

\vspace{0.1cm}
3: \textbf{Return} Classifications
\end{tcolorbox}\label{alg:simple_ablation_test}

\paragraph{Decision logic for term retention}
In the ablation test, we do not re-optimize the remaining coefficients after zeroing a term; instead, we reuse the previously optimized coefficient vector and set only the selected coefficient to zero. We adopted this approximation because a pilot comparison with full re-optimization changed only a very small number of keep/remove decisions while substantially increasing runtime. All optimization and ablation steps are performed on the given training trajectory rather than on a separate validation set.

The final decision for each term is determined by combining the qualitative evaluation from the Scientist LLM (Figure~\ref{fig:scientist_output_example}) and the quantitative evaluation from the Simple Ablation Test (Algorithm~\ref{alg:simple_ablation_test}). The decision matrix, including the accumulated hold mechanism, is presented in Table~\ref{tab:term_retention_logic}. Specifically, we enforce a ``2-strike'' rule where terms maintained in the \textit{Hold} state for two consecutive iterations are definitively removed (\textbf{Remove}) if they fail to improve.

\begin{table}[H]
    \centering
    \caption{Term Retention Decision Matrix. It summarizes the final action determination based on the dual evaluation results and the accumulated hold mechanism.}
    \label{tab:term_retention_logic}
    \small
    \renewcommand{\arraystretch}{1.3}
    \begin{tabular}{p{0.22\linewidth} p{0.22\linewidth} p{0.25\linewidth} p{0.18\linewidth}}
        \toprule
        \textbf{Semantic Quality} \newline (Scientist LLM) & \textbf{Quantitative Impact} \newline (Ablation Test) & \textbf{Previous State} \newline (Iteration $t-1$) & \textbf{Final Action} \newline (Iteration $t$) \\
        \midrule
        \textcolor{customred}{\textbf{Bad}} & Any & Any & \textbf{\textcolor{customred}{Remove}} \\
        \hline
        \textcolor{customgreen}{\textbf{Good}} & \textcolor{customgreen}{\textbf{Good}} (Positive) & Any & \textbf{\textcolor{customgreen}{Keep}} \\
        \hline
        \multirow{3}{*}{\shortstack[l]{\textcolor{customgreen}{\textbf{Good}}}} & \multirow{3}{*}{\shortstack[l]{\textbf{\textcolor{customblue}{Neutral} / \textcolor{customred}{Bad}}}} & None / Keep & \textbf{\textcolor{customblue}{Hold}} (1st) \\
         & & Hold (1st strike) & \textbf{\textcolor{customblue}{Hold}} (2nd) \\
         & & Hold (2nd strike) & \textbf{\textcolor{customred}{Remove}} \\
        \hline
        \multirow{3}{*}{\shortstack[l]{\textcolor{customblue}{\textbf{Neutral}}}} & \multirow{3}{*}{Any} & None / Keep & \textbf{\textcolor{customblue}{Hold}} (1st) \\
         & & Hold (1st strike) & \textbf{\textcolor{customblue}{Hold}} (2nd) \\
         & & Hold (2nd strike) & \textbf{\textcolor{customred}{Remove}} \\
        \bottomrule
    \end{tabular}
\end{table}

\subsection{Function construction}
\label{app:skeleton_conversion}

\begin{figure}[H]
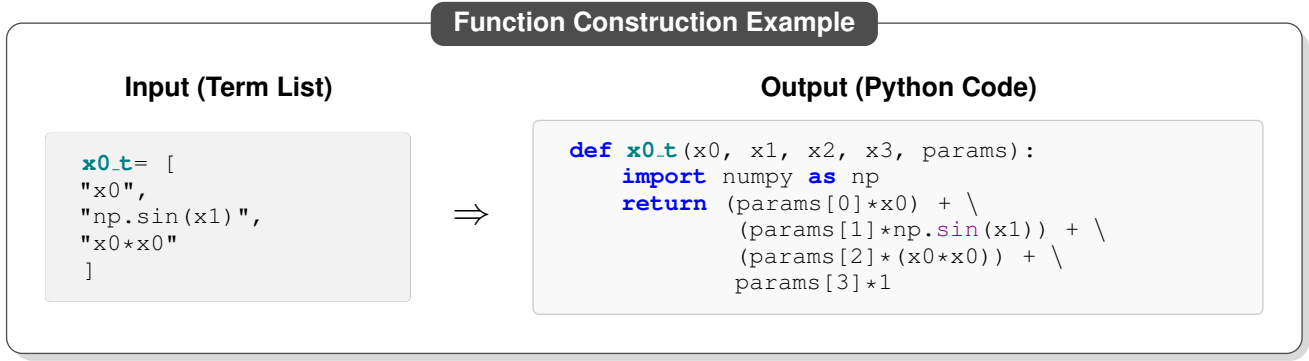

    \centering
    \begin{tcolorbox}[
        enhanced,
        title=Function Construction Example,
        colframe=black!70,
        colback=white,
        coltitle=white,
        colbacktitle=black!70,
        attach boxed title to top center={yshift=-10pt},
        boxed title style={
            frame hidden,
            rounded corners,
            arc=5pt,
        },
        fonttitle=\bfseries\sffamily,
        boxrule=0.5pt,
        drop shadow=black!30!white,
        left=4mm, right=4mm, top=6mm, bottom=4mm,
        arc=2mm
    ]
    \begin{minipage}[t]{0.30\linewidth}
        \centering
        \textbf{\textsf{Input (Term List)}}
    \end{minipage}
    \hfill
    \begin{minipage}[t]{0.05\linewidth}
        \centering
        ~
    \end{minipage}
    \hfill
    \begin{minipage}[t]{0.60\linewidth}
        \centering
        \textbf{\textsf{Output (Python Code)}}
    \end{minipage}
    
    \vspace{0.2cm}
    
    \begin{minipage}[c]{0.30\linewidth}
        \begin{tcolorbox}[colback=gray!10, boxrule=0pt, frame hidden, arc=2pt]
        \ttfamily\small
        \textbf{\textcolor{teal}{x0\_t}}= [\par
        "x0",\par
        "np.sin(x1)",\par
        "x0*x0"\par
        ]
        \end{tcolorbox}
    \end{minipage}
    \hfill
    \begin{minipage}[c]{0.05\linewidth}
        \centering
        \Large $\Rightarrow$
    \end{minipage}
    \hfill
    \begin{minipage}[c]{0.60\linewidth}
        \begin{tcolorbox}[colback=gray!5, colframe=gray!40, boxrule=0.5pt, arc=2pt, boxsep=2pt]
        \ttfamily\small
        \textbf{\textcolor{blue}{def}} \textbf{\textcolor{teal}{x0\_t}}(x0, x1, x2, x3, params):\\
        \hspace*{0.5cm} \textbf{\textcolor{blue}{import}} numpy \textbf{\textcolor{blue}{as}} np\\
        \hspace*{0.5cm} \textbf{\textcolor{blue}{return}} (params[0]*x0) + $\backslash$\\
        \hspace*{2.0cm} (params[1]*np.\textcolor{violet}{sin}(x1)) + $\backslash$\\
        \hspace*{2.0cm} (params[2]*(x0*x0)) + $\backslash$\\
        \hspace*{2.0cm} params[3]*1
        \end{tcolorbox}
    \end{minipage}
    \end{tcolorbox}
    \caption{Visual representation of the function construction process. A list of symbolic terms (left) is converted into an executable Python function (right) with learnable coefficients (\texttt{params}) and a bias term, enabling numerical optimization.}
    \label{fig:skeleton_conversion_example}
\end{figure}

As shown in Figure~\ref{fig:sampler_output_example}, the Sampler agent generates a list of candidate terms. To enable numerical optimization of the coefficients, we transform these symbolic terms into an executable Python function. Figure~\ref{fig:skeleton_conversion_example} illustrates this process, where each term is assigned a learnable parameter (e.g., \texttt{params[i]}) and combined into a formatted function body.

\section{Comprehensive experimental results}

\subsection{Discovered governing equations}
\label{app:discovered_equations}

In this section, we present the explicit mathematical forms of the governing equations discovered by each methodology for the benchmark ODE systems, detailing the structure and the optimized high-precision coefficients. These results are systematically organized in Table~\ref{tab:equations_appendix}, which provides a comprehensive comparison across all eight benchmark problems (ID 1-8 corresponding to the systems specified in Table~\ref{tab:benchmark_specs_detailed}). For each problem, the table lists the discovered equations from DoLQ and all baseline methods (EDL, ICSR, L{\small{A}}SR, LLM-SR), showing the complete equation structure with optimized coefficients for every dimension. This detailed presentation facilitates direct structural comparison between the equations recovered by our framework and those from baseline methods, enabling qualitative assessment of which methods successfully capture the ground truth terms versus those that propose spurious or incomplete dynamics.

\small
\renewcommand{\arraystretch}{1.4}
\begin{longtable}{ccc>{$}p{11.5cm}<{$}}
\caption{Governing equations discovered by each methodology.}
\label{tab:equations_appendix}\\
\toprule
ID & Method & Dim. & Discovered Equations \\
\midrule
\endfirsthead

\multicolumn{4}{c}{\tablename\ \thetable\ -- Continued from previous page} \\
\toprule
ID & Method & Dim. & Discovered Equations \\
\midrule
\endhead

\midrule
\multicolumn{4}{r}{Continued on next page...} \\
\endfoot

\bottomrule
\endlastfoot
  \multirow{14}{*}{1} & \multirow{2}{*}{DoLQ(ours)} & $\dot{x}_{0}$ & (-0.4000164923132323*(x0*x1)) + (1.1797068684934541e-06*(-x0*x1*x1)) + 0.00014926603696994873 * 1 \\
  \cline{3-4}
  \addlinespace[2pt]
   &  & $\dot{x}_{1}$ & (0.3999858850733014*(x0*x1)) + (0.3133017263306753*(-x1)) + (-0.00010236009691644724*(x1*x1)) - 0.000729280144169156 * 1 \\
  \cmidrule{2-4}
  \addlinespace[3pt]
   & \multirow{2}{*}{EDL} & $\dot{x}_{0}$ & -0.4*x0*x1 \\
  \cline{3-4}
  \addlinespace[2pt]
   &  & $\dot{x}_{1}$ & -0.1128*2*x0^2 + 0.0051*2*x1^3 + 0.0064*x0*x1 + 0.4826*x1^2 - 0.2024*x0*sqrt(x1) \\
  \cmidrule{2-4}
  \addlinespace[3pt]
   & \multirow{2}{*}{ICSR} & $\dot{x}_{0}$ & -0.4*x0*x1 + 0.0001 \\
  \cline{3-4}
  \addlinespace[2pt]
   &  & $\dot{x}_{1}$ & 0.4*x0*x1 - 0.314*x1 - 0.0002 \\
  \cmidrule{2-4}
  \addlinespace[3pt]
   & \multirow{2}{*}{LASR} & $\dot{x}_{0}$ & x1 * (x0 * -0.4000072832475115) \\
  \cline{3-4}
  \addlinespace[2pt]
   &  & $\dot{x}_{1}$ & x1 * ((x0 * 0.40001272551202366) + -0.314017558209762) \\
  \cmidrule{2-4}
  \addlinespace[3pt]
   & \multirow{2}{*}{LLM-SR} & $\dot{x}_{0}$ & (-0.5113286722182488 * (x0 * x1) / 1.2782811394075673) - (-1.6210813251309898e-05 * x1) \\
  \cline{3-4}
  \addlinespace[2pt]
   &  & $\dot{x}_{1}$ & (0.7940229825361336 / (1 + -0.0007829235554072554 * (x0 + x1))) * (1 / (1 + exp(-0.0023009370812258584 * ((x1 / (x0 + x1)) - -0.7319833222420157)))) * x0 * x1 - 0.31384689979086494 * x1 \\
  \midrule
  \multirow{14}{*}{2} & \multirow{2}{*}{DoLQ(ours)} & $\dot{x}_{0}$ & (-0.9999342734666715*(sin(x1))) + (-0.19999685243662685*(x0^2)) + (-3.159527567024422e-05*(x0*sin(x1))) + 8.291723665240451e-06 * 1 \\
  \cline{3-4}
  \addlinespace[2pt]
   &  & $\dot{x}_{1}$ & (-0.9996732126004018*(cos(x1)/x0)) + (-6.162357405925185e-05*(x0*sin(x1))) + (1.0002613154943063*(x0)) - 0.0006655465341315331 * 1 \\
  \cmidrule{2-4}
  \addlinespace[3pt]
   & \multirow{2}{*}{EDL} & $\dot{x}_{0}$ & -0.2*sin(x1) - 1.0*x0^2 \\
  \cline{3-4}
  \addlinespace[2pt]
   &  & $\dot{x}_{1}$ & 1.2906*x1*log(abs(x0)) + 3.7156*x1 + 0.4329*x0^2*x1 - 1.2452*x0 + 1.55*x1*cos(x0) \\
  \cmidrule{2-4}
  \addlinespace[3pt]
   & \multirow{2}{*}{ICSR} & $\dot{x}_{0}$ & -0.2*x0^2 - 1.0*sin(1.0*x1) - 0.0001 \\
  \cline{3-4}
  \addlinespace[2pt]
   &  & $\dot{x}_{1}$ & -0.5662/(x0^{1.894}*x1^{0.5626}) + 0.8685*x0 - 0.1773*x1 + 1.0748 \\
  \cmidrule{2-4}
  \addlinespace[3pt]
   & \multirow{2}{*}{LASR} & $\dot{x}_{0}$ & (x0 / -1.7671421572704085) - sin(x1) \\
  \cline{3-4}
  \addlinespace[2pt]
   &  & $\dot{x}_{1}$ & x0 - (cos(x1) / x0) \\
  \cmidrule{2-4}
  \addlinespace[3pt]
   & \multirow{2}{*}{LLM-SR} & $\dot{x}_{0}$ & (-1.387691981374681 * 0.9997683841467897 * sin(x1) + (-0.27884783934870216 * square(x0) - (-0.00023524831049328578) * power(x0, 3)) + ((-0.0005339077881395173) * x0 * cos(x1) + (-0.0009606900297937713) * x0 * sin(x1)) + ((1.1025607057532369e-05) * x0 * square(x1) + (0.0001392070749430701) * square(x0) * x1 + (1.0871856110837581e-05) * x0 * square(cos(x1)) + (0.0003566962804791643) * square(x0) * sin(x1))) / 1.387691981374681 \\
  \cline{3-4}
  \addlinespace[2pt]
   &  & $\dot{x}_{1}$ & ((1.00016068611199 * x0^2) + (-0.0003203621386719613 * x0) + (-0.00026883955217234126) - 0.9994481859297301 * cos(x1)) / maximum(x0, 1e-6) \\
  \midrule
  \multirow{14}{*}{3} & \multirow{2}{*}{DoLQ(ours)} & $\dot{x}_{0}$ & (-0.999924337295147*(x0)) + (0.0013713847096808362*(x0/(1.0 + x0^2))) + (4.0000051777750265*(-x1*x0/(1.0 + x0^2))) + 8.899331099033029 * 1 \\
  \cline{3-4}
  \addlinespace[2pt]
   &  & $\dot{x}_{1}$ & (1.3999732914208531*(-x0 * x1 / (1.0 + x0^2))) + (-4.4947665548779136e-05*(-x0^2 / (1.0 + x0^2))) + (-1.3999589124979535*(-x0)) + 4.897208257542713e-07 * 1 \\
  \cmidrule{2-4}
  \addlinespace[3pt]
   & \multirow{2}{*}{EDL} & $\dot{x}_{0}$ & -7.159*x1^3 + 10.3164*x0 - 4.0771*cos(x1)^2 - 0.3057*sqrt(x0) + 0.0662*x1 \\
  \cline{3-4}
  \addlinespace[2pt]
   &  & $\dot{x}_{1}$ & -0.7501*sqrt(x0) + 0.2166*x1^2 + 10.8516*x0^2 - 9.7161*x1 - 0.3873*cos(x0)^3 \\
  \cmidrule{2-4}
  \addlinespace[3pt]
   & \multirow{2}{*}{ICSR} & $\dot{x}_{0}$ & -4.0*x0*x1/(x0^2 + 0.9999) - 1.0002*x0*tanh(2.0678*x1) + 8.9004 \\
  \cline{3-4}
  \addlinespace[2pt]
   &  & $\dot{x}_{1}$ & -17.0218*x0*tanh(2.5653*x0) + 18.7118*x0 - 3.2262*x1/(x0 + 3.8981) - 0.6359 \\
  \cmidrule{2-4}
  \addlinespace[3pt]
   & \multirow{2}{*}{LASR} & $\dot{x}_{0}$ & ((sin(x0) + -0.7225424108770416) * -1.3716248441728514) + tan(sinh(sin(x1 + -1.271652676453637))) \\
  \cline{3-4}
  \addlinespace[2pt]
   &  & $\dot{x}_{1}$ & log(x0) \\
  \cmidrule{2-4}
  \addlinespace[3pt]
   & \multirow{2}{*}{LLM-SR} & $\dot{x}_{0}$ & (7.2413552436477575 * (x0 / (-0.09692412820488534 + x0))) * (1 - (x1^{1.6726750850832244}) / (4.267112440332994^{1.6726750850832244} + x1^{1.6726750850832244})) + (-1.6934497454181765 * x0 * x1 * ((x0^{-2.2251470718338093}) / (1.2897199381103897^{-2.2251470718338093} + x0^{-2.2251470718338093}))) \\
  \cline{3-4}
  \addlinespace[2pt]
   &  & $\dot{x}_{1}$ & (0.15305571645674615 * (x0 / (1.4598732465999258 + x0))) * (1 - 17.709388951838832 * x1) - (-2.096703476296624 * ((x0^{1.9666706388834827}) / (2.087291118909436^{1.9666706388834827} + x0^{1.9666706388834827})) * x1) - (-1.3593840857406783 * x0) \\
  \midrule
  \multirow{14}{*}{4} & \multirow{2}{*}{DoLQ(ours)} & $\dot{x}_{0}$ & (0.4973058292805835*(-x0)) + (0.21746562709380776*(-x0*x1)) + (0.6344054886410723*(-x1*x1)) + 0.5264411613474004 * 1 \\
  \cline{3-4}
  \addlinespace[2pt]
   &  & $\dot{x}_{1}$ & (0.41858642320557227*(x0*x1^3)) + (-1.6764582281044875*(x1/(1+x1))) + (1.5117151384013812*(tanh(x1))) + 0.028354488818915925 * 1 \\
  \cmidrule{2-4}
  \addlinespace[3pt]
   & \multirow{2}{*}{EDL} & $\dot{x}_{0}$ & -0.1566*sin(x1)^2 - 1.404*x0^2 + 0.5305*x1 \\
  \cline{3-4}
  \addlinespace[2pt]
   &  & $\dot{x}_{1}$ & 0.8039*x0*x1 + 0.0724*sin(x1)^2 \\
  \cmidrule{2-4}
  \addlinespace[3pt]
   & \multirow{2}{*}{ICSR} & $\dot{x}_{0}$ & 2.4649*x0*cos(0.9101*x1) - 2.9665*x0 + 0.5033 \\
  \cline{3-4}
  \addlinespace[2pt]
   &  & $\dot{x}_{1}$ & 0.0692*x0 + 0.6177*x1 - 0.1679 \\
  \cmidrule{2-4}
  \addlinespace[3pt]
   & \multirow{2}{*}{LASR} & $\dot{x}_{0}$ & abs(sinh(tanh(sin(tan(x1 + ((x0^{0.22334073977903218}) + (x0 / x0))) + x0))) + (x0 * 0.3514105571067361)) * -0.35812758021702057 \\
  \cline{3-4}
  \addlinespace[2pt]
   &  & $\dot{x}_{1}$ & ((x0 * x1) - 0.019991627704281027) * x1 \\
  \cmidrule{2-4}
  \addlinespace[3pt]
   & \multirow{2}{*}{LLM-SR} & $\dot{x}_{0}$ & (0.7575404568207408 * (x0^{1.0100603887529287}) * (x1^{0.6955603195414775})) + (-0.6206768205830506 * x0) + 0.39548215725015795 + (-1.6012562268622113 * x1) + (0.5863974471340245 * (x1^{0.698948044530037})) \\
  \cline{3-4}
  \addlinespace[2pt]
   &  & $\dot{x}_{1}$ & (0.007339212327735855 * x0) + (0.9624797409788882 * x1^2) + (0.7873542337819555 * x1^3) - (-0.06037485229609352 * x1) - (-0.06037485265191514 * x1) - (0.09930729898681726 * x0 * x1) - (1.2126452269944556 * x1^3) + (0.0073391823827276045 * x0) \\
  \midrule
  \multirow{14}{*}{5} & \multirow{2}{*}{DoLQ(ours)} & $\dot{x}_{0}$ & (-1.0015488038824854*(sin(x0))) + (0.33102674079192346*(sin(x0 - x1))) + (-0.0018115741826902861*(cos(x0))) + 0.0017490917848575025 * 1 \\
  \cline{3-4}
  \addlinespace[2pt]
   &  & $\dot{x}_{1}$ & (-0.9981457993630254*(sin(x1))) + (0.3300067495528232*(sin(x1 - x0))) + (-0.012578074617818194*(cos(x1))) + 0.012716069769122665 * 1 \\
  \cmidrule{2-4}
  \addlinespace[3pt]
   & \multirow{2}{*}{EDL} & $\dot{x}_{0}$ & -0.3375*sin(x0) - 0.684*x1 \\
  \cline{3-4}
  \addlinespace[2pt]
   &  & $\dot{x}_{1}$ & -0.2973*x1 - 0.6309*x0 \\
  \cmidrule{2-4}
  \addlinespace[3pt]
   & \multirow{2}{*}{ICSR} & $\dot{x}_{0}$ & 5.0117*x0*sin(x1) - 2.2897*x1 - 0.2639 \\
  \cline{3-4}
  \addlinespace[2pt]
   &  & $\dot{x}_{1}$ & -0.2981*x0 - 0.6191*x1 + 0.0019 \\
  \cmidrule{2-4}
  \addlinespace[3pt]
   & \multirow{2}{*}{LASR} & $\dot{x}_{0}$ & (x0 * -0.66306757779156) + 0.04301494511373006 \\
  \cline{3-4}
  \addlinespace[2pt]
   &  & $\dot{x}_{1}$ & x0 * ((x0 / -1.192434344729412) + 0.25749851031674253) \\
  \cmidrule{2-4}
  \addlinespace[3pt]
   & \multirow{2}{*}{LLM-SR} & $\dot{x}_{0}$ & -1.000043264361751 * sin(x0) + -0.3300718799861972 * sin(x1 - x0) + -1.568587192043395e-05 * cos(x1 + x0) \\
  \cline{3-4}
  \addlinespace[2pt]
   &  & $\dot{x}_{1}$ & -0.9938291044117465 * x1 + -0.3292639827583344 * sin(x0 - x1) - (-2.656564954764913e-06 / (x0^2 + x1^2 + 1e-9)^{1.5}) + -0.0003613964745632571 * x0 * sin(x1) \\
  \midrule
  \multirow{14}{*}{6} & \multirow{2}{*}{DoLQ(ours)} & $\dot{x}_{0}$ & (1.3286098112316618*(-x0)) + (1.178235675777727*(-x1)) + (-0.6010789351853646*(-power(x0, 3))) + 0.9285456156774677 * 1 \\
  \cline{3-4}
  \addlinespace[2pt]
   &  & $\dot{x}_{1}$ & (0.23753405260404614*(-x1)) + (-0.7898637222390873*(-x0)) + (-0.8360904268130745*(tanh(x1))) + (2.290634305077196*(-tanh(x0))) + 0.9338877404981648 * 1 \\
  \cmidrule{2-4}
  \addlinespace[3pt]
   & \multirow{2}{*}{EDL} & $\dot{x}_{0}$ & -1.1878*x1 - 0.467*log(sin(x0)) \\
  \cline{3-4}
  \addlinespace[2pt]
   &  & $\dot{x}_{1}$ & -1.1299*log(sin(x0)) - 0.4331*x1 \\
  \cmidrule{2-4}
  \addlinespace[3pt]
   & \multirow{2}{*}{ICSR} & $\dot{x}_{0}$ & -1.8485*x1 - 0.1948*sin(x0) + 0.7327 \\
  \cline{3-4}
  \addlinespace[2pt]
   &  & $\dot{x}_{1}$ & 0.2393*x0 - 2.1268*x1 + 0.6431 \\
  \cmidrule{2-4}
  \addlinespace[3pt]
   & \multirow{2}{*}{LASR} & $\dot{x}_{0}$ & cos(exp(tan(x0))) \\
  \cline{3-4}
  \addlinespace[2pt]
   &  & $\dot{x}_{1}$ & tanh(0.14631324378767832 / x1) - x0 \\
  \cmidrule{2-4}
  \addlinespace[3pt]
   & \multirow{2}{*}{LLM-SR} & $\dot{x}_{0}$ & ((-2.249593853293885 * (x0 / (1.1957556897109163 + x0)) + 1.0667454050960814) - (3.6741828960616694 * (x1 / (2.3342324749372967 + x1)))) \\
  \cline{3-4}
  \addlinespace[2pt]
   &  & $\dot{x}_{1}$ & x1 * (-2.9587039457791064 * (1 - (x1 / 0.5623115553288086)) - (1.966298296973735 * x0 * 1.966298237331697)) + 0.9508865699097846 \\
  \midrule
  \multirow{14}{*}{7} & \multirow{2}{*}{DoLQ(ours)} & $\dot{x}_{0}$ & (-1.1655257292555141*(sin(x1-x0))) + (1.22475958770386*(cos(x0))) + (-0.1090013166633977*(x0*sin(x0))) + 0.817067905629052 * 1 \\
  \cline{3-4}
  \addlinespace[2pt]
   &  & $\dot{x}_{1}$ & (-2.2117156170501504*(x1)) + (-1.291170751585586*(x0)) + (-4.676614135451172*(cos(x1))) + (1.8750516943971047*(sin(x1-x0))) + 10.469917369126234 * 1 \\
  \cmidrule{2-4}
  \addlinespace[3pt]
   & \multirow{2}{*}{EDL} & $\dot{x}_{0}$ & -0.817*x1^2 - 4.3577*x0^3*sin(x0) + 2.7509*x1 + 1.5327*x0*sqrt(x1) + 0.0552*x0*cos(x0) \\
  \cline{3-4}
  \addlinespace[2pt]
   &  & $\dot{x}_{1}$ & -1.7872*x0 + 0.3915*sqrt(x1) - 0.7313*cos(x0) + 0.1632*x0^2*x1 + 0.1673*x1^3 \\
  \cmidrule{2-4}
  \addlinespace[3pt]
   & \multirow{2}{*}{ICSR} & $\dot{x}_{0}$ & -0.2806*x0 + 0.6592*x1*cos(1.1398*x0) + 2.0662 \\
  \cline{3-4}
  \addlinespace[2pt]
   &  & $\dot{x}_{1}$ & -0.2806*x0 + 0.6592*x1*cos(1.1398*x0) + 1.6062 \\
  \cmidrule{2-4}
  \addlinespace[3pt]
   & \multirow{2}{*}{LASR} & $\dot{x}_{0}$ & (cos(x0) * sin(x1)) + 1.4320121866725906 \\
  \cline{3-4}
  \addlinespace[2pt]
   &  & $\dot{x}_{1}$ & (cos(x0) * sin(x1)) + 0.9720121867249 \\
  \cmidrule{2-4}
  \addlinespace[3pt]
   & \multirow{2}{*}{LLM-SR} & $\dot{x}_{0}$ & (-(-1.231629475981958) * x0 - 2.158066362365913 * x0^2 - (-0.3037486836573964) * x0^3) + (1.3169135770056555 * x1 + 2.287561908522607 * x1^2 + -1.2108364457957292 * x1^3) + (-0.8950111650180933 * (x1 - x0) + 1.609022512581006 * (x1 - x0)^2 + 0.3125475185298518 * (x1 - x0)^3) \\
  \cline{3-4}
  \addlinespace[2pt]
   &  & $\dot{x}_{1}$ & (0.14605034843290507 * x1) + (0.3954492324348049 * x0) + (1.249712176091898 * (x0 - x1)) + (-0.2869847813397451 * x1^3) + (0.03189090914482745 * x0^3) + (-0.08031202424973191 * x1^2) + (-0.5242674943108351 * x0^2) + (0.017802263147943736 * x0^4) + (-0.15007455215067977 * x1^4) + (-0.09580817521497335 * (x0 - x1)^3) \\
  \midrule
  \multirow{24}{*}{8} & \multirow{4}{*}{DoLQ(ours)} & $\dot{x}_{0}$ & (-9.809127338404517*(sin(x1))) + (-9.411027230248375e-05*(cos(x1))) + (-0.030620460686854464*(x0*x0)) - 4.9118855130869645e-05 * 1 \\
  \cline{3-4}
  \addlinespace[2pt]
   &  & $\dot{x}_{1}$ & (0.9990536032782628*(-9.81/x0*cos(x1))) + (0.6130165352306062*(x0)) + (-0.0032944410806144153*(cos(x1))) - 0.0023916253783730504 * 1 \\
  \cline{3-4}
  \addlinespace[2pt]
   &  & $\dot{x}_{2}$ & (1.0000423141055985*(x0*cos(x1))) + (-1.5764831178241946e-05*(x0*x0*cos(x1))) + 0.00014446800549605261 * 1 \\
  \cline{3-4}
  \addlinespace[2pt]
   &  & $\dot{x}_{3}$ & (1.0000402047694594*(x0*sin(x1))) + (-2.003183912960774e-05*(x0*x0*sin(x1))) + (-6.303122694043895e-06*(9.81*cos(x1))) - 4.353532868248261e-05 * 1 \\
  \cmidrule{2-4}
  \addlinespace[3pt]
   & \multirow{4}{*}{EDL} & $\dot{x}_{0}$ & -0.0306*sin(x1) - 9.8091*x0^2 \\
  \cline{3-4}
  \addlinespace[2pt]
   &  & $\dot{x}_{1}$ & -0.9504*x0^3 - 0.2996*x1^2 - 0.0006*x3 + 1.3296*sqrt(abs(x1)) - 175.3583*x0/2 \\
  \cline{3-4}
  \addlinespace[2pt]
   &  & $\dot{x}_{2}$ & 0.9999*x0*cos(x1) \\
  \cline{3-4}
  \addlinespace[2pt]
   &  & $\dot{x}_{3}$ & 0.9999*x0*sin(x1) \\
  \cmidrule{2-4}
  \addlinespace[3pt]
   & \multirow{4}{*}{ICSR} & $\dot{x}_{0}$ & -0.2813*x0 + 0.0015*x1*x2 - 0.0968*x3*sin(1.0007*x1) + 0.4387 \\
  \cline{3-4}
  \addlinespace[2pt]
   &  & $\dot{x}_{1}$ & -0.0605*x0 - 6.8291*x1*exp(-2.7884*x0) - 0.0042*x2*x3 + 5.306 \\
  \cline{3-4}
  \addlinespace[2pt]
   &  & $\dot{x}_{2}$ & 0.9998*x0*cos(x1) + 0.0217 \\
  \cline{3-4}
  \addlinespace[2pt]
   &  & $\dot{x}_{3}$ & 0.9999*x0*sin(x1) + 0.0002 \\
  \cmidrule{2-4}
  \addlinespace[3pt]
   & \multirow{4}{*}{LASR} & $\dot{x}_{0}$ & sin(x1) * -9.813480254368727 \\
  \cline{3-4}
  \addlinespace[2pt]
   &  & $\dot{x}_{1}$ & ((log(x0) - ((x2 - 5.963031610926088) * 0.24925925160994267)) * 8.24239810331568) / x0 \\
  \cline{3-4}
  \addlinespace[2pt]
   &  & $\dot{x}_{2}$ & cos(x1) * x0 \\
  \cline{3-4}
  \addlinespace[2pt]
   &  & $\dot{x}_{3}$ & sin(x1) * x0 \\
  \cmidrule{2-4}
  \addlinespace[3pt]
   & \multirow{4}{*}{LLM-SR} & $\dot{x}_{0}$ & (-9.807899351331447 * sin(x1)) + -((0.4145037567137417 + 1.5751662511703768 * (-0.0001796563617646803 + -8.696424882717339e-05 * x1)^2) * 0.07394440250282637 * x0^2) + ((-0.0001796563617646803 + -8.696424882717339e-05 * x1) * 0.07394440250282637 * x0^2 * sin(x1)) + (- -0.00026688377456063225 * x0) + (3.1382118431267775e-05 * x2) + (-7.49115792450938e-06 * x3) + (0.00024089007421140555 * x0 * cos(x1) * sin(x1)) \\
  \cline{3-4}
  \addlinespace[2pt]
   &  & $\dot{x}_{1}$ & ((0.4312345753897334 * x0) + (-7.190875450891451 * x0^2) + (115.22337211688654 * cos(x1))) / (-11.719230243958577 * x0 + -0.023905720212081497 + 1e-9) + (0.0004499464651382155 * x2) + (0.0002873146658744879 * x3) + (3.203979516768474e-05 * x0 * x1) \\
  \cline{3-4}
  \addlinespace[2pt]
   &  & $\dot{x}_{2}$ & (-(-8.313688998942643e-05) * sin(x1)) + (0.00012926385963329574 * -0.0002110727122192323 * x3 * x0^2 * sin(x1)) + (-0.00012926385963329574 * 0.05 * x0^2 * cos(x1)) + (-0.00012926385963329574 * 1.5157577205870516 * (-0.0002110727122192323 * x3)^2 * x0^2 * cos(x1)) + (-1.0 * exp(-((x3 - 1.0) / 1.0)^2) * x0^2 * cos(x1)) + (- -0.9999727429727798 * (x0 - -4.820794712947023e-05) * cos(x1)) + (1.7829345690512744e-08 * x3 * x1 * x0) \\
  \cline{3-4}
  \addlinespace[2pt]
   &  & $\dot{x}_{3}$ & (0.00013369546648802163 * x0 * cos(x1)) + (-6.586738720708613e-06 * x0^2 * sin(x1)) + (-4.30440238524879e-06 * x0^3 * cos(x1)) + (-2.158359661264577e-05 * x2 * x0 * cos(x1)) + (0.00011240049424943671 * x0) + (-0.0001459955779485213 * x0 * sin(x1)^2) + (1.2061925694746197e-05 * x0^2 * cos(x1)) + (-3.152838841683395e-05 * x0 * sin(x1) * cos(x1)) + 2.7038601280307466e-05 + (0.9999723594741682 * x0 * sin(x1)) \\
\end{longtable}
\normalsize

\subsection{Quantitative results}
\label{app:quantitative_results}

To quantitatively assess the fidelity of the discovered models, we employ the Normalized Mean Squared Error (NMSE) metric, evaluating the deviation between the ground truth dynamics and the identified governing equations for each state variable dimension. Table~\ref{tab:nmse_appendix} presents these NMSE values across all benchmark problems (ID 1-8) for both DoLQ and baseline methods. The table is structured to show four evaluation metrics for each dimension: ID NMSE (Residual) and ID NMSE (Integral) measure performance on the training domain, while ID-Ext NMSE (Residual) and ID-Ext NMSE (Integral) assess generalization to extended temporal regimes. Residual NMSE evaluates point-wise derivative matching, whereas Integral NMSE measures trajectory prediction accuracy through numerical integration from initial conditions. This comprehensive quantitative comparison enables systematic analysis of both fitting quality and extrapolation capability across different problem complexities and functional forms.

Let $\hat{\dot{x}}_{j,i}$ denote the derivative predicted by the discovered equation at time index $i$, and let $\tilde{x}_{j,i}$ denote the trajectory obtained by numerically integrating the discovered ODE from the initial condition. The two scale-normalized metrics are defined as follows:
\[
\mathrm{Residual\ NMSE}_j
=
\frac{\sum_i(\dot{x}_{j,i}-\hat{\dot{x}}_{j,i})^2}{\sum_i \dot{x}_{j,i}^2+\epsilon}
\]
\[
\mathrm{Integral\ NMSE}_j
=
\frac{\sum_i(x_{j,i}-\tilde{x}_{j,i})^2}{\sum_i x_{j,i}^2+\epsilon}
\]
Residual MSE in Eq.~\eqref{eq:residual} is used inside the search loop for parameter optimization and term ablation, whereas residual and integral NMSE are used only for final scale-normalized evaluation.

\scriptsize
\begin{longtable}{rllcccccccc}
\caption{NMSE comparison for each dimension.}
\label{tab:nmse_appendix}\\
\toprule
\multirow{2}{*}{ID} & \multirow{2}{*}{Method} & \multirow{2}{*}{Dim} & \multicolumn{2}{c}{ID NMSE (Residual)} & \multicolumn{2}{c}{ID NMSE (Integral)} & \multicolumn{2}{c}{ID-Ext NMSE (Residual)} & \multicolumn{2}{c}{ID-Ext NMSE (Integral)} \\
\cmidrule(lr){4-5} \cmidrule(lr){6-7} \cmidrule(lr){8-9} \cmidrule(lr){10-11}
 &  &  & Value & & Value & & Value & & Value & \\
\midrule
\endfirsthead

\multicolumn{11}{c}{\tablename\ \thetable\ -- Continued from previous page} \\
\toprule
\multirow{2}{*}{ID} & \multirow{2}{*}{Method} & \multirow{2}{*}{Dim} & \multicolumn{2}{c}{ID NMSE (Residual)} & \multicolumn{2}{c}{ID NMSE (Integral)} & \multicolumn{2}{c}{ID-Ext NMSE (Residual)} & \multicolumn{2}{c}{ID-Ext NMSE (Integral)} \\
\cmidrule(lr){4-5} \cmidrule(lr){6-7} \cmidrule(lr){8-9} \cmidrule(lr){10-11}
 &  &  & Value & & Value & & Value & & Value & \\
\midrule
\endhead

\midrule
\multicolumn{11}{r}{Continued on next page...} \\
\endfoot

\bottomrule
\endlastfoot
  \multirow{14}{*}{1} & \multirow{2}{*}{DoLQ(ours)} & $x_{0}$ & $2.13 \times 10^{-8}$ & & $3.45 \times 10^{-9}$ & & $1.04 \times 10^{-8}$ & & $2.36 \times 10^{-9}$ & \\
   &  & $x_{1}$ & $1.28 \times 10^{-8}$ & & $4.52 \times 10^{-9}$ & & $2.07 \times 10^{-8}$ & & $5.03 \times 10^{-8}$ & \\
  \cmidrule{2-11}
   & \multirow{2}{*}{EDL} & $x_{0}$ & $2.39 \times 10^{-8}$ & & --- & & $9.16 \times 10^{-9}$ & & --- & \\
   &  & $x_{1}$ & $3.71 \times 10^{1}$ & & --- & & $2.80 \times 10^{1}$ & & --- & \\
  \cmidrule{2-11}
   & \multirow{2}{*}{ICSR} & $x_{0}$ & $2.92 \times 10^{-8}$ & & $1.89 \times 10^{-9}$ & & $1.22 \times 10^{-8}$ & & $1.32 \times 10^{-9}$ & \\
   &  & $x_{1}$ & $2.69 \times 10^{-8}$ & & $7.36 \times 10^{-9}$ & & $1.62 \times 10^{-8}$ & & $2.24 \times 10^{-8}$ & \\
  \cmidrule{2-11}
   & \multirow{2}{*}{LASR} & $x_{0}$ & $2.22 \times 10^{-8}$ & & $5.62 \times 10^{-9}$ & & $8.56 \times 10^{-9}$ & & $3.42 \times 10^{-9}$ & \\
   &  & $x_{1}$ & $1.34 \times 10^{-8}$ & & $8.77 \times 10^{-9}$ & & $6.64 \times 10^{-9}$ & & $1.38 \times 10^{-8}$ & \\
  \cmidrule{2-11}
   & \multirow{2}{*}{LLM-SR} & $x_{0}$ & $2.12 \times 10^{-8}$ & & $3.70 \times 10^{-9}$ & & $8.59 \times 10^{-9}$ & & $2.21 \times 10^{-9}$ & \\
   &  & $x_{1}$ & $1.19 \times 10^{-8}$ & & $4.58 \times 10^{-9}$ & & $2.37 \times 10^{-8}$ & & $5.36 \times 10^{-8}$ & \\
  \midrule
  \multirow{14}{*}{2} & \multirow{2}{*}{DoLQ(ours)} & $x_{0}$ & $4.93 \times 10^{-9}$ & & $1.42 \times 10^{-8}$ & & $5.83 \times 10^{-9}$ & & $2.19 \times 10^{-8}$ & \\
   &  & $x_{1}$ & $3.51 \times 10^{-8}$ & & $4.69 \times 10^{-8}$ & & $3.64 \times 10^{-8}$ & & $5.06 \times 10^{-8}$ & \\
  \cmidrule{2-11}
   & \multirow{2}{*}{EDL} & $x_{0}$ & $1.43 \times 10^{1}$ & & $1.02 \times 10^{0}$ & & $1.19 \times 10^{1}$ & & $1.43 \times 10^{0}$ & \\
   &  & $x_{1}$ & $1.64 \times 10^{2}$ & & --- & & $3.19 \times 10^{2}$ & & --- & \\
  \cmidrule{2-11}
   & \multirow{2}{*}{ICSR} & $x_{0}$ & $1.21 \times 10^{-8}$ & & 0.000110 & & $1.51 \times 10^{-8}$ & & 0.0618 & \\
   &  & $x_{1}$ & 0.005294 & & 0.000763 & & 0.0772 & & 0.0222 & \\
  \cmidrule{2-11}
   & \multirow{2}{*}{LASR} & $x_{0}$ & 0.1127 & & 0.0653 & & 0.1478 & & 0.0721 & \\
   &  & $x_{1}$ & $5.61 \times 10^{-8}$ & & 0.1116 & & $5.33 \times 10^{-8}$ & & 0.1431 & \\
  \cmidrule{2-11}
   & \multirow{2}{*}{LLM-SR} & $x_{0}$ & $1.41 \times 10^{0}$ & & $9.32 \times 10^{0}$ & & $1.15 \times 10^{0}$ & & $1.80 \times 10^{1}$ & \\
   &  & $x_{1}$ & $1.19 \times 10^{0}$ & & $8.20 \times 10^{0}$ & & $1.12 \times 10^{0}$ & & $1.59 \times 10^{1}$ & \\
  \midrule
  \multirow{14}{*}{3} & \multirow{2}{*}{DoLQ(ours)} & $x_{0}$ & $4.22 \times 10^{-8}$ & & $3.79 \times 10^{-8}$ & & $3.65 \times 10^{-8}$ & & $1.16 \times 10^{-7}$ & \\
   &  & $x_{1}$ & $3.52 \times 10^{-9}$ & & $1.58 \times 10^{-8}$ & & $1.06 \times 10^{-8}$ & & $1.14 \times 10^{-7}$ & \\
  \cmidrule{2-11}
   & \multirow{2}{*}{EDL} & $x_{0}$ & $2.14 \times 10^{5}$ & & --- & & $2.65 \times 10^{5}$ & & --- & \\
   &  & $x_{1}$ & $4.31 \times 10^{2}$ & & --- & & $4.15 \times 10^{2}$ & & --- & \\
  \cmidrule{2-11}
   & \multirow{2}{*}{ICSR} & $x_{0}$ & $5.11 \times 10^{-8}$ & & $4.03 \times 10^{-5}$ & & $5.11 \times 10^{-8}$ & & 0.000210 & \\
   &  & $x_{1}$ & 0.001031 & & $8.09 \times 10^{-5}$ & & 0.000756 & & 0.000303 & \\
  \cmidrule{2-11}
   & \multirow{2}{*}{LASR} & $x_{0}$ & 0.0688 & & $1.66 \times 10^{1}$ & & 0.0758 & & $3.08 \times 10^{1}$ & \\
   &  & $x_{1}$ & 0.5129 & & $6.14 \times 10^{0}$ & & 0.5047 & & $6.62 \times 10^{1}$ & \\
  \cmidrule{2-11}
   & \multirow{2}{*}{LLM-SR} & $x_{0}$ & 0.005255 & & 0.004934 & & 0.006184 & & 0.007153 & \\
   &  & $x_{1}$ & $8.93 \times 10^{-5}$ & & 0.005235 & & $9.19 \times 10^{-5}$ & & 0.0116 & \\
  \midrule
  \multirow{14}{*}{4} & \multirow{2}{*}{DoLQ(ours)} & $x_{0}$ & $1.92 \times 10^{-6}$ & & $3.27 \times 10^{-9}$ & & 0.2675 & & 0.000975 & \\
   &  & $x_{1}$ & $9.95 \times 10^{-8}$ & & $1.71 \times 10^{-9}$ & & 0.001386 & & $5.48 \times 10^{-5}$ & \\
  \cmidrule{2-11}
   & \multirow{2}{*}{EDL} & $x_{0}$ & $3.05 \times 10^{3}$ & & $2.94 \times 10^{1}$ & & $1.03 \times 10^{3}$ & & $5.83 \times 10^{0}$ & \\
   &  & $x_{1}$ & $7.11 \times 10^{1}$ & & $9.47 \times 10^{0}$ & & $1.92 \times 10^{0}$ & & $2.35 \times 10^{0}$ & \\
  \cmidrule{2-11}
   & \multirow{2}{*}{ICSR} & $x_{0}$ & $7.54 \times 10^{-6}$ & & $1.35 \times 10^{-6}$ & & 0.0120 & & 0.001100 & \\
   &  & $x_{1}$ & 0.000154 & & $1.88 \times 10^{-5}$ & & 0.0267 & & 0.007869 & \\
  \cmidrule{2-11}
   & \multirow{2}{*}{LASR} & $x_{0}$ & 0.000159 & & $6.88 \times 10^{-7}$ & & $1.05 \times 10^{0}$ & & 0.007504 & \\
   &  & $x_{1}$ & $1.19 \times 10^{-7}$ & & $2.29 \times 10^{-8}$ & & $6.71 \times 10^{-8}$ & & 0.000871 & \\
  \cmidrule{2-11}
   & \multirow{2}{*}{LLM-SR} & $x_{0}$ & $6.08 \times 10^{-6}$ & & $2.45 \times 10^{-8}$ & & 0.5790 & & 0.004555 & \\
   &  & $x_{1}$ & $7.52 \times 10^{-7}$ & & $2.13 \times 10^{-8}$ & & 0.007890 & & 0.000784 & \\
  \midrule
  \multirow{14}{*}{5} & \multirow{2}{*}{DoLQ(ours)} & $x_{0}$ & $2.43 \times 10^{-7}$ & & $5.61 \times 10^{-10}$ & & $1.27 \times 10^{-7}$ & & $2.87 \times 10^{-9}$ & \\
   &  & $x_{1}$ & $4.85 \times 10^{-7}$ & & $6.86 \times 10^{-8}$ & & $4.56 \times 10^{-7}$ & & $1.97 \times 10^{-7}$ & \\
  \cmidrule{2-11}
   & \multirow{2}{*}{EDL} & $x_{0}$ & $5.86 \times 10^{0}$ & & $7.16 \times 10^{0}$ & & $2.68 \times 10^{0}$ & & $2.61 \times 10^{1}$ & \\
   &  & $x_{1}$ & $2.08 \times 10^{1}$ & & $6.26 \times 10^{2}$ & & $1.56 \times 10^{1}$ & & $8.86 \times 10^{2}$ & \\
  \cmidrule{2-11}
   & \multirow{2}{*}{ICSR} & $x_{0}$ & 0.002610 & & $4.30 \times 10^{-5}$ & & 0.2313 & & 0.0435 & \\
   &  & $x_{1}$ & $2.58 \times 10^{-6}$ & & $1.89 \times 10^{-5}$ & & $5.54 \times 10^{-5}$ & & 0.0310 & \\
  \cmidrule{2-11}
   & \multirow{2}{*}{LASR} & $x_{0}$ & 0.000346 & & $1.33 \times 10^{-5}$ & & 0.0117 & & 0.001680 & \\
   &  & $x_{1}$ & 0.009038 & & 0.0114 & & 0.0339 & & 0.0892 & \\
  \cmidrule{2-11}
   & \multirow{2}{*}{LLM-SR} & $x_{0}$ & $2.48 \times 10^{-7}$ & & $6.19 \times 10^{-10}$ & & $1.17 \times 10^{-7}$ & & $4.77 \times 10^{-8}$ & \\
   &  & $x_{1}$ & $4.91 \times 10^{-7}$ & & $6.40 \times 10^{-8}$ & & 0.000219 & & 0.000103 & \\
  \midrule
  \multirow{14}{*}{6} & \multirow{2}{*}{DoLQ(ours)} & $x_{0}$ & $1.10 \times 10^{-6}$ & & $3.28 \times 10^{-8}$ & & $2.33 \times 10^{-6}$ & & $7.31 \times 10^{-8}$ & \\
   &  & $x_{1}$ & $1.07 \times 10^{-6}$ & & $3.71 \times 10^{-8}$ & & $7.50 \times 10^{-6}$ & & $4.55 \times 10^{-6}$ & \\
  \cmidrule{2-11}
   & \multirow{2}{*}{EDL} & $x_{0}$ & $1.11 \times 10^{-5}$ & & --- & & $1.19 \times 10^{-5}$ & & --- & \\
   &  & $x_{1}$ & $5.01 \times 10^{1}$ & & --- & & $7.15 \times 10^{1}$ & & --- & \\
  \cmidrule{2-11}
   & \multirow{2}{*}{ICSR} & $x_{0}$ & 0.000298 & & $1.76 \times 10^{-5}$ & & 0.004846 & & $7.17 \times 10^{-5}$ & \\
   &  & $x_{1}$ & 0.000426 & & $3.68 \times 10^{-5}$ & & 0.0170 & & 0.006141 & \\
  \cmidrule{2-11}
   & \multirow{2}{*}{LASR} & $x_{0}$ & 0.0366 & & 0.0161 & & 0.0281 & & 0.0158 & \\
   &  & $x_{1}$ & 0.0185 & & 0.0180 & & 0.0262 & & 0.0620 & \\
  \cmidrule{2-11}
   & \multirow{2}{*}{LLM-SR} & $x_{0}$ & $7.03 \times 10^{-6}$ & & $6.29 \times 10^{-7}$ & & $5.25 \times 10^{-6}$ & & $7.12 \times 10^{-5}$ & \\
   &  & $x_{1}$ & $2.47 \times 10^{-6}$ & & $4.72 \times 10^{-7}$ & & 0.000166 & & 0.000228 & \\
  \midrule
  \multirow{14}{*}{7} & \multirow{2}{*}{DoLQ(ours)} & $x_{0}$ & 0.000467 & & $3.62 \times 10^{-5}$ & & $6.53 \times 10^{0}$ & & 0.3966 & \\
   &  & $x_{1}$ & 0.000239 & & 0.000125 & & $3.70 \times 10^{2}$ & & 0.5569 & \\
  \cmidrule{2-11}
   & \multirow{2}{*}{EDL} & $x_{0}$ & $3.30 \times 10^{5}$ & & --- & & $1.56 \times 10^{7}$ & & --- & \\
   &  & $x_{1}$ & $3.93 \times 10^{1}$ & & --- & & $2.10 \times 10^{4}$ & & --- & \\
  \cmidrule{2-11}
   & \multirow{2}{*}{ICSR} & $x_{0}$ & 0.000510 & & $2.55 \times 10^{-5}$ & & $1.30 \times 10^{1}$ & & 0.4600 & \\
   &  & $x_{1}$ & 0.000510 & & 0.000116 & & $1.30 \times 10^{1}$ & & $1.01 \times 10^{0}$ & \\
  \cmidrule{2-11}
   & \multirow{2}{*}{LASR} & $x_{0}$ & $1.20 \times 10^{-7}$ & & $3.14 \times 10^{-8}$ & & $4.15 \times 10^{-7}$ & & $1.79 \times 10^{-8}$ & \\
   &  & $x_{1}$ & $1.20 \times 10^{-7}$ & & $1.43 \times 10^{-7}$ & & $4.15 \times 10^{-7}$ & & $3.93 \times 10^{-8}$ & \\
  \cmidrule{2-11}
   & \multirow{2}{*}{LLM-SR} & $x_{0}$ & $9.64 \times 10^{-6}$ & & $8.79 \times 10^{-8}$ & & $1.67 \times 10^{3}$ & & --- & \\
   &  & $x_{1}$ & $9.78 \times 10^{-6}$ & & $3.77 \times 10^{-7}$ & & $5.84 \times 10^{3}$ & & --- & \\
  \midrule
  \multirow{24}{*}{8} & \multirow{4}{*}{DoLQ(ours)} & $x_{0}$ & $7.23 \times 10^{-8}$ & & $1.10 \times 10^{-6}$ & & $6.82 \times 10^{-8}$ & & $3.94 \times 10^{-6}$ & \\
   &  & $x_{1}$ & $4.78 \times 10^{-6}$ & & $2.01 \times 10^{-7}$ & & $3.20 \times 10^{-6}$ & & $3.38 \times 10^{-7}$ & \\
   &  & $x_{2}$ & $1.23 \times 10^{-8}$ & & $3.40 \times 10^{-6}$ & & $1.68 \times 10^{-8}$ & & $6.47 \times 10^{-6}$ & \\
   &  & $x_{3}$ & $9.38 \times 10^{-9}$ & & $1.43 \times 10^{-6}$ & & $3.13 \times 10^{-8}$ & & $1.01 \times 10^{-5}$ & \\
  \cmidrule{2-11}
   & \multirow{4}{*}{EDL} & $x_{0}$ & $2.69 \times 10^{3}$ & & --- & & $2.60 \times 10^{3}$ & & --- & \\
   &  & $x_{1}$ & $5.67 \times 10^{4}$ & & --- & & $5.20 \times 10^{4}$ & & --- & \\
   &  & $x_{2}$ & $1.61 \times 10^{-8}$ & & --- & & $2.91 \times 10^{-8}$ & & --- & \\
   &  & $x_{3}$ & $1.08 \times 10^{-8}$ & & --- & & $3.22 \times 10^{-8}$ & & --- & \\
  \cmidrule{2-11}
   & \multirow{4}{*}{ICSR} & $x_{0}$ & 0.000678 & & $1.01 \times 10^{0}$ & & 0.000945 & & $2.85 \times 10^{0}$ & \\
   &  & $x_{1}$ & 0.1014 & & $1.14 \times 10^{0}$ & & $1.02 \times 10^{0}$ & & $2.74 \times 10^{0}$ & \\
   &  & $x_{2}$ & $3.72 \times 10^{-5}$ & & 0.1605 & & $6.03 \times 10^{-5}$ & & 0.4885 & \\
   &  & $x_{3}$ & $1.52 \times 10^{-8}$ & & $2.32 \times 10^{0}$ & & $3.65 \times 10^{-8}$ & & $6.67 \times 10^{0}$ & \\
  \cmidrule{2-11}
   & \multirow{4}{*}{LASR} & $x_{0}$ & 0.0264 & & $4.24 \times 10^{0}$ & & 0.0255 & & $1.64 \times 10^{2}$ & \\
   &  & $x_{1}$ & 0.0735 & & $1.31 \times 10^{0}$ & & $1.95 \times 10^{0}$ & & $3.39 \times 10^{0}$ & \\
   &  & $x_{2}$ & $1.84 \times 10^{-8}$ & & $1.84 \times 10^{0}$ & & $2.05 \times 10^{-8}$ & & 0.8254 & \\
   &  & $x_{3}$ & $1.51 \times 10^{-8}$ & & $6.70 \times 10^{0}$ & & $3.40 \times 10^{-8}$ & & $3.85 \times 10^{3}$ & \\
  \cmidrule{2-11}
   & \multirow{4}{*}{LLM-SR} & $x_{0}$ & $6.89 \times 10^{-8}$ & & $7.58 \times 10^{-7}$ & & $7.08 \times 10^{-8}$ & & $1.01 \times 10^{-5}$ & \\
   &  & $x_{1}$ & $3.89 \times 10^{-6}$ & & $1.92 \times 10^{-7}$ & & $3.24 \times 10^{-6}$ & & $5.18 \times 10^{-7}$ & \\
   &  & $x_{2}$ & $1.29 \times 10^{-8}$ & & $2.81 \times 10^{-6}$ & & $1.71 \times 10^{-8}$ & & $2.65 \times 10^{-6}$ & \\
   &  & $x_{3}$ & $9.91 \times 10^{-9}$ & & $7.05 \times 10^{-7}$ & & $8.81 \times 10^{-8}$ & & $5.76 \times 10^{-6}$ & \\
\end{longtable}
\normalsize

\subsection{Scores}
\label{app:scores}

In this section, we summarize the comprehensive evaluation scores for each benchmark problem. Figure~\ref{fig:success_scores} and Table~\ref{tab:detailed_scores} present a binary success/failure assessment across all eight problems (P1-P8, corresponding to benchmark IDs 1-8 from Table~\ref{tab:benchmark_specs_detailed}) for each method. The evaluation is based on two distinct criteria: the NMSE test row indicates whether the normalized mean squared error satisfies the designated convergence threshold (integral NMSE $< 10^{-3}$ across all dimensions), and the Term test row verifies whether the discovered equation accurately recovers the correct symbolic terms of the ground truth dynamics after excluding terms with negligible impact. A blue checkmark (\bluecheck) denotes success for that problem, while an empty cell indicates failure. The rightmost Total column sums the number of successful problems for each method and evaluation criterion, providing an aggregate view of discovery performance across the benchmark suite.

As prior ODE-discovery baselines do not report structural recovery under a unified false-positive/false-negative (FP/FN) protocol, precision/recall metrics are omitted from the main comparison. Instead, the Term test evaluates whether the recovered equation matches the ground-truth term set after excluding negligible terms, which serves as a strict structure-recovery criterion on benchmarks with known governing equations.

\begin{table}[H]
\centering
\caption{Detailed scores for each problem by evaluation type and method.}
\label{tab:detailed_scores}
\begin{center}
\begin{small}
\begin{sc}
\begin{tabular}{llcccccccc|c}
\toprule
Evaluation & Method & P1 & P2 & P3 & P4 & P5 & P6 & P7 & P8 & Total \\
\midrule
NMSE test & DoLQ & \bluecheck & \bluecheck & \bluecheck & \bluecheck & \bluecheck & \bluecheck &  & \bluecheck & 7 \\
NMSE test & EDL &  &  &  &  &  &  &  &  & 0 \\
NMSE test & ICSR & \bluecheck &  & \bluecheck &  &  &  &  &  & 2 \\
NMSE test & LASR & \bluecheck &  &  &  &  &  & \bluecheck &  & 2 \\
NMSE test & LLM-SR & \bluecheck &  &  &  & \bluecheck & \bluecheck &  & \bluecheck & 4 \\
\midrule
Term test & DoLQ & \bluecheck & \bluecheck & \bluecheck &  & \bluecheck &  &  & \bluecheck & 5 \\
Term test & EDL &  &  &  &  &  &  &  &  & 0 \\
Term test & ICSR &  &  &  &  &  &  &  &  & 0 \\
Term test & LASR & \bluecheck &  &  &  &  &  & \bluecheck &  & 2 \\
Term test & LLM-SR &  & \bluecheck &  &  & \bluecheck &  &  &  & 2 \\
\bottomrule
\end{tabular}
\end{sc}
\end{small}
\end{center}
\vskip -0.1in
\end{table}

\subsection{Token usage details}
\label{app:token_usage}

This section presents a detailed comparison of token usage between the proposed framework and baseline models. The token consumption data in the following table is reported in thousands (K) of tokens, distinguishing between input tokens (In) and output tokens (Out). For each benchmark problem (ID 1-8), we track two metrics: ``Found'' columns show the cumulative tokens consumed up to the point when the method successfully discovered a reasonable equation (if applicable), while ``Total'' columns report the tokens used throughout the entire experimental run until completion or timeout. This distinction enables analysis of both the efficiency of successful discovery and the overall computational cost. All values are reported directly from the experiment logs collected during our experiments. Note that for DoLQ, the reported token counts represent the combined usage of both the Sampler Agent and the Scientist Agent, and no reasoning tokens were generated during our experiments.

\begin{table}[H]
  \centering
  \caption{Token usage until discovery.}
  \footnotesize
  \setlength{\tabcolsep}{3pt}

  \begin{minipage}[t]{0.49\linewidth}
  \centering
  \begin{tabular}{rlcccc}
  \toprule
  \multirow{2}{*}{ID} & \multirow{2}{*}{Method} & \multicolumn{2}{c}{Found (K)} & \multicolumn{2}{c}{Total (K)} \\
  \cmidrule(lr){3-4} \cmidrule(lr){5-6}
   &  & In & Out & In & Out \\
  \midrule
  \multirow{5}{*}{1} & DoLQ(ours) & 20.2 & 10.1 & 226 & 96.8 \\
   & EDL & 4.99 & 6.96 & 551 & 439 \\
   & ICSR & 243 & 24.8 & 243 & 24.8 \\
   & LASR & 357 & 387 & 357 & 387 \\
   & LLM-SR & 338 & 443 & 338 & 443 \\
  \midrule
  \multirow{5}{*}{2} & DoLQ(ours) & 69.6 & 40.0 & 289 & 156 \\
   & EDL & 149 & 42.5 & 465 & 118 \\
   & ICSR & 262 & 39.0 & 262 & 39.0 \\
   & LASR & 426 & 426 & 426 & 426 \\
   & LLM-SR & 533 & 602 & 703 & 816 \\
  \midrule
  \multirow{5}{*}{3} & DoLQ(ours) & 188 & 110 & 282 & 166 \\
   & EDL & 129 & 302 & 296 & 296 \\
   & ICSR & 309 & 44.1 & 309 & 44.1 \\
   & LASR & 483 & 447 & 483 & 447 \\
   & LLM-SR & 652 & 692 & 813 & 813 \\
  \midrule
  \multirow{5}{*}{4} & DoLQ(ours) & 44.6 & 26.4 & 201 & 109 \\
   & EDL & 1.49 & 2.97 & 193 & 239 \\
   & ICSR & 260 & 35.1 & 260 & 35.1 \\
   & LASR & 401 & 404 & 401 & 404 \\
   & LLM-SR & 781 & 775 & 781 & 775 \\
  \bottomrule
  \end{tabular}
  \end{minipage}
  \hfill
  \begin{minipage}[t]{0.49\linewidth}
  \centering
  \begin{tabular}{rlcccc}
  \toprule
  \multirow{2}{*}{ID} & \multirow{2}{*}{Method} & \multicolumn{2}{c}{Found (K)} & \multicolumn{2}{c}{Total (K)} \\
  \cmidrule(lr){3-4} \cmidrule(lr){5-6}
   &  & In & Out & In & Out \\
  \midrule
  \multirow{5}{*}{5} & DoLQ(ours) & 20.4 & 12.4 & 223 & 106 \\
   & EDL & 0.61 & 0.82 & 186 & 222 \\
   & ICSR & 256 & 31.1 & 256 & 31.1 \\
   & LASR & 368 & 420 & 368 & 420 \\
   & LLM-SR & 624 & 784 & 624 & 784 \\
  \midrule
  \multirow{5}{*}{6} & DoLQ(ours) & 10.3 & 8.08 & 239 & 136 \\
   & EDL & 0.41 & 0.53 & 190 & 233 \\
   & ICSR & 255 & 29.4 & 255 & 29.4 \\
   & LASR & 345 & 394 & 345 & 394 \\
   & LLM-SR & 730 & 701 & 730 & 701 \\
  \midrule
  \multirow{5}{*}{7} & DoLQ(ours) & 183 & 89.4 & 233 & 114 \\
   & EDL & 26.9 & 29.8 & 279 & 287 \\
   & ICSR & 256 & 33.9 & 256 & 33.9 \\
   & LASR & 383 & 407 & 383 & 407 \\
   & LLM-SR & 758 & 854 & 758 & 854 \\
  \midrule
  \multirow{5}{*}{8} & DoLQ(ours) & 173 & 68.7 & 784 & 264 \\
   & EDL & 73.6 & 22.9 & 989 & 265 \\
   & ICSR & 772 & 72.9 & 772 & 72.9 \\
   & LASR & 809 & 930 & 809 & 930 \\
   & LLM-SR & 1238 & 1379 & 1852 & 1997 \\
  \bottomrule
  \end{tabular}
  \end{minipage}
\end{table}
\normalsize

\section{Analysis of optimizer adoption}
\label{app:optimizer_adoption}

\begin{figure}[h]
  \centering
  \includegraphics[width=1.0\columnwidth]{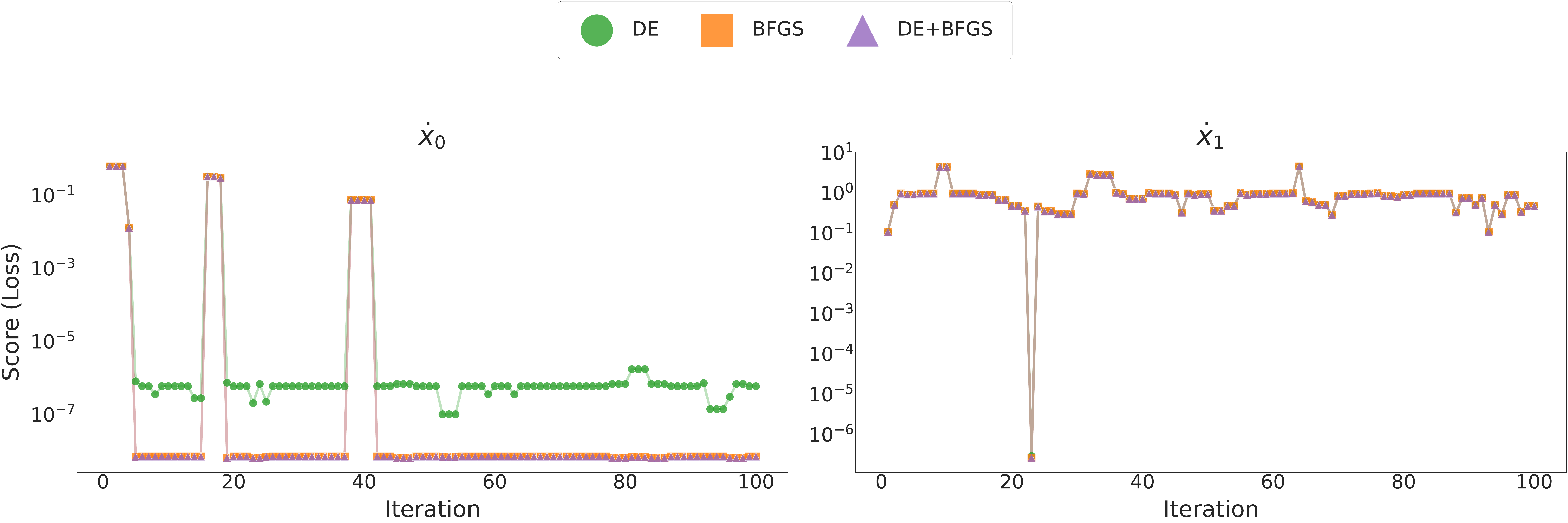}
  \caption{Adoption frequency of BFGS and differential evolution across different ODE systems. Differential evolution is more frequently selected for systems with intricate functional forms where the loss landscape is more rugged, while BFGS is often sufficient for simpler systems.}
  \label{fig:de_bfgs_adoption}
\end{figure}

To further analyze the behavior of the hybrid optimizer, we investigate the adoption frequency of BFGS and differential evolution across different problem types. As shown in Figure~\ref{fig:de_bfgs_adoption}, the selection rate of these methods varies significantly depending on the mathematical complexity of the system. For systems with simpler functional forms, BFGS is frequently sufficient to find the optimal parameters due to the relatively smooth loss landscapes. Conversely, for systems with complex functional forms (e.g., Glider), differential evolution is adopted more frequently. This demonstrates that the global search capability of differential evolution is critical for escaping local minima and accurately identifying parameters in rugged loss landscapes where local gradient-based methods often fail.

Moreover, the left panel of Figure~\ref{fig:de_bfgs_adoption} reveals a critical advantage of the hybrid optimizer. When examining $\dot{x}_0$ with the ground truth skeleton structure, we observe that differential evolution achieves lower MSE compared to using BFGS alone. This is particularly significant because assigning better scores to correct skeleton structures provides a crucial advantage during the symbolic regression search process, enabling more effective exploration and convergence to the true underlying dynamics.

\section{Quantitative results under shifted initial conditions}
\label{app:quantitative_shifted_ic}

To evaluate robustness against initial-state variations, the discovered equations are re-assessed using an alternative initial condition (ic\_1), maintaining the original ID and ID-Ext temporal regimes and NMSE metrics. The results are summarized in Table~\ref{tab:quantitative_shifted_ic}.

\begin{table}[!htbp]
  \centering
  \caption{Quantitative performance comparison measured by NMSE under shifted initial conditions. We report dimension-averaged results for the same discovered equations, re-evaluated from an alternative initial condition (ic\_1) rather than the default initial state, demonstrating robustness and generalization across initial conditions. Bold with underline indicates the best, bold indicates the second best. NaN indicates solver failure or numerical instability.}
  \label{tab:quantitative_shifted_ic}
  \footnotesize
  \setlength{\tabcolsep}{4pt}
  \begin{tabular}{lllccccc}
    \toprule
    \multirow{2}{*}{Benchmark} & \multirow{2}{*}{Metric} & \multirow{2}{*}{Regime} & \multicolumn{5}{c}{Model} \\
    \cmidrule(lr){4-8}
     &  &  & ICSR & L{\small{A}}SR & LLM-SR & EDL & DoLQ(ours) \\
    \midrule
    \multirow{4}{*}{SIR(2D)}
     & \multirow{2}{*}{Residual} & ID & \textbf{1.47e-8} & \textbf{\underline{1.41e-8}} & 2.18e-4 & 1.23e2 & 1.35e-6 \\
     &  & ID-Ext & \textbf{\underline{1.22e-9}} & \textbf{3.30e-9} & 2.97e-3 & NaN & 7.15e-5 \\
    \cmidrule(lr){2-8}
     & \multirow{2}{*}{Integral} & ID & \textbf{1.44e-8} & \textbf{\underline{1.39e-8}} & 2.09e-4 & 1.29e2 & 1.37e-6 \\
     &  & ID-Ext & \textbf{\underline{1.13e-9}} & \textbf{3.67e-9} & 8.02e-4 & NaN & 3.57e-5 \\
    \midrule
    \multirow{4}{*}{CDIMA(2D)}
     & \multirow{2}{*}{Residual} & ID & \textbf{4.53e-2} & 4.35e-1 & 4.80e2 & 6.10e5 & \textbf{\underline{2.32e-8}} \\
     &  & ID-Ext & 1.12e-1 & NaN & \textbf{6.89e-3} & NaN & \textbf{\underline{3.41e-8}} \\
    \cmidrule(lr){2-8}
     & \multirow{2}{*}{Integral} & ID & 2.60e-2 & 4.04e-1 & \textbf{2.42e-2} & 3.67e5 & \textbf{\underline{1.78e-8}} \\
     &  & ID-Ext & 1.26e-1 & NaN & \textbf{1.59e-2} & NaN & \textbf{\underline{9.22e-8}} \\
    \midrule
    \multirow{4}{*}{Glider(4D)}
     & \multirow{2}{*}{Residual} & ID & 7.12e0 & 6.17e0 & \textbf{2.44e-4} & 1.17e5 & \textbf{\underline{1.05e-6}} \\
     &  & ID-Ext & 1.18e3 & 9.13e2 & \textbf{7.71e-4} & NaN & \textbf{\underline{5.79e-6}} \\
    \cmidrule(lr){2-8}
     & \multirow{2}{*}{Integral} & ID & 7.43e0 & 5.90e1 & \textbf{2.76e-4} & 1.99e5 & \textbf{\underline{1.85e-6}} \\
     &  & ID-Ext & 3.44e3 & 1.94e3 & \textbf{7.22e-4} & NaN & \textbf{\underline{7.49e-6}} \\
    \bottomrule
  \end{tabular}
\end{table}

\section{Additional comparison on the 2D dimensionless Glider system (ID 2)}
\label{app:glider2d_comparison}

A detailed analysis of the 2D dimensionless Glider system (ID 2) is provided to supplement the main evaluation. Table~\ref{tab:ode033_nmse_appendix} reports per-dimension NMSE values, Table~\ref{tab:ode033_tokens_appendix} summarizes token consumption, and Table~\ref{tab:ode033_equations_appendix} lists the discovering equations across varying LLM backbones, facilitating a comprehensive comparison of accuracy, efficiency, and symbolic structure.

\begin{table}[!htbp]
\centering
\scriptsize
\setlength{\tabcolsep}{4pt}
\caption{Per-dimension NMSE on the 2D dimensionless Glider system (ID 2) when DoLQ is executed with different LLM models. All settings are identical to Table~\ref{tab:dolq_hyperparams}, with only the model changed. The table reports residual and integral NMSE on the ID and ID-Ext evaluation regimes for each state dimension.}
\label{tab:ode033_nmse_appendix}
\begin{tabular}{llcccc}
\toprule
 \multirow{2}{*}{Model} & \multirow{2}{*}{Dim} & \multicolumn{2}{c}{Residual} & \multicolumn{2}{c}{Integral} \\
 \cmidrule(lr){3-4} \cmidrule(lr){5-6}
 & & ID & ID-Ext & ID & ID-Ext \\
\midrule
\multirow{2}{*}{Gemini 2.5 Flash-Lite} & $x_0$ & $4.93 \times 10^{-9}$ & $5.83 \times 10^{-9}$ & $1.42 \times 10^{-8}$ & $2.19 \times 10^{-8}$ \\
 & $x_1$ & $3.51 \times 10^{-8}$ & $3.64 \times 10^{-8}$ & $4.69 \times 10^{-8}$ & $5.06 \times 10^{-8}$ \\
\midrule
\multirow{2}{*}{GPT-4o mini} & $x_0$ & $5.16 \times 10^{-9}$ & $5.07 \times 10^{-8}$ & $1.67 \times 10^{-3}$ & $8.87 \times 10^{-1}$ \\
 & $x_1$ & $4.04 \times 10^{-2}$ & $8.48 \times 10^{-1}$ & $7.75 \times 10^{-3}$ & $3.28 \times 10^{-1}$ \\
\midrule
\multirow{2}{*}{DeepSeek-V3.2} & $x_0$ & $4.01 \times 10^{-9}$ & $6.93 \times 10^{-8}$ & $1.33 \times 10^{-8}$ & $6.26 \times 10^{-7}$ \\
 & $x_1$ & $2.85 \times 10^{-8}$ & $5.66 \times 10^{-7}$ & $1.78 \times 10^{-8}$ & $3.22 \times 10^{-7}$ \\
\midrule
\multirow{2}{*}{Grok 4.1 Fast} & $x_0$ & $4.04 \times 10^{-9}$ & $6.88 \times 10^{-8}$ & $1.40 \times 10^{-8}$ & $9.11 \times 10^{-7}$ \\
 & $x_1$ & $2.85 \times 10^{-8}$ & $7.77 \times 10^{-7}$ & $1.79 \times 10^{-8}$ & $4.24 \times 10^{-7}$ \\
\bottomrule
\end{tabular}
\end{table}

\begin{table}[t]
\centering
\small
\caption{Token usage on the 2D dimensionless Glider system (ID 2) when DoLQ is executed with different LLM models. All settings are identical to Table~\ref{tab:dolq_hyperparams}, with only the model changed. We report the total input and output tokens consumed during symbolic ODE discovery.}
\label{tab:ode033_tokens_appendix}
\begin{tabular}{lcc}
\toprule
Model & Total input tokens & Total output tokens \\
\midrule
Gemini 2.5 Flash-Lite & 289K & 156K \\
GPT-4o mini & 259K & 89K \\
DeepSeek-V3.2 & 258K & 100K \\
Grok 4.1 Fast & 337K & 725K \\
\bottomrule
\end{tabular}
\end{table}

\begin{table}[!htbp]
\centering
\scriptsize
\caption{Final equations discovered on the 2D dimensionless Glider system (ID 2) when DoLQ is executed with different LLM models. All settings are identical to Table~\ref{tab:dolq_hyperparams}, with only the model changed. For each model, we list the identified governing equations for $\dot{x}_0$ and $\dot{x}_1$.}
\label{tab:ode033_equations_appendix}
\begin{tabular}{>{\raggedright\arraybackslash}p{0.18\textwidth} c >{\raggedright\arraybackslash}p{0.68\textwidth}}
\toprule
Model & Dim & Equation \\
\midrule
\multirow{2}{*}{Gemini 2.5 Flash-Lite} & $\dot{x}_0$ &
$-0.999934\sin(x_1) - 0.199969x_0^2 - 3.160\times10^{-5}(x_0\sin(x_1)) + 8.292\times10^{-6}$ \\
 & $\dot{x}_1$ &
$-0.999673\frac{\cos(x_1)}{x_0} - 6.162\times10^{-5}(x_0\sin(x_1)) + 1.000261x_0 - 6.655\times10^{-4}$ \\
\midrule
\multirow{2}{*}{GPT-4o mini} & $\dot{x}_0$ &
$-4.597\times10^{-5}\cos(x_1) - 0.200003x_0^2 - 0.999975\sin(x_1) + 4.598\times10^{-5}$ \\
 & $\dot{x}_1$ &
$2.654411\log(x_0) - 0.239982\cos(x_1) - 0.112718(x_0x_1) + 1.348938$ \\
\midrule
\multirow{2}{*}{DeepSeek-V3.2} & $\dot{x}_0$ &
$-6.490\times10^{-6}x_0 - 1.000029\sin(x_1) - 0.199997x_0^2 - 1.215\times10^{-5}(x_0\cos(x_1)) + 7.643\times10^{-5}\sin(2x_1) + 1.517\times10^{-5}$ \\
 & $\dot{x}_1$ &
$-1.000027\frac{\cos(x_1)}{x_0} + 0.999992x_0 + 3.221\times10^{-4}\frac{\cos(2x_1)}{x_0} + 5.269\times10^{-5}$ \\
\midrule
\multirow{2}{*}{Grok 4.1 Fast} & $\dot{x}_0$ &
$-1.000030\sin(x_1) - 0.199998x_0^2 + 7.786\times10^{-5}\sin(2x_1) - 4.216\times10^{-6}(x_0^2\cos(x_1)) + 5.934\times10^{-6}$ \\
 & $\dot{x}_1$ &
$-1.000170\frac{\cos(x_1)}{x_0} + 0.999566x_0 + 4.069\times10^{-4}\frac{\cos(2x_1)}{x_0} + 2.158\times10^{-4}(x_0\log(x_0)) + 6.012\times10^{-4}$ \\
\bottomrule
\end{tabular}
\end{table}

\section{Robustness analyses}
\label{app:robustness_analyses}

\subsection{Sigma-noise robustness}
\label{app:sigma_noise_analysis}

This section reports a sigma-noise robustness comparison on the 2D dimensionless Glider system (ID 2). DoLQ stores one system-level run per sigma, whereas ICSR, L{\small{A}}SR, and LLM-SR store per-dimension outputs. To ensure a fair comparison, all final discovered equations are re-evaluated under a common protocol. Following the paper definition, the ID regime corresponds to the trajectory over $t_{\text{start}} \sim t_{\text{end}}$, while the ID-Ext regime corresponds to the full trajectory over $t_{\text{start}} \sim t_{\text{ood\_end}}$. For each noise level $\sigma$ and state dimension, NMSE is averaged across initial conditions. Tables~\ref{tab:sigma_avg_nmse}, \ref{tab:sigma_dim_nmse}, and \ref{tab:sigma_equations} jointly summarize the averaged metrics, per-dimension metrics, and final discovered equations under these noise levels.

The results indicate that DoLQ does not intrinsically filter observation noise; its absolute accuracy deteriorates as $\sigma$ increases, despite remaining competitive with baseline methods. Robust equation discovery under noisy trajectories therefore remains a clear limitation and an important direction for future work.

\begin{table}[H]
\centering
\footnotesize
\setlength{\tabcolsep}{4pt}
\caption{Average NMSE measured across different sigma-noise levels for the dimensionless Glider system (ID 2). DoLQ stores system-level runs, whereas ICSR, L{\small{A}}SR, EDL, and LLM-SR store method-specific outputs; all values are re-evaluated under the same ID and ID-Ext regimes. Cases where optimization exceeded the 5-minute time limit or the loss became $\infty$ are reported as NaN.}
\label{tab:sigma_avg_nmse}
\begin{tabular}{lcccccc}
\toprule
Method & \multicolumn{2}{c}{$\sigma=0.001$} & \multicolumn{2}{c}{$\sigma=0.01$} & \multicolumn{2}{c}{$\sigma=0.1$} \\
\cmidrule(lr){2-3}\cmidrule(lr){4-5}\cmidrule(lr){6-7}
 & ID & ID-Ext & ID & ID-Ext & ID & ID-Ext \\
\midrule
DoLQ & 0.1086 & 0.1246 & 0.6135 & 0.7668 & 32.5799 & 50.3185 \\
ICSR & 0.1289 & 0.2188 & 1.1742 & 2.0300 & 1.2066 & 2.1584 \\
L{\small{A}}SR & 6.0070 & 6.0182 & 0.5390 & 0.6768 & NaN & NaN \\
EDL & NaN & NaN & NaN & NaN & NaN & NaN \\
LLM-SR & 4.6368 & 4.8001 & 370.2495 & 1.04e+04 & 2.17e+04 & 1.40e+05 \\
\bottomrule
\end{tabular}
\end{table}

\begin{table}[H]
\centering
\footnotesize
\setlength{\tabcolsep}{6pt}
\caption{Per-dimension NMSE measured across different sigma-noise levels for the dimensionless Glider system (ID 2). All values are recomputed from the final discovered equations using the same ID and ID-Ext evaluation protocol. Cases where optimization exceeded the 5-minute time limit or the loss became $\infty$ are reported as NaN.}
\label{tab:sigma_dim_nmse}
\begin{tabular}{ccccc}
\toprule
Method & $\sigma$ & Dim & ID & ID-Ext \\
\midrule
\multirow{6}{*}{DoLQ} & \multirow{2}{*}{0.001} & $\dot{x}_0$ & 0.0052 & 0.0078 \\
 &  & $\dot{x}_1$ & 0.2120 & 0.2413 \\
\cmidrule{2-5}
 & \multirow{2}{*}{0.01} & $\dot{x}_0$ & 0.4654 & 0.6868 \\
 &  & $\dot{x}_1$ & 0.7616 & 0.8469 \\
\cmidrule{2-5}
 & \multirow{2}{*}{0.1} & $\dot{x}_0$ & 12.5009 & 14.2665 \\
 &  & $\dot{x}_1$ & 52.6590 & 86.3704 \\
\midrule
\multirow{6}{*}{ICSR} & \multirow{2}{*}{0.001} & $\dot{x}_0$ & 0.0048 & 0.0100 \\
 &  & $\dot{x}_1$ & 0.2530 & 0.4276 \\
\cmidrule{2-5}
 & \multirow{2}{*}{0.01} & $\dot{x}_0$ & 0.5786 & 0.5851 \\
 &  & $\dot{x}_1$ & 1.7697 & 3.4750 \\
\cmidrule{2-5}
 & \multirow{2}{*}{0.1} & $\dot{x}_0$ & 0.5829 & 0.5762 \\
 &  & $\dot{x}_1$ & 1.8304 & 3.7405 \\
\midrule
\multirow{6}{*}{L{\small{A}}SR} & \multirow{2}{*}{0.001} & $\dot{x}_0$ & 11.8347 & 11.8571 \\
 &  & $\dot{x}_1$ & 0.1792 & 0.1792 \\
\cmidrule{2-5}
 & \multirow{2}{*}{0.01} & $\dot{x}_0$ & 0.6409 & 0.6412 \\
 &  & $\dot{x}_1$ & 0.4371 & 0.7125 \\
\cmidrule{2-5}
 & \multirow{2}{*}{0.1} & $\dot{x}_0$ & NaN & NaN \\
 &  & $\dot{x}_1$ & 0.2028 & 0.2428 \\
\midrule
\multirow{6}{*}{EDL} & \multirow{2}{*}{0.001} & $\dot{x}_0$ & NaN & NaN \\
 &  & $\dot{x}_1$ & NaN & NaN \\
\cmidrule{2-5}
 & \multirow{2}{*}{0.01} & $\dot{x}_0$ & NaN & NaN \\
 &  & $\dot{x}_1$ & NaN & NaN \\
\cmidrule{2-5}
 & \multirow{2}{*}{0.1} & $\dot{x}_0$ & NaN & NaN \\
 &  & $\dot{x}_1$ & NaN & NaN \\
\midrule
\multirow{6}{*}{LLM-SR} & \multirow{2}{*}{0.001} & $\dot{x}_0$ & 5.5721 & 5.8846 \\
 &  & $\dot{x}_1$ & 3.7015 & 3.7157 \\
\cmidrule{2-5}
 & \multirow{2}{*}{0.01} & $\dot{x}_0$ & 511.0606 & 1.89e+04 \\
 &  & $\dot{x}_1$ & 229.4384 & 2035.1681 \\
\cmidrule{2-5}
 & \multirow{2}{*}{0.1} & $\dot{x}_0$ & 4.31e+04 & 2.79e+05 \\
 &  & $\dot{x}_1$ & 373.0867 & 1072.5261 \\
\bottomrule
\end{tabular}
\end{table}

\begin{table*}[t]
\centering
\scriptsize
\caption{Final discovered equations under different noise levels for the dimensionless Glider system (ID 2). A dash (-) indicates that no valid equation was available for reporting.}
\label{tab:sigma_equations}
\setlength{\tabcolsep}{3pt}
\resizebox{\textwidth}{!}{%
\begin{tabular}{ccc>{\raggedright\arraybackslash}p{0.74\textwidth}}
\toprule
Method & $\sigma$ & Dim & Equation \\
\midrule
\multirow{6}{*}[-3.5em]{DoLQ} & \multirow{2}{*}{0.001} & $\dot{x}_0$ & $((((-0.17845980618743792 \cdot (x_0^2)) + (-1.0941877346694517 \cdot \sin(x_1))) + (0.05314606941174399 \cdot x_1)) + (-0.3391220551372564))$ \\
 &  & $\dot{x}_1$ & $((((-1.8940271186983297 \cdot (-9.81)) + (-0.8283632477740318 \cdot \cos(x_1))) + (-0.20272251166224334 \cdot (x_1/x_0))) + (-15.836236841336438))$ \\
\cmidrule{2-4}
 & \multirow{2}{*}{0.01} & $\dot{x}_0$ & $((((-10.284937863923174 \cdot 9.81) + (0.4911498533070019 \cdot x_1)) + (-1.9129771153833335 \cdot \sin(x_1))) + 97.78128237249153)$ \\
 &  & $\dot{x}_1$ & $((((-1.3512098187497759 \cdot \sin(x_1)) + (2.080726220812281 \cdot x_0)) + (-0.16870678464779224 \cdot (x_1 \sin(x_0)))) + (-2.1465100178836565))$ \\
\cmidrule{2-4}
 & \multirow{2}{*}{0.1} & $\dot{x}_0$ & $((((4.610553824850981 \cdot \sin(x_1)) + (-12.839898886682368 \cdot (x_0 \sin(x_1)))) + (2.454264804272983 \cdot (x_0^2 \sin(x_1)))) + 1.9735115401721728)$ \\
 &  & $\dot{x}_1$ & $((((-0.8346838384737242 \cdot (x_1 \cos(x_1))) + (-4.785671048569711 \cdot (x_0 x_1))) + (1.7968322398349674 \cdot (x_0^2 x_1))) + 9.272749877208437)$ \\
\midrule
\multirow{6}{*}[-1.5em]{ICSR} & \multirow{2}{*}{0.001} & $\dot{x}_0$ & $-0.185x_0^2 + 0.0035x_0x_1 + 0.0041x_1^2 - 1.0961\sin(x_1) - 0.1851$ \\
 &  & $\dot{x}_1$ & $0.5688x_0 - 0.0262x_1^2 + 2.0966 - 1.0063/x_0$ \\
\cmidrule{2-4}
 & \multirow{2}{*}{0.01} & $\dot{x}_0$ & $-0.9509x_0\sin(0.9597x_1) + 0.0595x_1 - 0.822$ \\
 &  & $\dot{x}_1$ & $-1.2089x_1\cos(0.4621x_0) + 5.343$ \\
\cmidrule{2-4}
 & \multirow{2}{*}{0.1} & $\dot{x}_0$ & $-0.9509x_0\sin(0.9597x_1) + 0.0595x_1 - 0.822$ \\
 &  & $\dot{x}_1$ & $-1.2089x_1\cos(0.4621x_0) + 5.343$ \\
\midrule
\multirow{6}{*}[-1.5em]{L{\small{A}}SR} & \multirow{2}{*}{0.001} & $\dot{x}_0$ & $-4.162585368621153/x_1$ \\
 &  & $\dot{x}_1$ & $\log(\sinh(x_0))$ \\
\cmidrule{2-4}
 & \multirow{2}{*}{0.01} & $\dot{x}_0$ & $-\sin(x_1) - 0.4299814235724901$ \\
 &  & $\dot{x}_1$ & $\log(|x_0| \cdot ||\cosh(x_0)| - 1.5820667442977299|) + 0.2913439363183297$ \\
\cmidrule{2-4}
 & \multirow{2}{*}{0.1} & $\dot{x}_0$ & $-1.339672350584124 \cdot \tan(\exp(x_0/x_1)/12.367644780162692)$ \\
 &  & $\dot{x}_1$ & $x_0$ \\
\midrule
\multirow{6}{*}{EDL} & \multirow{2}{*}{0.001} & $\dot{x}_0$ & - \\
 &  & $\dot{x}_1$ & - \\
\cmidrule{2-4}
 & \multirow{2}{*}{0.01} & $\dot{x}_0$ & - \\
 &  & $\dot{x}_1$ & - \\
\cmidrule{2-4}
 & \multirow{2}{*}{0.1} & $\dot{x}_0$ & - \\
 &  & $\dot{x}_1$ & - \\
\midrule
\multirow{6}{*}[-4.5em]{LLM-SR} & \multirow{2}{*}{0.001} & $\dot{x}_0$ & $5.7922x_1 + 0.3353x_0 - 1.2690x_0^2 - 10.4843 - 0.9018x_1^2 + 0.2287x_0^3 + 0.0289x_0x_1^2 + 0.0321x_0^3x_1 + 0.0383x_1^3 - 0.0019x_0^2x_1^3$ \\
 &  & $\dot{x}_1$ & $-1.6745x_0 - 0.6770\cos(x_1)/(x_0+10^{-9}) - 0.2463x_0\sin(x_1) + 1.9791x_0^2 - 0.2591x_0^3 + 0.0039x_1^2 + 0.3814x_0x_1 - 0.2232x_0^2x_1 - 0.2732/(x_0+10^{-9}) - 0.2510\sin(2x_1)$ \\
\cmidrule{2-4}
 & \multirow{2}{*}{0.01} & $\dot{x}_0$ & $-4.1139x_0^2 + 0.1036x_1^2 + 2.3645x_0x_1 + 0.0819x_0^2x_1 - 0.1884x_0x_1^2 + 0.7252x_0^3 - 0.0395x_1^3 - 1.9019\sin(\mathrm{clip}(x_0)) - 1.5588\log(\mathrm{clip}(x_0)) + 0.0039\exp(x_1)$ \\
 &  & $\dot{x}_1$ & $\big[-1.3646x_0^2\cos(x_1)\sin(x_1) - 3.1803x_0^2\sin(x_1) + 12.0188\cos(x_1) + 57.1401 - 3.1391x_0\sin(x_1) + 1.0047x_0\cos(x_1) + 0.2624x_1^3\big]/(x_0+10^{-6}) + 57.1401/(x_0+10^{-6}) - 30.5360x_1/(x_0+10^{-6}) + 0.5511x_0x_1\sin(x_1)$ \\
\cmidrule{2-4}
 & \multirow{2}{*}{0.1} & $\dot{x}_0$ & $\frac{0.5 \cdot 0.813984 \cdot x_0^2 \cdot 0.813984 \cdot \max\!\left(-125.0561 + 0.0961(-44.7349 + 22.1524x_1 - 0.2354x_1^3)^2, 0.01\right)\cos(x_1)}{0.769795} - \frac{0.5 \cdot 0.813984 \cdot x_0^2 \cdot 0.813984 \cdot 0.004265(-44.7349 + 22.1524x_1 - 0.2354x_1^3)\sin(x_1)}{0.769795} - 3.9405\sin(x_1)$ \\
 &  & $\dot{x}_1$ & $-10.8801x_1 + 0.1683x_1^3 - 9.8663\sin(x_1) - 44.2731x_0 - 0.1425x_0^2 - 87.9659/x_0 + 11.9680/x_0^2 - 15.7989x_0x_1 + 5.6508x_0^2x_1 + 199.1791$ \\
\bottomrule
\end{tabular}
}
\end{table*}

Different LLM backbones can converge to slightly different symbolic forms due to disparities in mathematical inductive bias and because early term proposals continually alter subsequent search trajectories in iterative mechanisms like DoLQ. Several distinct LLM backbones—Gemini 2.5 Flash-Lite, GPT-4o mini, DeepSeek-V3.2, and Grok 4.1 Fast—were evaluated to assess DoLQ's robustness across providers. Empirically, Gemini 2.5 Flash-Lite, DeepSeek-V3.2, and Grok 4.1 Fast successfully recovered the Glider system dynamics with only minor discrepancies in extraneous terms or coefficients. Conversely, GPT-4o mini frequently yielded qualitatively mismatched structures and inferior integral NMSE. For established benchmarks, discovery performance is evaluated primarily by NMSE and secondarily by term-level agreement with the ground truth; for entirely unknown systems, deeper scientific validation would be essential to confirm the physical validity of the inferred structures.

\subsection{Robustness to noisy descriptions}
To assess sensitivity to description noise, DoLQ was evaluated on the Glider(2D) system with modified natural-language descriptions but fixed numerical trajectories. When provided with a partially mismatched description, DoLQ retained certain meaningful components, such as $\sin(x_1)$ and $1/x_0$, yet failed to completely recover the ground-truth equations; average residual NMSE degraded to $1.19\times10^{-2}$ (ID) and $7.37\times10^{-2}$ (ID-Ext). Under a fully mismatched description, the core dynamics of $x_1$ ($x_0-\cos(x_1)/x_0$) were completely lost, and the residual NMSE further increased to $3.15\times10^{-2}$ (ID) and $1.83\times10^{-1}$ (ID-Ext). This suggests that DoLQ inherently trusts the provided qualitative description and lacks an intrinsic mechanism to correct erroneous prior information.
\section{Reasoning-trace analysis}
\label{app:reasoning_trace_analysis}

\subsection{Hallucination audit of reasoning traces}
We audited one representative Glider(2D) run at the log level and found that the most common hallucination pattern is premature mechanistic storytelling. Representative exploratory terms included $x_0\cos(x_1)$, $\log(x_0)$, and $\tanh(x_0)$. These terms were sometimes assigned confident physical narratives before sufficient numerical evidence had accumulated. This effect appeared most clearly when the prompts required explicit semantic justification for exploratory terms whose physical meaning was only weakly grounded by the system description. In our audit, these narratives affected search direction but not final acceptance, because every candidate still had to pass coefficient optimization, residual evaluation, and iterative comparison against competing candidates. The over-interpreted terms were not retained in the final discovered equation.

\section{Time complexity analysis}
\label{app:time_complexity}

The computational cost of DoLQ is analyzed by decoupling LLM token processing from numerical parameter optimization. As these operations utilize fundamentally different resources, merging them into a single asymptotic bound would obfuscate the system's true computational load. Table~\ref{tab:dolq_time_complexity_notation} outlines the notation adopted for this complexity analysis.

\begin{table}[h]
  \centering
  \caption{Notation used in the time complexity analysis of DoLQ.}
  \label{tab:dolq_time_complexity_notation}
  \small
  \renewcommand{\arraystretch}{1.15}
  \begin{tabular}{ll}
    \toprule
    Symbol & Meaning \\
    \midrule
    $T$ & Total number of DoLQ search cycles \\
    $H$ & Number of candidate hypotheses evaluated per cycle \\
    $d_{\mathrm{sys}}$ & Number of state dimensions in the ODE system \\
    $N$ & Number of sampled time points used in the residual objective \\
    $k_j$ & Number of symbolic terms in the $j$-th equation \\
    $p_j$ & Number of trainable parameters in the $j$-th equation \\
    $G$ & Number of differential evolution generations \\
    $NP$ & Differential evolution population size \\
    $I_j$ & Number of BFGS refinement iterations for dimension $j$ \\
    $L_{\mathrm{samp}}$ & Total input and output tokens processed by the Sampler Agent per cycle \\
    $L_{\mathrm{sci}}$ & Total input and output tokens processed by the Scientist Agent per cycle \\
    \bottomrule
  \end{tabular}
\end{table}

\paragraph{LLM cost.}
At each DoLQ cycle, the Sampler Agent and Scientist Agent are each invoked once. If their total input and output token counts are denoted by $L_{\mathrm{samp}}$ and $L_{\mathrm{sci}}$, respectively, then the token-processing cost per cycle is
\begin{equation}
O(L_{\mathrm{samp}} + L_{\mathrm{sci}}).
\end{equation}
The Scientist Agent can reduce practical runtime by filtering poor structures early, but this affects the average search trajectory rather than the worst-case asymptotic form of a single LLM call.

\paragraph{Parameter optimization cost.}
For the $j$-th dimension, one evaluation of the residual objective in Eq.~\eqref{eq:residual} over $N$ time points costs $O(N k_j)$, assuming each symbolic primitive in a candidate term is evaluated in constant time. Differential evolution therefore incurs
\begin{equation}
O(G \cdot NP \cdot N \cdot k_j).
\end{equation}
Using dense BFGS refinement adds repeated residual evaluations together with quasi-Newton updates over $p_j$ trainable parameters, yielding
\begin{equation}
O\left(I_j (N k_j + p_j^2)\right).
\end{equation}
Since DoLQ evaluates BFGS alone, differential evolution alone, and the hybrid differential-evolution-plus-BFGS strategy before selecting the best result, the constant factor increases, but the asymptotic optimization cost per dimension remains
\begin{equation}
O\left((G \cdot NP + I_j) N k_j + I_j p_j^2\right).
\end{equation}

\paragraph{Scientist quantitative evaluation cost.}
The quantitative evaluation in Algorithm~\ref{alg:simple_ablation_test} performs one baseline residual evaluation and then one additional residual evaluation per term after temporarily zeroing its coefficient. For the $j$-th dimension, this requires $k_j + 1$ residual evaluations, so the total cost is
\begin{equation}
O\left((k_j + 1) N k_j\right) = O(N k_j^2).
\end{equation}
This term is separate from the LLM reasoning cost and must be counted explicitly in the overall complexity.

\paragraph{Lightweight bookkeeping.}
Function construction, JSON parsing, and keep/hold/remove bookkeeping are linear in the number of proposed terms, i.e.,
\begin{equation}
O\left(\sum_{j=1}^{d_{\mathrm{sys}}} k_j\right),
\end{equation}
which is dominated by parameter optimization and ablation.

\paragraph{Overall complexity.}
For one DoLQ cycle that evaluates $H$ candidate hypotheses, the numerical computation scales as
\begin{equation}
O\left(
H \sum_{j=1}^{d_{\mathrm{sys}}}
\left[
(G \cdot NP + I_j) N k_j + I_j p_j^2 + N k_j^2
\right]
\right),
\end{equation}
while the LLM token-processing cost scales as
\begin{equation}
O(L_{\mathrm{samp}} + L_{\mathrm{sci}}).
\end{equation}
Over $T$ search cycles, the total complexity becomes
\begin{equation}
O\left(T (L_{\mathrm{samp}} + L_{\mathrm{sci}})\right)
\end{equation}
for token processing and
\begin{equation}
O\left(
T H \sum_{j=1}^{d_{\mathrm{sys}}}
\left[
(G \cdot NP + I_j) N k_j + I_j p_j^2 + N k_j^2
\right]
\right)
\end{equation}
for numerical computation.

\paragraph{Practical implication.}
The main numerical advantage of DoLQ comes from using residual MSE during optimization. Unlike integral-based objectives such as Eq.~\eqref{eq:mse}, Eq.~\eqref{eq:residual} does not require solving the ODE at every optimizer step, so the inner loop is reduced to direct residual evaluation on sampled data. Integral NMSE remains useful for final evaluation, but it is not part of the optimization loop analyzed above. Similarly, early filtering by the Scientist Agent improves practical efficiency by reducing the average number of poor candidates and unnecessary terms, rather than by changing the worst-case asymptotic order.

\section{Efficiency analysis}
\label{app:efficiency_analysis}

To ensure an equitable baseline comparison, LLM-SR's runtime metrics are aggregated across all target dimensions, reflecting its dimension-wise execution paradigm compared to DoLQ's joint system-level discovery. Evaluations were performed in identical environments (Intel Core i7-12700K, 20 logical cores, 94 GiB RAM). Table~\ref{tab:rebuttal_all_cpu_runtime} details the empirical CPU-time, wall-clock duration, total elapsed time, and peak memory usage.

\begin{table}[h]
\centering
\caption{CPU, runtime, and peak-memory comparison of DoLQ and LLM-SR. Total CPU time is the summed CPU time, total wall time is the wall-clock runtime, end-to-end elapsed time is the full elapsed time, and peak memory is the maximum resident memory usage in GiB.}
\footnotesize
\setlength{\tabcolsep}{4pt}
\begin{tabular}{llrrrr}
\toprule
Problem (ID) & Method & Total CPU time (s) & Total wall time (s) & End-to-end elapsed (s) & Peak memory (GiB) \\
\midrule
\multirow{2}{*}{2} & DoLQ & 12.169 & 568.095 & 608 & 0.234 \\
& LLM-SR & 6.781 & 3080.691 & 3080.691 & 0.421 \\
\midrule
\multirow{2}{*}{3} & DoLQ & 12.300 & 744.724 & 775 & 0.235 \\
& LLM-SR & 6.848 & 2293.902 & 2296.985 & 0.420 \\
\midrule
\multirow{2}{*}{8} & DoLQ & 12.660 & 712.452 & 746 & 0.238 \\
& LLM-SR & 14.363 & 6406.439 & 6411.136 & 0.422 \\
\bottomrule
\end{tabular}
\label{tab:rebuttal_all_cpu_runtime}
\end{table}

Given LLM-SR's decoupled execution, direct comparison necessitates aggregating its CPU and temporal metrics while extracting the highest peak memory recorded across per-dimension runs. The maximum value accurately captures the instantaneous memory ceiling rather than an artificial sum. Consequently, while DoLQ does not uniformly outperform baselines in raw CPU computation time, it exhibits systematically lower wall-clock duration, total elapsed time, and peak memory footprints.

Floating-point operations (FLOPs) are intentionally omitted because the hybrid pipeline integrates remote LLM inference with local numerical optimization, rendering hardware-agnostic FLOP estimation both ill-defined and challenging to reproduce. Efficiency is thus benchmarked primarily through computation time and memory allocation.

\section{Prompts and responses}
\label{app:prompts}

\subsection{Sampler agent prompt and response}
\label{app:sampler_prompt}

We present example prompts and outputs for the Sampler and Scientist agents in Figure~\ref{fig:sampler_prompt_box} through Figure~\ref{fig:scientist_output_example}. The labeled sections in these figures explicitly correspond to the components illustrated in the framework overview (Figure~\ref{fig:overview}), detailing the specific instructions and constraints provided to each agent during the discovery process.

\begin{figure}[!htbp]
    \centering
    \begin{tcolorbox}[
        enhanced,
        title=Sampler Prompt,
        colframe=black!70,
        colback=white,
        coltitle=white,
        colbacktitle=black!70,
        attach boxed title to top center={yshift=-10pt},
        boxed title style={
            frame hidden,
            rounded corners,
            arc=5pt,
        },
        fonttitle=\bfseries\sffamily,
        boxrule=0.5pt,
        drop shadow=black!30!white,
        left=4mm, right=4mm, top=6mm, bottom=4mm,
        arc=2mm
    ]
    \ttfamily\scriptsize
    \begin{tcolorbox}[enhanced, colframe=red!75!black, colback=white!95!gray, boxrule=1pt, arc=0pt, 
        overlay={\node[anchor=south east, fill=red!75!black, text=white, inner sep=2pt] at (frame.south east) {\bfseries (1a)};}]
    You are a helpful assistant tasked with discovering mathematical term structures for scientific systems. 
    Complete the 'term\_list' below, considering the physical meaning and relationships of inputs.
    
    \vspace{0.2cm}
    \textbf{\# System Description} \\
    Glider flight experiment. State variables: forward velocity x0, path angle x1, horizontal position x2, vertical altitude x3. [.... omitted....]
    
    \vspace{0.2cm}
    \textbf{\#\#\# SCIENTIST AGENT GUIDANCE} \\
    The Scientist agent has analyzed previous experiments and provides the following guidance:
    
    \textbf{\#\#\#\# Accumulated Knowledge (Theory)} \\
    - The kinematic terms for horizontal and vertical position (x2\_t and x3\_t) show strong physical relevance, particularly components involving x0 and trigonometric functions of x1. The damping term in x0\_t and the velocity-dependent term in x1\_t demonstrate good physical grounding. However, constant terms lack clear physical justification.
    \end{tcolorbox}

    \begin{tcolorbox}[enhanced, colframe=red!75!black, colback=white!95!gray, boxrule=1pt, arc=0pt,
        overlay={\node[anchor=south east, fill=red!75!black, text=white, inner sep=2pt] at (frame.south east) {\bfseries (1b)};}]
    \textbf{\#\#\#\# Term-by-Term Evaluation (Previous Attempt Analysis)} \\
    Evaluation results for each term. keep = retain, hold = hold/modify, remove = eliminate: \\
    x0\_t: \\
    - C*(x0) : \textbf{\textcolor{customgreen}{KEEP}} \\
    - C*(np.sin(x1)) : \textbf{\textcolor{customgreen}{KEEP}} \\
    - C*x0**3 : \textbf{\textcolor{customred}{REMOVE}} \\
    x1\_t: \\
    - C*(np.cos(x1)) : \textbf{\textcolor{customgreen}{KEEP}} \\
    - C*(x0) : \textbf{\textcolor{customblue}{HOLD}} \\
    - C*x1**3 : \textbf{\textcolor{customred}{REMOVE}} \\
    x2\_t: \\
    - C*(x0*np.cos(x1)) : \textbf{\textcolor{customgreen}{KEEP}} \\
    x3\_t: \\
    - C*(x0*np.sin(x1)) : \textbf{\textcolor{customgreen}{KEEP}} \\
    - C*x0*x1*x2 : \textbf{\textcolor{customred}{REMOVE}} \\
    
    \vspace{0.2cm}
    \textbf{\#\#\#\# Removed Terms List (\textcolor{customred}{Ban List})} \\
    The following term structures have negatively impacted performance. \textbf{Do NOT propose them again}: \\
    x0\_t: C*(9.81), C*(np.cos(x1)), C*(x0*x1) \\
    x1\_t: C*(1), C*(np.sin(x1)), C*(x0*x1), C*(x3) \\
    x2\_t: C*(9.81), C*(x1) \\
    x3\_t: C*(x0), C*(x1), C*(x3)
    \end{tcolorbox}

    \begin{tcolorbox}[enhanced, colframe=red!75!black, colback=white!95!gray, boxrule=1pt, arc=0pt,
        overlay={\node[anchor=south east, fill=red!75!black, text=white, inner sep=2pt] at (frame.south east) {\bfseries (1c)};}]
    \textbf{Goal:} Reflect the Scientist's insights and guidance in the equation structure.
    
    \vspace{0.2cm}
    [Required Conditions (Violation Will Cause Errors)] \\
    1. You can use: import numpy as np \\
    2. Target System Context: Input variables are x0, x1, x2, x3. \\
    - This system is 4-dimensional \\
    - Variables x4 and above do not exist. \\
    3. Term Format: Propose terms WITHOUT coefficients. The system will automatically attach trainable parameters. \\
    - Correct: "x0", "np.sin(x0)", "x0*x1" \\
    - Incorrect: "params[0]*x0", "C*x0", "0.5*x0" \\
    4. Term Complexity: You MAY use internal constants if they have physical meaning (e.g., frequency, phase). \\
    - Example: "np.sin(2*x0)" is allowed and encouraged if the factor 2 is significant. \\
    - Note: The system will still attach an outer trainable parameter (e.g., params[0]*np.sin(2*x0)). \\
    5. Symbolic Constants: Do NOT use symbolic constants like 'g', 'k', 'm'. Use numerical values. \\
    - Correct: "9.81*x0" (if g=9.81 is known), "np.pi*x0" \\
    - Incorrect: "g*x0" (will cause NameError) \\
    6. No duplicates: Equations identical to previous attempts are forbidden. Structural modifications are required. \\
    7. Reasoning required: When proposing each term, provide a physical/mathematical reasoning based on the system description (desc).
    
    \vspace{0.2cm}
    [Example (4D System)] \\
    x0\_t: ["x0", "x1*x2", "x3"] \\
    x1\_t: ["x0", "np.sin(x1)"] \\
    x2\_t: ["x0*x1"] \\
    x3\_t: ["-9.81", "x0"]
    \end{tcolorbox}
    \end{tcolorbox}
    \caption{The Sampler Prompt. The labeled sections \textbf{(1a)-(1c)} correspond to the Sampler components in Figure~\ref{fig:overview}: \textbf{(1a)} System Description \& Insight, \textbf{(1b)} Feedback \& Constraints, and \textbf{(1c)} Action \& Instruction.}
    \label{fig:sampler_prompt_box}
\end{figure}

\begin{figure}[!htbp]
    \centering
    \begin{tcolorbox}[
        enhanced,
        title=Sampler LLM Response,
        colframe=black!70,
        colback=white,
        coltitle=white,
        colbacktitle=black!70,
        attach boxed title to top center={yshift=-10pt},
        boxed title style={
            frame hidden,
            rounded corners,
            arc=5pt,
        },
        fonttitle=\bfseries\sffamily,
        boxrule=0.5pt,
        drop shadow=black!30!white,
        left=4mm, right=4mm, top=6mm, bottom=4mm,
        arc=2mm
    ]
    \ttfamily\scriptsize
    \{\\
    \hspace*{0.4cm} "ode\_pairs": [\\
    \hspace*{0.8cm} \{\\
    \hspace*{1.2cm} "x0\_t": ["x0", "np.sin(x1)", "x0*x0"],\\
    \hspace*{1.2cm} "x1\_t": ["np.cos(x1)", "x0", "x0*x0"],\\
    \hspace*{1.2cm} "x2\_t": ["x0*np.cos(x1)", "x0"],\\
    \hspace*{1.2cm} "x3\_t": ["x0*np.sin(x1)", "-9.81", "x2*x2"]\\
    \hspace*{0.8cm} \},\\
    \hspace*{0.8cm} {[.... other pairs omitted....]}\\
    \hspace*{0.4cm} ]\\
    \}
    \end{tcolorbox}
    \caption{Example response from the Sampler LLM showing the structured JSON response containing candidate ODE terms for each dimension.}
    \label{fig:sampler_output_example}
\end{figure}

\subsection{Scientist agent prompt and response}
\label{app:scientist_prompt}

Figure~\ref{fig:scientist_prompt_box} and Figure~\ref{fig:scientist_output_example} present a representative Scientist-agent prompt/response pair, showing how accumulated insights, experiment summaries, and evaluation instructions are organized before the model returns term-level judgments and actions.

\begin{figure}[!htbp]
    \centering
    \begin{tcolorbox}[
        enhanced,
        title=Scientist Prompt,
        colframe=black!70,
        colback=white,
        coltitle=white,
        colbacktitle=black!70,
        attach boxed title to top center={yshift=-10pt},
        boxed title style={
            frame hidden,
            rounded corners,
            arc=5pt,
        },
        fonttitle=\bfseries\sffamily,
        boxrule=0.5pt,
        drop shadow=black!30!white,
        left=4mm, right=4mm, top=6mm, bottom=4mm,
        arc=2mm
    ]
    \ttfamily\scriptsize
    \begin{tcolorbox}[enhanced, colframe=red!75!black, colback=white!95!gray, boxrule=1pt, arc=0pt,
        overlay={\node[anchor=south east, fill=red!75!black, text=white, inner sep=2pt] at (frame.south east) {\bfseries (2a)};}]
    You are a senior scientist specializing in ODE discovery. Your role is to evaluate proposed mathematical terms and provide guidance to improve the term\_list in the next iteration. 
    
    \vspace{0.2cm}
    \textbf{Progress:} Currently on iteration 50 of 100 total
    
    \vspace{0.2cm}
    \textbf{System Description:} \\
    Glider flight experiment. State variables: forward velocity x0, path angle x1, horizontal position x2, vertical altitude x3. [.... omitted....]
    \end{tcolorbox}

    \begin{tcolorbox}[enhanced, colframe=red!75!black, colback=white!95!gray, boxrule=1pt, arc=0pt,
        overlay={\node[anchor=south east, fill=red!75!black, text=white, inner sep=2pt] at (frame.south east) {\bfseries (2b)};}]
    \textbf{Accumulated Insights:} \\
    The kinematic terms for x2\_t and x3\_t show strong physical relevance, particularly components involving x0 and trigonometric functions of x1. The damping term in x0\_t and the velocity-dependent term in x1\_t demonstrate good physical grounding. [.... omitted....]
    \end{tcolorbox}

    \begin{tcolorbox}[enhanced, colframe=red!75!black, colback=white!95!gray, boxrule=1pt, arc=0pt,
        overlay={\node[anchor=south east, fill=red!75!black, text=white, inner sep=2pt] at (frame.south east) {\bfseries (2c)};}]
    \textbf{Previous Experiment Results:} \\
    {[Global Best]} (x0\_t: 2.187e-02, x1\_t: 6.520e-01, x2\_t: 2.846e-07, x3\_t: 2.005e-07) \\
    x0\_t = -0.2688*(x0) - 9.8336*(np.sin(x1)) - 0.0279*(x0*np.cos(x1)) + 0.4777 \\
    {[.... other dimensions omitted....]}
    
    \vspace{0.2cm}
    \textbf{Current Attempt:} (x0\_t: 2.761e-02, x1\_t: 3.555e+00, x2\_t: 2.846e-07, x3\_t: 2.086e-07) \\
    x0\_t = -0.2982*(x0) - 9.8333*(np.sin(x1)) \\
    {[.... other dimensions omitted....]}
    \end{tcolorbox}
    
    \begin{tcolorbox}[enhanced, colframe=red!75!black, colback=white!95!gray, boxrule=1pt, arc=0pt,
        overlay={\node[anchor=south east, fill=red!75!black, text=white, inner sep=2pt] at (frame.south east) {\bfseries (2d)};}]
    \textbf{Task:} Perform term-by-term evaluation.
    
    \vspace{0.2cm}
    For all terms in the current attempt, evaluate the following 2 aspects: \\
    1. \textbf{Semantic Quality:} \\
    - \textbf{\textcolor{customgreen}{good}}: Clearly aligns with the physical/mathematical meaning (\textbf{Max 3 per function}) \\
    - \textbf{\textcolor{customblue}{neutral}}: Has some relevance but not essential \\
    - \textbf{\textcolor{customred}{bad}}: Unrelated to or contradicts the system \\
    
    2. \textbf{Reasoning:} Explain your evaluation in 1-2 sentences. Focus on physical meaning.
    
    \vspace{0.2cm}
    \textbf{Important Notes:} \\
    - Use coefficient analysis: Terms with coefficients near 0 are removal candidates.
    \end{tcolorbox}
    \end{tcolorbox}
    \caption{The Scientist Prompt. The labeled sections \textbf{(2a)-(2d)} correspond to the Scientist components in Figure~\ref{fig:overview}: \textbf{(2a)} Role \& Description, \textbf{(2b)} Accumulated Insights, \textbf{(2c)} Experiment Results, and \textbf{(2d)} Evaluation Task.}
    \label{fig:scientist_prompt_box}
\end{figure}

\begin{figure}[!htbp]
    \centering
    \begin{tcolorbox}[
        enhanced,
        title=Scientist LLM Response,
        colframe=black!70,
        colback=white,
        coltitle=white,
        colbacktitle=black!70,
        attach boxed title to top center={yshift=-10pt},
        boxed title style={
            frame hidden,
            rounded corners,
            arc=5pt,
        },
        fonttitle=\bfseries\sffamily,
        boxrule=0.5pt,
        drop shadow=black!30!white,
        left=4mm, right=4mm, top=6mm, bottom=4mm,
        arc=2mm
    ]
    \ttfamily\scriptsize
    \{\\
    \hspace*{0.4cm} "x0\_t": [\\
    \hspace*{0.8cm} \{"term": "C*(x0)", "semantic\_quality": "\textbf{\textcolor{customgreen}{good}}", "action": "\textbf{\textcolor{customgreen}{keep}}"\},\\
    \hspace*{0.8cm} \{"term": "C*(np.sin(x1))", "semantic\_quality": "\textbf{\textcolor{customgreen}{good}}", "action": "\textbf{\textcolor{customgreen}{keep}}"\},\\
    \hspace*{0.8cm} \{"term": "C*(x0*x0)", "semantic\_quality": "\textbf{\textcolor{customblue}{neutral}}", "action": "\textbf{\textcolor{customblue}{hold}}"\}\\
    \hspace*{0.4cm} ],\\
    \hspace*{0.4cm} {[.... x1\_t, x2\_t, x3\_t omitted....]}\\
    \}
    \end{tcolorbox}
    \caption{Example response from the Scientist LLM showing term-by-term evaluation with semantic quality assessment and action recommendations.}
    \label{fig:scientist_output_example}
\end{figure}

\subsection{Initial sampler prompt at iteration 1}
\label{app:initial_sampler_prompt}

Figure~\ref{fig:initial_sampler_prompt} shows the initialization-time Sampler prompt used before any prior evaluations are available, making explicit what context is given when accumulated knowledge, term-level feedback, and removed-term history are still empty.

\begin{figure}[H]
\centering
\begin{tcolorbox}[
        enhanced,
        title=Initial Sampler Prompt (Iteration 1),
        colframe=black!70,
        colback=white,
        coltitle=white,
        colbacktitle=black!70,
        attach boxed title to top center={yshift=-8pt},
        boxed title style={
            frame hidden,
            rounded corners,
            arc=5pt,
        },
        fonttitle=\bfseries\sffamily,
        boxrule=0.5pt,
        drop shadow=black!30!white,
        left=3mm, right=3mm, top=4mm, bottom=3mm,
        arc=2mm
    ]
    \ttfamily\scriptsize
    You are a helpful assistant tasked with discovering mathematical term structures for scientific systems.
    Complete the \texttt{term\_list} below, considering physical meaning and relationships of inputs.

    \vspace{0.05cm}
    \textbf{\# System Description} \\
    The glider is viewed as an idealized system whose motion is expected to arise from the interplay of gravity and aerodynamic forces, with speed and flight path angle evolving from assumed initial conditions.

    \vspace{0.05cm}
    \textbf{\#\#\# SCIENTIST AGENT GUIDANCE} \\
    The Scientist agent has analyzed previous experiments and provides the following guidance:

    \begin{tcolorbox}[colframe=customred, colback=white, boxrule=0.8pt, arc=0pt,
        left=0.8mm, right=0.8mm, top=0.5mm, bottom=0.5mm]
    \textbf{\#\#\#\# Accumulated Knowledge (Theory)} \\
    None. This is the first iteration, so no accumulated insight had been produced yet.
    \end{tcolorbox}

    \begin{tcolorbox}[colframe=customred, colback=white, boxrule=0.8pt, arc=0pt,
        left=0.8mm, right=0.8mm, top=0.5mm, bottom=0.5mm]
    \textbf{\#\#\#\# Term-by-Term Evaluation (Previous Attempt Analysis)} \\
    Evaluation results for each term. \texttt{keep} = retain, \texttt{hold} = hold/modify, \texttt{remove} = eliminate: \\
    None. There was no previous attempt because this prompt was given before any sampler output had been evaluated.
    \end{tcolorbox}

    \vspace{0.05cm}
    \begin{tcolorbox}[colframe=customred, colback=white, boxrule=0.8pt, arc=0pt,
        left=0.8mm, right=0.8mm, top=0.5mm, bottom=0.5mm]
    \textbf{\#\#\#\# Removed Terms List (\textcolor{customred}{Ban List})} \\
    The following term structures have negatively impacted performance. \textbf{Do NOT propose them again}: \\
    None. At iteration 1, no term skeleton had yet been removed.
    \end{tcolorbox}

    \textbf{Goal:} Reflect the Scientist's insights and guidance in the equation structure.

    \vspace{0.05cm}
    [Required Conditions (Violation Will Cause Errors)] \\
    1. You can use: \texttt{import numpy as np} \\
    2. Target System Context: Input variables are \texttt{x0}, \texttt{x1}. \\
    \hspace*{0.4cm}-- This system is 2-dimensional \\
    \hspace*{0.4cm}-- Variables \texttt{x2} and above do not exist. \\
    3. Term Format: Propose terms WITHOUT coefficients. The system will automatically attach trainable parameters. \\
    \hspace*{0.4cm}-- Correct: \texttt{"x0"}, \texttt{"np.sin(x0)"}, \texttt{"x0*x1"} \\
    \hspace*{0.4cm}-- Incorrect: \texttt{"params[0]*x0"}, \texttt{"C*x0"}, \texttt{"0.5*x0"} \\
    4. Term Complexity: You MAY use internal constants if they have physical meaning (e.g., frequency, phase). \\
    \hspace*{0.4cm}-- Example: \texttt{"np.sin(2*x0)"} is allowed and encouraged if the factor 2 is significant. \\
    \hspace*{0.4cm}-- Note: The system will still attach an outer trainable parameter (e.g., \texttt{params[0]*np.sin(2*x0)}). \\
    5. Symbolic Constants: Do NOT use symbolic constants like \texttt{'g'}, \texttt{'k'}, \texttt{'m'}. Use numerical values. \\
    \hspace*{0.4cm}-- Correct: \texttt{"9.81*x0"} (if \texttt{g=9.81} is known), \texttt{"np.pi*x0"} \\
    \hspace*{0.4cm}-- Incorrect: \texttt{"g*x0"} (will cause \texttt{NameError}) \\
    6. No duplicates: Equations identical to previous attempts are forbidden. Structural modifications are required. \\
    7. Reasoning required: When proposing each term, provide a physical/mathematical reasoning based on the system description (\texttt{desc}).

    \vspace{0.05cm}
    [Example (2D System)] \\
    \texttt{x0\_t: ["x0", "x1"]} \\
    \texttt{x1\_t: ["x0", "np.sin(x1)"]}
\end{tcolorbox}
\caption{Initial prompt to the Sampler agent at iteration 1, where accumulated knowledge, term-level evaluation, and removed-term history are empty at initialization.}
\label{fig:initial_sampler_prompt}
\end{figure}

\end{document}